%% file: main.tex
\documentclass{article}
\usepackage{iclr2027_conference,times}
\usepackage{graphicx}
\usepackage{booktabs}
\usepackage{array}
\usepackage{tabularx}
\usepackage{amsmath}
\usepackage{amssymb}
\usepackage{multirow}
\usepackage{placeins}
\usepackage{hyperref}
\usepackage{url}
\usepackage{iftex}
\ifPDFTeX\else
  \usepackage{fontspec}
  \IfFontExistsTF{TeX Gyre Termes}
    {\setmainfont{TeX Gyre Termes}}
    {\IfFontExistsTF{Times New Roman}{\setmainfont{Times New Roman}}{}}
\fi

\newcolumntype{Y}{>{\centering\arraybackslash}X}

\title{Every Cache Entry Earns Its Place: Global Allocation of Resolution and Coverage for KV Cache Compression}
\author{
Haolin Tian$^{1*}$ \quad Yuzhe Liu$^{1*}$ \quad Tonghan Wang$^{1\dagger}$\\
\normalfont $^{1}$Tsinghua University
}

\iclrfinalcopy

\begin{document}

\maketitle
\fancyhead{}
\renewcommand{\headrulewidth}{0.4pt}
\begingroup
\renewcommand{\thefootnote}{\fnsymbol{footnote}}
\footnotetext[1]{These authors contributed equally.}
\footnotetext[2]{Corresponding author. Correspondence: \texttt{thw@mail.tsinghua.edu.cn}.}
\endgroup

\begin{abstract}
As large language models (LLMs) process increasingly long contexts, KV cache storage and repeated access have become a major bottleneck. Existing KV cache compression methods rely on predefined, fixed compression rules and are typically developed around either token eviction or merging. As a result, cache resources can neither flow freely across layers, heads, and context slots, nor be jointly allocated to balance local resolution and information coverage. 
Therefore, we propose GraceKV, a global approach for the allocation of resolution and coverage in KV cache compression, and formulate the compression process as a global resource allocation problem under a fixed cache budget. 
GraceKV treats each layer--KV-head--slot combination as an atomic unit and builds a prototype tree. Leaf nodes correspond to token-level KV entries, while each internal node uses a single prototype to compress the KV space covered by its children. A set of non-overlapping nodes in the tree forms the representation of an atomic unit. Adding the root of a new tree expands information coverage, whereas splitting a selected node improves local resolution. All candidate actions compete globally for a shared cache budget. Finally, the nodes retained across all trees form the compressed KV cache.
This process adaptively determines the allocation of cache resources among atomic units globally and the balance between resolution and coverage. 
GraceKV requires no additional training, and the entire compression and inference process is performed on the GPU. 
Systematic experiments across diverse long-context tasks and compression ratios show that GraceKV ranks first in 25 of 32 settings and remains robust up to $128\times$ compression. These results validate the effectiveness of global budget allocation in coordinating information coverage and local resolution.
\end{abstract}

\section{Introduction}
Applications such as long-document understanding and complex reasoning continually push LLMs to process longer contexts \citep{bai-etal-2024-longbench,hsieh2024ruler,zhang-etal-2024-bench}. To avoid redundant computation during autoregressive generation, LLMs cache the Key and Value associated with each context token in the KV cache. As context length grows, the storage footprint of the KV cache increases linearly and is repeatedly accessed at every decoding step, leading to continuously growing GPU memory usage and memory-access overhead. These costs have become a major bottleneck limiting the efficiency and scalability of long-context inference.

\begin{figure*}[!t]
\centering
\includegraphics[width=0.95\textwidth]{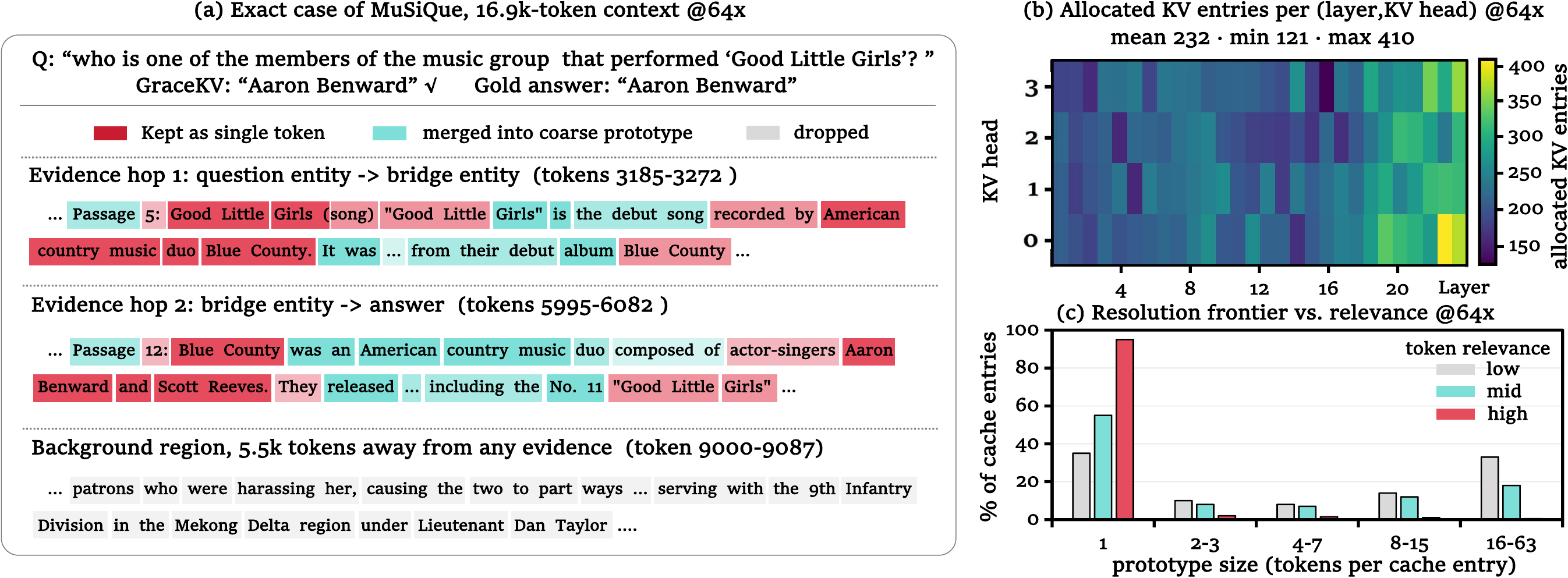} 
\caption{A representative case illustrates GraceKV's global allocation of cache resources across context regions, layers, and KV heads, preserving high-value evidence at fine resolution while aggressively compressing less relevant content.}
\label{fig:main}
\end{figure*}

To alleviate this bottleneck, KV cache compression has been widely studied. Existing methods mainly follow two directions: token eviction and KV merging. The former retains a subset of tokens according to importance scores and removes the rest, whereas the latter aggregates multiple KV entries into a small number of prototypes, trading lower local representation resolution for broader information coverage \citep{zhang2023h2o,li2024snapkv,zhang2024cam}. Moreover, existing methods typically rely on predefined, fixed compression rules. Although subsequent studies introduce adaptivity at the layer or head level, they still require the allocation granularity or local cache quotas to be specified in advance \citep{ge2023model,cai2024pyramidkv,fu2025not}. Consequently, this ``first constrain the representation form and local quota, then make compression decisions'' paradigm prevents the limited post-compression cache resources from jointly coordinating local resolution and information coverage, or from flowing freely across layers, attention heads, and context regions. Therefore, enabling limited cache resources to be adaptively allocated to more valuable places, thereby organizing and retaining contextual information more effectively, is central to efficient long-context inference.

Therefore, we propose \textbf{GraceKV}, a global approach for the allocation of resolution and coverage in KV cache compression, and formulate the compression process as a global resource allocation problem under a limited budget. To coordinate local resolution and information coverage within a unified representation space, GraceKV first partitions the context into slots through curvature-guided semantic segmentation. It then treats each layer-KV head-slot combination as an atomic unit and constructs a progressively refinable prototype tree for each atomic unit. The root uses a single prototype to represent the entire slot, each internal node uses one prototype to compressively represent the contiguous interval covered by its subtree, and the leaves correspond to the token-level K/V entries. A set of non-overlapping nodes in the tree thus constitutes a cache representation of the atomic unit at a particular resolution. Based on this representation, GraceKV defines two basic operations: \emph{Add} admits the root of a new tree to expand information coverage, whereas \emph{Split} refines a selected node to improve local resolution in a covered region. Accordingly, whether a region is covered and, if so, at what resolution it is represented can both be uniformly expressed by a set of nodes in the prototype tree.

Within this unified representation space, GraceKV further performs global resource allocation through a bottom-up \emph{value flow} and a utility-guided \emph{budget flow}. The value flow starts from token-level contextual values, aggregates them upward along the tree to nodes at different levels, and combines them with head sensitivity and the reduction in attention distortion to form the marginal utility of every candidate operation. The budget flow brings all candidate operations across atomic units into a shared global priority queue, where they compete according to marginal utility, directing the limited budget toward operations with higher returns. Together, the two flows determine the final node set of every tree, enabling globally adaptive allocation of cache resources across layers, KV heads, and context regions while balancing information coverage and local resolution. To compensate for the underestimation of non-smooth gains caused by greedy allocation, GraceKV introduces a budget-aware singleton floor, allowing a small number of high-value tokens to receive exact representations while charging their costs to the same budget. Finally, all nodes retained across the prototype trees form the physically compressed KV cache, as illustrated in Figure~\ref{fig:main}. GraceKV requires no additional training, and the entire compression and inference process runs on GPUs. When the budget is restored to the FullKV scale, GraceKV can exactly recover the original cache.

We conduct systematic experiments on LongBench and RULER across question answering, summarization, few-shot learning, aggregation, and retrieval \citep{bai-etal-2024-longbench,hsieh2024ruler}. GraceKV ranks first in 21 of 24 LongBench settings and achieves the best result in 25 of 32 settings overall. Its performance remains stable from $4\times$ to $128\times$ compression, while several baselines degrade sharply on specific tasks or under tight budgets. GraceKV also substantially reduces KV-cache memory usage and lowers post-construction per-token decode latency as context grows. Ablation studies on two backbones further confirm the contribution and consistent effectiveness of its core components across task types and cache budgets.

The main contributions are summarized as follows:
\begin{itemize}
    \item We formulate KV cache compression as a global resource allocation problem under a limited cache budget and use progressively refinable prototype trees to uniformly represent the information coverage and different resolution levels of atomic units.

    \item We propose \textbf{GraceKV}, a global approach for the allocation of resolution and coverage in KV cache compression. GraceKV performs global resource allocation through value and budget flows, requires no additional training, and runs entirely on the GPU.

    \item We validate GraceKV across multiple models, long-context tasks, and compression ratios, demonstrating strong performance, robust generalization, and efficiency gains, with ablations confirming each component.
\end{itemize}

\section{Related Work}
We focus on inference-time KV cache compression along the sequence dimension by reducing the number of cached entries.

\paragraph{Token Eviction.}
Token eviction retains a subset of the original KV entries according to importance scores and removes the rest \citep{zhang2023h2o,liu2023scissorhands,adnan2024keyformer}. StreamingLLM, H$_2$O, and SnapKV select cache positions using attention sinks, accumulated attention, and local observation windows, respectively \citep{xiao2024efficient,zhang2023h2o,li2024snapkv}. D2O performs dynamic compression using layer- and token-level signals, PyramidKV assigns different cache capacities across layers, and ChunkKV uses contiguous semantic chunks as the retention unit \citep{wan2024d2o,cai2024pyramidkv,liu2026chunkkv}. Despite their different selection strategies and granularities, the resulting cache remains a subset of the original KV entries. Consequently, each position can only be preserved exactly or removed entirely, while unselected regions can no longer participate in subsequent attention computation.

\paragraph{KV Merging.}
KV merging aggregates multiple cache entries into fewer representations, preserving partial information from compressed regions while shortening the cache \citep{zhang2024cam,wang2024model,liu2025zsmerge}. CaM merges discarded states into retained positions, and KVMerger groups entries according to Key similarity and represents each group with a weighted prototype \citep{zhang2024cam,wang2024model}. Recent methods further introduce semantic segmentation and clustering-based merging to improve the semantic integrity of merged representations \citep{hu2025efficient,wu2026semanticache}. Compared with token eviction, merging maintains broader information coverage, but fixed compression granularities and rules can lead to the loss of precise information and distortion of attention quality. In addition, some methods rely on CPU storage to recover fine-grained information on demand \citep{abaskohi2026sekv,jie2025specache}. Such designs do not truly eliminate cache overhead and additionally introduce data transfer, synchronization delays, and more complex storage management.

\paragraph{Cache Allocation.}
Early methods typically apply uniform compression rules and cache capacities across different layers and attention heads \citep{xiao2024efficient,zhang2023h2o,liu2023scissorhands,adnan2024keyformer,li2024snapkv}, ignoring the varying contextual demands of different model components. PyramidKV and Ada-KV allocate differentiated budgets at the layer and head levels, respectively, while D2O and ChunkKV adjust local compression decisions using input-dependent signals and semantic chunks \citep{cai2024pyramidkv,feng2026ada,wan2024d2o,liu2026chunkkv}. These methods gradually relax the uniform-budget assumption, but their allocation remains confined to predefined compression rules and allocation hierarchies. In contrast, GraceKV globally allocates cache resources across layers, KV heads, and context regions, jointly coordinating information coverage and local resolution without requiring additional training.

\begin{figure*}[!t]
\centering
\includegraphics[width=0.95\textwidth]{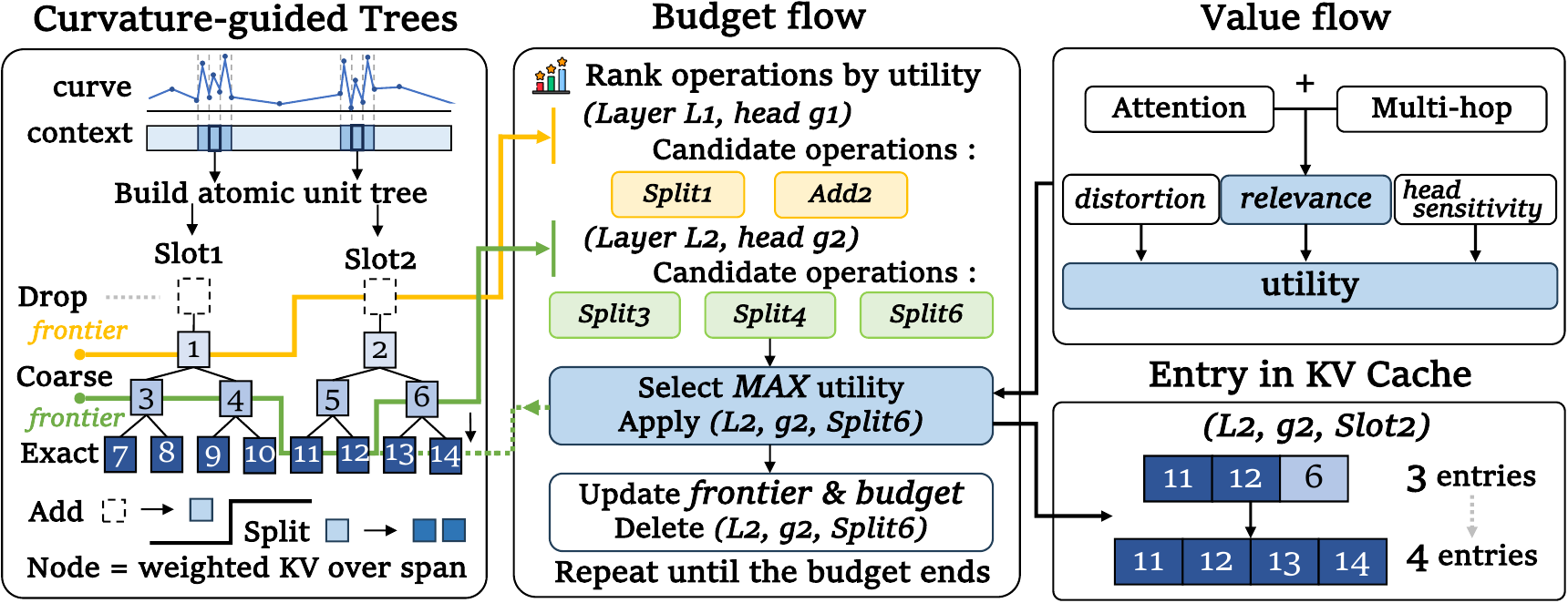} 
\caption{Overview of GraceKV: a global approach for the allocation of resolution and coverage in KV cache compression.}
\label{fig:pipeline}
\end{figure*}

\FloatBarrier
\section{Method}

The overall framework of GraceKV is illustrated in
Figure~\ref{fig:pipeline}; additional implementation details are
provided in Appendix~\ref{app:complete_method}.

\subsection{Problem Formulation}

Consider an autoregressive Transformer with \(L\) layers, where each
layer contains \(H\) query heads and \(G\) KV heads. Under grouped-query
attention, the \(g\)-th KV head is shared by the query-head set
\(\mathcal{H}(g)\). After prefill, the context cache of KV head \(g\)
at layer \(l\) is
\begin{equation}
K_{l,g},V_{l,g}\in\mathbb{R}^{T\times d},
\end{equation}
where \(T\) and \(d\) denote the input length and attention-head
dimension, respectively.

Let \(N_{\mathrm{ctx}}\) denote the context length between the prompt
prefix and the question suffix, and let \(x\) denote the target
compression ratio. The prefix and suffix are kept intact and excluded
from the cache budget. A recent window of length \(r\) at the end of
the context is also kept intact, but its physical entries are counted
toward the budget. GraceKV additionally selects a small set of
layer-KV head-token singletons outside the recent window, denoted by
\(\mathcal{E}_{\mathrm{floor}}\), to exactly preserve high-value
information. The corresponding budgets are
\begin{equation}
\begin{aligned}
B_{\mathrm{total}}
&=
\left\lfloor
\frac{LGN_{\mathrm{ctx}}}{x}
\right\rfloor,
\;
B_{\mathrm{floor}}
=
\left|\mathcal{E}_{\mathrm{floor}}\right|,
\qquad
B_{\mathrm{alloc}}
&=
B_{\mathrm{total}}-L\cdot G\cdot r-B_{\mathrm{floor}}.
\end{aligned}
\end{equation}

One budget unit corresponds to one physically stored K/V pair at a
particular layer and KV head. The remaining budget
\(B_{\mathrm{alloc}}\) is globally allocated to Add and Split
operations over the compressible context. The singleton-floor
selection procedure is introduced in the section on utility-guided
budget flow.

GraceKV performs only one standard prefill. During this pass, it
collects post-RoPE queries and keys, values, hidden states from sampled
layers, head-wise outputs in the question region, and token surprisal.
No training or additional model forward pass is required afterward.

For each layer-KV head-slot combination, GraceKV represents its
cache state using a frontier \(\mathcal{F}_{l,g,j}\) composed of
non-overlapping tree nodes. After excluding the recent window and
singleton floor, all remaining regions are initially uncovered. Let
\(\mathcal{A}\) denote the set of candidate Add and Split operations.
The allocation problem is
\begin{equation}
\begin{aligned}
\max_{\mathcal{A}'\subseteq\mathcal{A}}\;
&\sum_{a\in\mathcal{A}'}\Delta(a),
\;
\mathrm{s.t.}\;
\sum_{a\in\mathcal{A}'}c(a)
\leq B_{\mathrm{alloc}},
\\
&\mathrm{Pred}(a)\subseteq\mathcal{A}',
\quad
\forall a\in\mathcal{A}'.
\end{aligned}
\end{equation}
Here, \(\Delta(a)\), \(c(a)\), and \(\mathrm{Pred}(a)\) denote the
marginal utility, the number of newly added physical KV entries, and
the set of prerequisite ancestor operations, respectively.

\subsection{Progressively
Refinable Prototype Trees}

\paragraph{Curvature-guided semantic slot segmentation.}
We view normalized hidden states as a trajectory evolving along token
positions. For sampled layer \(l\), position \(i\), and window scale
\(w\), we define
\begin{equation}
\begin{aligned}
\mu_{i,-}^{(l,w)}
&=
\frac{1}{w}
\sum_{\tau=0}^{w-1}
\bar{h}_{i-\tau}^{(l)},
\;
\mu_{i,+}^{(l,w)}
=
\frac{1}{w}
\sum_{\tau=1}^{w}
\bar{h}_{i+\tau}^{(l)},
\\
\delta_{i,-}^{(l,w)}
&=
\bar{h}_{i}^{(l)}
-
\mu_{i,-}^{(l,w)},
\;
\delta_{i,+}^{(l,w)}
=
\mu_{i,+}^{(l,w)}
-
\bar{h}_{i}^{(l)},
\\
\kappa_{i}^{(l,w)}
&=
1-
\cos\!\left(
\delta_{i,-}^{(l,w)},
\delta_{i,+}^{(l,w)}
\right).
\end{aligned}
\end{equation}

We stabilize $\kappa_{i}^{(l,w)}$, the multi-layer and multi-scale curvature, using its mean
and standard deviation, and then combine it with token surprisal $s_i^{\mathrm{surp}}$:
\begin{equation}
\begin{aligned}
\kappa_i
&=
\operatorname{Mean}_{l,w}
\kappa_i^{(l,w)}
-
\lambda_{\mathrm{stab}}
\operatorname{Std}_{l,w}
\kappa_i^{(l,w)},
\qquad
b_i
&=
z_{\mathrm{robust}}(\kappa_i)
+
\lambda_s
z_{\mathrm{robust}}\!\left(
s_i^{\mathrm{surp}}
\right).
\end{aligned}
\end{equation}
Here, \(z_{\mathrm{robust}}\) denotes robust normalization based on the
median absolute deviation. 
We produce the slot partition
\(\mathcal{S}\)
by dynamic programming that maximizes the
boundary score conditioned on $b_i$ (see Appendix~\ref{app:segmentation_objective} for details) and penalizes the dispersion of adjacent
hidden-state differences within each slot $S$ and deviations from the
preferred slot length,
with constraint \(n_{\min}\leq |S|\leq n_{\max}\).

\paragraph{Multi-resolution prototype tree.}
For each layer \(l\), KV head \(g\), and slot \(S_j\), GraceKV
constructs a prototype tree \(\mathcal{T}_{l,g,j}\). Each node \(v\)
corresponds to a continuous interval \(I_v=[a_v,b_v)\). The root,
internal nodes, and leaves encode progressively finer representation
resolutions.

Since post-RoPE keys at different positions lie in different rotated
coordinate systems, we first map them into a shared coordinate system
and then restore the prototype at a representative position:
\begin{equation}
\begin{aligned}
k_t^{\circ}
&=
\mathrm{RoPE}^{-1}(k_t,t),
\;
p_v
=
\operatorname{round}\biggl(
\frac{1}{|I_v|}
\sum_{t\in I_v}t
\biggr),
\\
\widetilde{k}_v
&=
\mathrm{RoPE}\biggl(
\frac{1}{|I_v|}
\sum_{t\in I_v}k_t^{\circ},
p_v
\biggr).
\end{aligned}
\end{equation}

The value prototype uses token-value $R_{l,g,t}^{\mathrm{tok}}$ (defined by Eq.~\ref{eq:R_token}) weighting with a \(10\%\)
uniform base. The nominal multiplicity is further corrected by key
coherence:
\begin{equation}
\begin{aligned}
\beta_v
&=
0.1
\operatorname{Mean}_{t\in I_v}
R_{l,g,t}^{\mathrm{tok}},
\ \ 
\widetilde{v}_v
=
\frac{
\sum_{t\in I_v}
\left(
R_{l,g,t}^{\mathrm{tok}}+\beta_v
\right)v_t
}{
\sum_{t\in I_v}
\left(
R_{l,g,t}^{\mathrm{tok}}+\beta_v
\right)
},
\\
\operatorname{\hbar}_v
&=
\frac{
\left\|
\operatorname{Mean}_{t\in I_v}
k_t^{\circ}
\right\|_2
}{
\operatorname{Mean}_{t\in I_v}
\left\|k_t^{\circ}\right\|_2+\epsilon
},
\ \ 
m_v^{\mathrm{eff}}
=
\operatorname{clamp}\!\left(
|I_v|\operatorname{\hbar}_v^2,
1,
|I_v|
\right)
\end{aligned}
\end{equation}

Let \(\mathcal{Q}_v\) denote a probe set drawn from the question
suffix, the sequence suffix, and high-evidence slots. For probes that
can fully observe \(I_v\), let \(Z_q(I_v)=\sum_{t\in I_v}e^{q^\top k_t/\sqrt d}\) and \(M_q(I_v)=\sum_{t\in I_v}e^{q^\top k_t/\sqrt d}v_t\) denote
the attention partition and value aggregate of the original interval.
We define
\begin{equation}
\begin{aligned}
\widetilde{Z}_q(I_v)
&=
m_v^{\mathrm{eff}}
\exp\!\left(
\frac{q^{\top}\widetilde{k}_v}{\sqrt{d}}
\right),
\quad
\widetilde{M}_q(I_v)
=
\widetilde{Z}_q(I_v)\widetilde{v}_v,
\\
D_q(I_v)
&=
\lambda_Z
\frac{(Z_q-\widetilde Z_q)^2}{Z_q^2+\epsilon}
+
\lambda_M
\frac{\|M_q-\widetilde M_q\|_2^2}
{\|M_q\|_2^2+\epsilon}
\\
D(I_v)
&=
\operatorname{Mean}_{q\in\mathcal Q_v}
\left[
Z_q(I_v)/\sum\nolimits_{I_v}Z_q(I_v) \cdot D_q(I_v)
\right]
\\
D^{\mathrm{drop}}(I_v)
&=
\operatorname{Mean}_{q\in\mathcal Q_v}
Z_q(I_v)/\sum\nolimits_{I_v}Z_q(I_v).
\end{aligned}
\end{equation}

The attention-mass weighting prevents rarely accessed regions from
receiving high priority solely because their local approximation
errors 
$D_q(I_v)$
are large.

Each non-leaf node chooses between a curvature-aligned binary split
and salient-token extraction. The former creates two continuous
subintervals at a high-curvature position and adds one net cache
entry. The latter isolates the highest-value token in the interval
while preserving the non-empty intervals on both sides. GraceKV
constructs a unit-cost score from distortion reduction, boundary
reward, and the number of added entries, and selects the
higher-scoring construction.

\subsection{Bottom-Up Value Flow}

Prototype distortion characterizes whether a region can be represented at low resolution, but it does not indicate whether that region is relevant to the current question. GraceKV therefore estimates input-conditioned task relevance from the same
prefill pass and uses it to construct token-level context values.
Queries from the question suffix and sequence suffix form the probe
set \(\mathcal{U}\). For a compressible token \(t\), its direct
relevance is
\begin{equation}
s_{l,g,t}
=
\sum_{u\in\mathcal{U}}
\sum_{h\in\mathcal{H}(g)}
\left[
\operatorname{softmax}_{\tau\in\mathcal{V}(u)}
\left(
\frac{
\left(q_u^{l,h}\right)^{\top}
k_{\tau}^{l,g}
}{
\sqrt{d}
}
\right)
\right]_t
\end{equation}
Here, \(\mathcal{V}(u)\) is the complete visible range of probe \(u\)
under the causal mask.

We normalize \(s_{l,g,t}\) and average it across layers and KV heads to
obtain a global score $\widehat{s}_{l,g,t}$. Tokens covering a high cumulative score
form the first-hop seeds. GraceKV then performs hop-wise decayed
propagation over token attention graphs from sampled layers.
Previously selected seeds are removed at each hop, producing the
multi-hop relevance score \(\widehat{d}_t\). The final token and node
values are
\begin{equation}
\begin{aligned}
R_{l,g,t}^{\mathrm{tok}}
&=
\left(
1-\lambda_{\mathrm{graph}}
\right)
\widehat{s}_{l,g,t}
+
\lambda_{\mathrm{graph}}
\widehat{d}_t,
\qquad
R_{l,g}(v)
&=
\sum_{t\in I_v}
R_{l,g,t}^{\mathrm{tok}}.
\end{aligned}\label{eq:R_token}
\end{equation}

Node values are aggregated bottom-up from token values. The same value flow can therefore evaluate both whether a coarse region should be covered and whether an already covered region should be further refined.

Layer-KV head combinations differ in their effects on model output.
Let \(\mathcal{U}_q\) denote the probes in the question region. We
define the static output-projection strength and dynamic head output as
\begin{equation}
\begin{aligned}
a_{l,g}
&=
\left(
\sum_{h\in\mathcal{H}(g)}
\left\|
W_O^{l,h}
\right\|_F^2
\right)^{1/2},
\\
b_{l,g}
&=
\operatorname{Mean}_{u\in\mathcal{U}_q}
\left\|
\sum_{h\in\mathcal{H}(g)}
W_O^{l,h}o_u^{l,h}
\right\|_2,
\;
\omega_{l,g}
=
\sqrt{a_{l,g}b_{l,g}}.
\end{aligned}
\end{equation}
The normalized \(\omega_{l,g}\) is shared by all operations of the
same KV head as a sensitivity multiplier.

\subsection{Utility-Guided Budget Flow}

For an uncovered slot with root \(r_j\), Add introduces its root
prototype to expand coverage; for a frontier node \(v\), Split replaces
it with its children to improve resolution. Their gains and costs are
\begin{equation}
\begin{aligned}
\Delta_{\mathrm{add}}(l,g,j)
&=
\omega_{l,g}R_{l,g}(r_j)
\left[
D^{\mathrm{drop}}(I_{r_j})
-
D(I_{r_j})
\right],
\\
\Delta_{\mathrm{split}}(l,g,v)
&=
\omega_{l,g}R_{l,g}(v)
\left[
D(I_v)
-
\sum_{u\in\operatorname{child}(v)}
D(I_u)
\right],
\\
c_{\mathrm{add}}
&=
1,
\quad
c_{\mathrm{split}}(v)
=
\left|
\operatorname{child}(v)
\right|-1.
\end{aligned}
\end{equation}

Step-wise marginal allocation may underestimate non-smooth gains toward
high-value leaves. GraceKV therefore exactly preserves
\(\mathcal{E}_{\mathrm{floor}}\) with original K/V singletons before
global competition; its cost is included in \(B_{\mathrm{floor}}\) and
excluded from subsequent Add--Split allocation.

All remaining operations enter the same max-priority queue according
to their marginal utility per unit cost:
\begin{equation}
U(a)=\frac{\Delta(a)}{c(a)}.
\end{equation}
An operation is queued only after its prerequisites are completed. The
allocator repeatedly executes the operation with the highest
utility until \(B_{\mathrm{alloc}}\) is exhausted or none remains.

The retained frontier nodes and singleton floor form a physical ragged
cache for GPU decoding. When \(x=1\), all trees expand to token-level
leaves, exactly recovering FullKV.

\begin{table*}[!b]
\centering
\setlength{\abovecaptionskip}{4pt}
{
\fontsize{7.5pt}{8.5pt}\selectfont
\setlength{\tabcolsep}{2.2pt}
\renewcommand{\arraystretch}{0.90}
\begin{tabularx}{\textwidth}{@{}l@{\hspace{4pt}}*{4}{Y}@{\hspace{10pt}}*{4}{Y}@{}}
\toprule
& \multicolumn{4}{c}{\textbf{RULER (Aggregation, Set Recall $\times 100$)}}
& \multicolumn{4}{c}{\textbf{RULER (Retrieval, Set Recall $\times 100$)}}
\\
\cmidrule(lr){2-5}
\cmidrule(lr){6-9}
\textbf{Method}
& 4K & 2K & 1K & 512
& 4K & 2K & 1K & 512
\\
\midrule
FullKV
& \multicolumn{4}{c}{84.67}
& \multicolumn{4}{c}{87.00}
\\
\midrule
H2O \citep{zhang2023h2o}
& 82.00 & 79.00 & \underline{77.33} & 72.00
& 8.75 & 2.00 & 0.00 & 0.00
\\
StreamingLLM \citep{xiao2024efficient}
& 81.33 & 74.33 & 64.67 & 52.67
& 22.50 & 13.25 & 6.25 & 2.75
\\
CaM \citep{zhang2024cam}
& \textbf{85.67} & \textbf{85.00} & \textbf{80.33} & \textbf{76.67}
& 10.25 & 1.75 & 0.00 & 0.00
\\
SnapKV \citep{li2024snapkv}
& 72.67 & 67.67 & 62.67 & 60.00
& \underline{87.00} & 85.75 & 85.75 & 75.00
\\
Ada-SnapKV \citep{feng2026ada}
& 72.67 & 67.00 & 63.00 & 61.33
& \underline{87.00} & 86.00 & 85.50 & 70.00
\\
D2O \citep{wan2024d2o}
& \underline{85.00} & \underline{84.00} & 76.67 & \underline{74.33}
& 11.25 & 1.50 & 0.00 & 0.00
\\
PyramidKV \citep{cai2024pyramidkv}
& 70.33 & 64.67 & 62.67 & 59.67
& 86.00 & 85.25 & 81.00 & 60.00
\\
ChunkKV \citep{liu2026chunkkv}
& 82.00 & 75.33 & 69.00 & 64.33
& 86.25 & \underline{86.25} & \underline{86.00} & \underline{75.50}
\\
\textbf{GraceKV (Ours)}
& 82.33 & 76.67 & 70.67 & 65.00
& \textbf{87.25} & \textbf{86.75} & \textbf{86.25} & \textbf{76.50}
\\
\bottomrule
\end{tabularx}
}
\caption{Performance comparison on RULER under different KV cache budgets. The best and second-best compressed-cache results are highlighted in bold and underlined.}
\label{tab:ruler_results}
\end{table*}

\begin{table*}[!t]
\centering
\setlength{\abovecaptionskip}{4pt}

{
\fontsize{7.5pt}{8.5pt}\selectfont
\setlength{\tabcolsep}{1.0pt}
\renewcommand{\arraystretch}{0.90}

\begin{tabularx}{\textwidth}{@{}l@{\hspace{3pt}}*{6}{Y}@{\hspace{7pt}}*{6}{Y}@{}}
\toprule
&
\multicolumn{6}{c}{\textbf{LongBench (Single-Document QA, F1 $\times 100$)}}
&
\multicolumn{6}{c}{\textbf{LongBench (Multi-Document QA, F1 $\times 100$)}}
\\
\cmidrule(lr){2-7}
\cmidrule(lr){8-13}
\textbf{Method}
& $4{\times}$ & $8{\times}$ & $16{\times}$ & $32{\times}$ & $64{\times}$ & $128{\times}$
& $4{\times}$ & $8{\times}$ & $16{\times}$ & $32{\times}$ & $64{\times}$ & $128{\times}$
\\
\midrule
FullKV
& \multicolumn{6}{c}{43.95}
& \multicolumn{6}{c}{29.86}
\\
\midrule
H2O \citep{zhang2023h2o}
& 39.67 & 36.84 & 35.28 & 31.60 & 27.86 & \underline{25.27}
& 26.80 & 25.36 & 25.57 & 24.18 & 22.08 & 20.95
\\
StreamingLLM \citep{xiao2024efficient}
& 31.47 & 28.86 & 27.63 & 26.05 & 23.69 & 22.98
& 22.65 & 20.10 & 18.18 & 17.11 & 17.17 & 16.40
\\
CaM \citep{zhang2024cam}
& 39.32 & 36.93 & 34.07 & 29.93 & 25.60 & 23.96
& 26.92 & 25.52 & 25.78 & 23.78 & 23.28 & 20.15
\\
SnapKV \citep{li2024snapkv}
& 42.30 & 39.11 & 36.00 & 34.04 & 27.91 & 25.08
& 29.82 & 29.14 & \underline{28.43} & 26.86 & 25.06 & \underline{24.65}
\\
Ada-SnapKV \citep{feng2026ada}
& \underline{42.62} & 39.91 & \underline{37.09}
& \underline{34.75} & 29.25 & 25.15
& \underline{29.87} & 28.92 & 27.87
& \underline{27.20} & 24.87 & 23.88
\\
D2O \citep{wan2024d2o}
& 39.33 & 36.22 & 34.28 & 30.49 & 26.73 & 24.48
& 25.70 & 25.68 & 25.40 & 24.57 & 22.53 & 20.29
\\
PyramidKV \citep{cai2024pyramidkv}
& 42.21 & 38.81 & 35.85 & 33.76 & 26.55 & 23.86
& 29.37 & 29.57 & 27.54 & 25.78 & 23.87 & 23.48
\\
ChunkKV \citep{liu2026chunkkv}
& 42.13 & \underline{40.32} & 36.89 & 33.80 & \underline{30.22} & 24.94
& 29.50 & \textbf{30.44} & 28.40 & 26.77 & \underline{25.92} & 24.20
\\
\textbf{GraceKV (Ours)}
& \textbf{42.97} & \textbf{41.58} & \textbf{38.45}
& \textbf{35.57} & \textbf{31.49} & \textbf{25.81}
& \textbf{30.18} & \underline{29.90} & \textbf{28.78}
& \textbf{28.18} & \textbf{27.09} & \textbf{25.19}
\\
\midrule
&
\multicolumn{6}{c}{\textbf{LongBench (Summarization, Rouge-L $\times 100$)}}
&
\multicolumn{6}{c}{\textbf{LongBench (Few-Shot, Rouge-L $\times 100$)}}
\\
\cmidrule(lr){2-7}
\cmidrule(lr){8-13}
\textbf{Method}
& $4{\times}$ & $8{\times}$ & $16{\times}$ & $32{\times}$ & $64{\times}$ & $128{\times}$
& $4{\times}$ & $8{\times}$ & $16{\times}$ & $32{\times}$ & $64{\times}$ & $128{\times}$
\\
\midrule
FullKV
& \multicolumn{6}{c}{15.97}
& \multicolumn{6}{c}{36.23}
\\
\midrule
H2O \citep{zhang2023h2o}
& 14.59 & 14.25 & 13.63 & 12.99 & 11.71 & 10.79
& 35.13 & 34.44 & 33.71 & 33.10 & 32.85 & 32.43
\\
StreamingLLM \citep{xiao2024efficient}
& 14.48 & 13.52 & 12.99 & 12.55 & 12.05 & \underline{11.81}
& 34.59 & 34.47 & 33.18 & 33.02 & 32.80 & 32.31
\\
CaM \citep{zhang2024cam}
& \textbf{15.50} & \textbf{14.74} & 13.87 & 12.98 & 11.94 & 10.93
& 34.99 & \underline{34.63} & \underline{33.79} & 33.20 & 32.81 & \underline{32.79}
\\
SnapKV \citep{li2024snapkv}
& 14.81 & 14.24 & 13.68 & 13.03 & \underline{12.40} & 11.69
& 34.96 & 34.51 & 33.40 & 33.14 & 32.76 & 32.66
\\
Ada-SnapKV \citep{feng2026ada}
& 14.80 & 14.23 & 13.68 & 13.09 & 12.38 & 11.66
& \underline{35.16} & 34.27 & 33.56 & 33.42 & 32.57 & 32.06
\\
D2O \citep{wan2024d2o}
& 14.90 & 14.43 & \underline{13.90} & 13.15 & 11.78 & 11.21
& 34.80 & 34.32 & 33.78 & \underline{33.65} & \underline{33.05} & 32.76
\\
PyramidKV \citep{cai2024pyramidkv}
& 14.49 & 13.81 & 13.37 & 12.97 & 12.09 & 11.53
& 34.98 & 33.90 & 33.49 & 33.20 & \underline{33.05} & 32.17
\\
ChunkKV \citep{liu2026chunkkv}
& 14.75 & 14.24 & 13.79 & \underline{13.16} & 12.35 & 11.65
& 34.59 & 33.86 & 33.52 & 33.40 & 32.68 & 32.22
\\
\textbf{GraceKV (Ours)}
& \underline{15.03} & \underline{14.51} & \textbf{14.04}
& \textbf{13.48} & \textbf{12.58} & \textbf{11.88}
& \textbf{35.82} & \textbf{34.77} & \textbf{33.92}
& \textbf{33.80} & \textbf{33.18} & \textbf{32.83}
\\
\bottomrule
\end{tabularx}
}
\caption{Performance comparison on LongBench under different KV cache compression ratios. The best and second-best compressed-cache results are highlighted in bold and underlined.}
\label{tab:longbench_results}
\end{table*}

\section{Experiments}
\subsection{Experimental Setup}

\paragraph{Models and Datasets.}
We evaluate GraceKV on two representative long-context benchmarks,
LongBench and RULER
\citep{bai-etal-2024-longbench,hsieh2024ruler},
covering six subtasks in total. We evaluate Qwen2.5-3B-Instruct,
Qwen2.5-7B-Instruct, and Qwen2.5-14B-Instruct
\citep{qwen2025qwen25technicalreport}; Llama-3.1-8B-Instruct
\citep{grattafiori2024llama3herdmodels}; and Mistral-7B-Instruct-v0.3
\citep{jiang2023mistral7b,mistralai2024mistral7bv03}. The main results
use Qwen2.5-7B-Instruct, while the ablation study additionally includes
Llama-3.1-8B-Instruct. Complete cross-backbone results are reported in
Appendix~\ref{app:cross_backbone}.

\paragraph{Baselines.}
We compare GraceKV with FullKV and eight representative KV cache
compression methods~\citep{zhang2023h2o,xiao2024efficient,zhang2024cam,li2024snapkv,feng2026ada,wan2024d2o,cai2024pyramidkv,liu2026chunkkv}. All methods are evaluated under a unified
framework, following the commonly used protocols of the corresponding
benchmarks and using identical cache budgets at each compression ratio.

\paragraph{Evaluation Metrics.}
We adopt the official metric for each subtask. For readability, all
reported scores in the main paper are multiplied by 100. 

Additional implementation and experimental setup details are provided in Appendix Sections~\ref{app:models_tasks_metrics}--\ref{app:fairness_reproducibility}.

\subsection{Main Results}

The main results are shown in Tables~\ref{tab:ruler_results} and~\ref{tab:longbench_results}.

\paragraph{GraceKV achieves strong overall performance across the evaluated settings.}
GraceKV obtains the highest reported score in 21 out of 24 LongBench settings and ranks second in the remaining three. Together with RULER, it obtains the highest score in 25 out of 32 settings. Under the same cache-entry budget, these results demonstrate that GraceKV provides a favorable overall balance between information coverage and local resolution relative to representative token-eviction, KV-merging, and adaptive allocation methods. The result pattern is not concentrated in one favorable benchmark slice: strong scores recur across question answering, summarization, few-shot evaluation, and multiple compression ratios. RULER provides a complementary view, with the strongest advantage on Retrieval and a more competitive but task-dependent outcome on Aggregation. Together, the results support a strong balance across the evaluated conditions rather than relying on an isolated peak.

\FloatBarrier

\paragraph{GraceKV degrades gracefully as the cache budget becomes tighter.}
Beyond its overall ranking, GraceKV degrades smoothly as cache budgets tighten, rather than collapsing abruptly in particular task--budget combinations as several fixed compression strategies do. This trend indicates that global resource allocation can adjust the balance between information coverage and local resolution as the available budget changes.

\paragraph{Fixed compression strategies exhibit strong task dependence.}
The relative rankings of different baselines vary substantially across tasks, indicating that a fixed compression strategy cannot easily accommodate different information requirements. This phenomenon is particularly evident on RULER. Methods that emphasize merging and broad information coverage generally perform better on Aggregation, but degrade considerably on Retrieval, which requires precise localization of the original information. In contrast, methods that preserve more original token entries are more effective for Retrieval, but do not necessarily maintain the same advantage on Aggregation. GraceKV provides a favorable cross-task balance: while coarse-merging methods retain an advantage on Aggregation, GraceKV achieves leading performance on Retrieval and avoids the pronounced retrieval degradation observed in merging-dominated baselines. This contrast supports the value of jointly coordinating information coverage and local resolution, rather than committing to a single fixed representation form across tasks.

\subsection{Fine-grained Analysis}

\paragraph{Ablation Study.}
Table~\ref{tab:ablation_results} analyzes the contribution of each GraceKV component under two backbone models and two cache budgets. Overall, removing any component leads to performance degradation in most settings. Multi-Resolution and Global Allocation have the most pronounced effects, indicating that progressive representation and global budget allocation are key designs of GraceKV, while the remaining components also contribute under different settings. Single-document question answering is more sensitive to these components than summarization, suggesting that precise evidence localization and cross-region association rely more heavily on fine-grained representation and resource allocation. In contrast, the relatively small variations on summarization suggest that tasks dominated by distributed contextual information are more tolerant to local changes in cache representation and allocation. The two backbone models exhibit similar overall trends, indicating that these observations are not limited to a specific model.

\begin{table*}[t]
\centering
\setlength{\abovecaptionskip}{4pt}
{
\scriptsize
\setlength{\tabcolsep}{1pt}

\resizebox{\textwidth}{!}{%
\begin{tabular*}{\textwidth}{
@{\extracolsep{\fill}}
l
cc
cc
cc
cc
cc
cc
@{}
}
\toprule
&
\multicolumn{12}{c}{
\textbf{LongBench (Single-Document QA, F1 $\times 100$)}
}
\\
\cmidrule(lr){2-13}

\textbf{Model}
& \multicolumn{2}{c}{\textbf{Full}}
& \multicolumn{2}{c}{\textbf{w/o MH}}
& \multicolumn{2}{c}{\textbf{w/o SF}}
& \multicolumn{2}{c}{\textbf{w/o CS}}
& \multicolumn{2}{c}{\textbf{w/o MR}}
& \multicolumn{2}{c}{\textbf{w/o GA}}
\\
\cmidrule(lr){2-3}
\cmidrule(lr){4-5}
\cmidrule(lr){6-7}
\cmidrule(lr){8-9}
\cmidrule(lr){10-11}
\cmidrule(lr){12-13}

& $8{\times}$ & $64{\times}$
& $8{\times}$ & $64{\times}$
& $8{\times}$ & $64{\times}$
& $8{\times}$ & $64{\times}$
& $8{\times}$ & $64{\times}$
& $8{\times}$ & $64{\times}$
\\
\midrule

Qwen2.5-7B-Instruct\citep{qwen2025qwen25technicalreport}
& 41.58 & 31.49
& $-0.57$ & $-0.37$
& $-1.14$ & $-1.45$
& $-0.43$ & $-0.93$
& $-2.53$ & $-2.37$
& $-2.25$ & $-1.98$
\\

Llama-3.1-8B-Instruct\citep{grattafiori2024llama3herdmodels}
& 44.13 & 33.00
& $-0.04$ & $-1.07$
& $-0.66$ & $-0.77$
& $-0.47$ & $-0.49$
& $-2.74$ & $-1.40$
& $-1.35$ & $-1.57$
\\
\bottomrule
\end{tabular*}
}

\resizebox{\textwidth}{!}{%
\begin{tabular*}{\textwidth}{
@{\extracolsep{\fill}}
l
cc
cc
cc
cc
cc
cc
@{}
}
\toprule
&
\multicolumn{12}{c}{
\textbf{LongBench (Summarization, Rouge-L $\times 100$)}
}
\\
\cmidrule(lr){2-13}

\textbf{Model}
& \multicolumn{2}{c}{\textbf{Full}}
& \multicolumn{2}{c}{\textbf{w/o MH}}
& \multicolumn{2}{c}{\textbf{w/o SF}}
& \multicolumn{2}{c}{\textbf{w/o CS}}
& \multicolumn{2}{c}{\textbf{w/o MR}}
& \multicolumn{2}{c}{\textbf{w/o GA}}
\\
\cmidrule(lr){2-3}
\cmidrule(lr){4-5}
\cmidrule(lr){6-7}
\cmidrule(lr){8-9}
\cmidrule(lr){10-11}
\cmidrule(lr){12-13}

& $8{\times}$ & $64{\times}$
& $8{\times}$ & $64{\times}$
& $8{\times}$ & $64{\times}$
& $8{\times}$ & $64{\times}$
& $8{\times}$ & $64{\times}$
& $8{\times}$ & $64{\times}$
\\
\midrule

Qwen2.5-7B-Instruct\citep{qwen2025qwen25technicalreport}
& 14.51 & 12.58
& $-0.02$ & $-0.01$
& $-0.01$ & $+0.08$
& $-0.27$ & $-0.14$
& $-0.21$ & $-0.15$
& $-0.12$ & $-0.21$
\\

Llama-3.1-8B-Instruct\citep{grattafiori2024llama3herdmodels}
& 15.51 & 12.94
& $-0.77$ & $+0.03$
& $-0.70$ & $+0.05$
& $-0.70$ & $-0.16$
& $-0.81$ & $-0.21$
& $-0.82$ & $-0.14$
\\
\bottomrule
\end{tabular*}
}
}
\caption{Ablation study on LongBench using two backbone models.
Full reports the absolute performance of the complete GraceKV,
while the remaining variants report performance changes ($\Delta$) relative
to Full. MH, SF, CS, MR, and GA denote Multi-hop, Singleton Floor,
Curvature Segmentation, Multi-Resolution, and Global Allocation.}
\label{tab:ablation_results}
\end{table*}
\FloatBarrier

\begin{figure}[!ht]
\centering
\setlength{\abovecaptionskip}{4pt}
\includegraphics[width=0.75\textwidth]{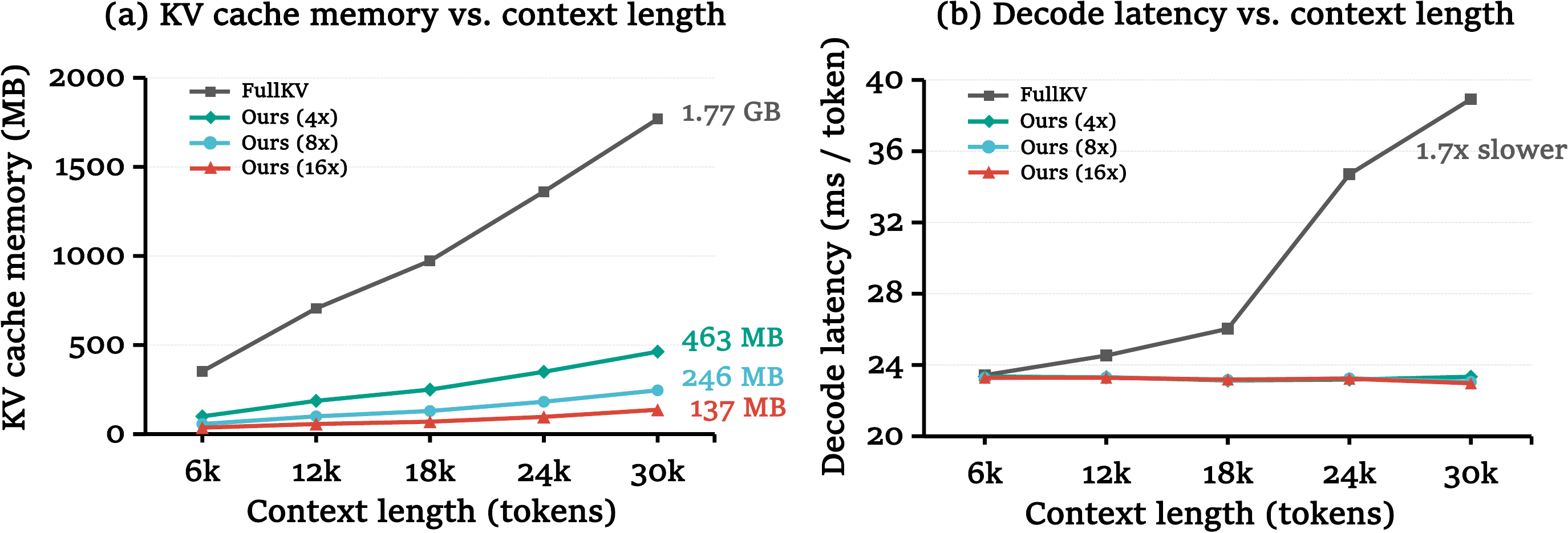}
\caption{Physical KV-cache memory and post-construction per-token decoding latency across context lengths.}
\label{fig:analysis}
\end{figure}

\paragraph{Efficiency and Scalability.}
Figure~\ref{fig:analysis} shows that GraceKV reduces physical KV-cache memory and post-construction per-token latency as context grows. Building the unified prototype trees incurs a one-time post-prefill cost of approximately 2.35 seconds, with about 8\% peak-memory overhead in the reported measurement. This cost is generally not amortized by a single response on the evaluated short-output benchmarks, whereas memory savings are immediate. Appendix~\ref{app:system_efficiency} provides detailed steady-state and break-even results; parallel tree construction and operator fusion may further reduce the one-time cost.

\paragraph{Hyperparameter Study.}
We study the singleton-floor budget, multi-hop value weight $\lambda_{\mathrm{graph}}$, target slot length, and distortion ratio $\lambda_Z:\lambda_M$. Complete results in Appendix Sections~\ref{app:sensitivity_floor_graph} and~\ref{app:sensitivity_slot_distortion} show that GraceKV remains stable over a broad range around the default configuration. Performance degrades when the singleton-floor ratio becomes too small, suggesting that greedy allocation may underestimate the non-smooth gains of high-value tokens and supporting the singleton floor.

\section{Limitations}
Achieving globally optimal allocation requires searching an exponentially large space, making it difficult to satisfy the latency requirements of online inference. GraceKV therefore adopts global greedy allocation and uses the singleton floor to mitigate the underestimation of non-smooth gains, thereby approximating the globally optimal solution. However, this strategy still cannot guarantee global optimality in the combinatorial optimization sense. In addition, prototype-tree construction introduces one-time latency and peak memory overhead, which could be further reduced through low-level and hardware-specific optimizations.

\bibliography{references}
\bibliographystyle{iclr2027_conference}

\clearpage
\appendix
\setcounter{secnumdepth}{2}

\input{appendix}

\end{document}

%% file: appendix.tex

\section{Complete Method Specification}
\label{app:complete_method}

This section presents the complete specification of GraceKV.
To eliminate implementation ambiguities arising from the space
constraints of the main paper, we start from the features
collected during a single prefill pass and describe, in order,
context partitioning, prototype-tree construction, the bottom-up
value flow, and the top-down budget flow. Unless otherwise
specified, all positional indices refer to absolute token
positions in the complete input sequence, and one cache-budget
unit corresponds to one physically stored key--value pair.

\subsection{Overview and Notation}
\label{app:overview_notation}

Consider an autoregressive Transformer with $L$ layers. Each
layer contains $H$ query heads and $G$ KV heads. Under
grouped-query attention (GQA), the $g$-th KV head is shared by
the query-head set
$\mathcal{H}(g)\subseteq\{1,\ldots,H\}$. Let $d$ denote the
dimension of one attention head and $T$ denote the total length
of the prefill input. After one standard prefill pass, the cache
of KV head $g$ at layer $l$ is
\begin{equation}
\begin{aligned}
K_{l,g},V_{l,g}
&\in\mathbb{R}^{T\times d},\\
l&\in\{1,\ldots,L\},
\qquad
g\in\{1,\ldots,G\}.
\end{aligned}
\label{eq:app_kv_cache}
\end{equation}
Here,
$k_t^{l,g}\in\mathbb{R}^{d}$ and
$v_t^{l,g}\in\mathbb{R}^{d}$ denote the post-RoPE key and
value at position $t$, respectively, while
$q_t^{l,h}\in\mathbb{R}^{d}$ denotes the post-RoPE query of
query head $h$ at layer $l$ and position $t$. When $(l,g)$ is
fixed and no ambiguity arises, we omit the superscript $(l,g)$
of the keys and values below.

The complete input is divided into a prompt prefix, a long
context, and a question suffix. Let $c_s$ and $c_e$ denote the
start and end positions of the context, respectively. The
context interval and its length are
\begin{equation}
\mathcal{C}_{\mathrm{ctx}}
=
[c_s,c_e),
\qquad
N_{\mathrm{ctx}}
=
c_e-c_s.
\label{eq:app_context_range}
\end{equation}
Given a target compression ratio $x\geq 1$, the total physical
entry budget for the contextual cache is
\begin{equation}
B_{\mathrm{total}}
=
\left\lfloor
\frac{LGN_{\mathrm{ctx}}}{x}
\right\rfloor.
\label{eq:app_total_budget}
\end{equation}
The prompt prefix $[0,c_s)$ and the question and template
suffix $[c_e,T)$ are always preserved in full and are excluded
from the budget in Eq.~\eqref{eq:app_total_budget}. A recent
window of length $r$ at the end of the context is also
preserved. It does not participate in compression decisions,
but its $LGr$ physical entries are counted toward the context
budget. We use $r=32$ by default. Only when the total budget
cannot accommodate this window is $r$ adaptively shortened,
with a minimum length of $4$. The positions that participate
in tree construction and budget competition are therefore
\begin{equation}
\mathcal{C}
=
[c_s,c_e-r),
\qquad
N_c
=
|\mathcal{C}|
=
N_{\mathrm{ctx}}-r.
\label{eq:app_compressible_context}
\end{equation}

GraceKV performs only one standard prefill pass, during which
it collects: (i) post-RoPE queries, keys, and values from all
layers; (ii) residual hidden states from several sampled
layers; (iii) per-query-head attention outputs in the question
region; and (iv) token-wise surprisal obtained from the prefill
logits. Let
$\mathcal{L}_{\mathrm{seg}}\subseteq\{1,\ldots,L\}$ denote the
sampled layers used for context segmentation and
$\mathcal{L}_{\mathrm{graph}}\subseteq\{1,\ldots,L\}$ denote
the sampled layers used for multi-hop propagation. By default,
the segmentation layers are located at relative network depths
$0.33$, $0.50$, and $0.67$, while the propagation layers are
located at depths $0.50$ and $0.75$. All these features are
obtained from the same forward pass. No training or additional
model forward pass is required afterward.

GraceKV constructs one tree for every
layer--KV-head--slot triple and represents the corresponding
cache state using a tree frontier. Let
$\mathcal{S}=\{S_j\}_{j=1}^{J}$ be a contiguous partition of
$\mathcal{C}$, where $J$ is the number of slots. The prototype
tree of layer $l$, KV head $g$, and slot $j$ is denoted by
$\mathcal{T}_{l,g,j}$, with root node $r_{l,g,j}$. A tree node
$v$ corresponds to a contiguous interval
$I_v=[a_v,b_v)\subseteq S_j$. The current frontier
$\mathcal{F}_{l,g,j}$ is a set of tree nodes whose intervals do
not overlap. An empty frontier means that the entire slot is
physically dropped. A frontier containing only the root
represents the entire slot with one coarse prototype. Expanding
frontier nodes increases local resolution, while a single-token
leaf stores the original K/V pair exactly.

GraceKV additionally selects a small number of high-value
layer--KV-head--token triples outside the recent window as
exact singleton anchors. Realizing an anchor may also
materialize prerequisite root and internal frontier entries.
We denote the complete set of physical cache entries reserved
by this stage by $\mathcal{E}_{\mathrm{floor}}$; it contains
the exact singleton leaves and all prerequisite frontier
entries that remain in the cache. Shared entries created by
multiple floor paths are included only once. Since every
element corresponds to one physically stored key--value pair,
the singleton-floor budget is
$B_{\mathrm{floor}}=|\mathcal{E}_{\mathrm{floor}}|$.
After accounting for the recent window and the singleton
floor, the budget available for free competition is
\begin{equation}
B_{\mathrm{free}}
=
B_{\mathrm{total}}
-
LGr
-
B_{\mathrm{floor}}.
\label{eq:app_free_budget}
\end{equation}
The algorithm always checks
\begin{equation}
LGr+B_{\mathrm{floor}}+B_{\mathrm{used}}
\leq B_{\mathrm{total}},
\end{equation}
where $B_{\mathrm{used}}$ is the budget already consumed by
free competition. The resulting cache is therefore physically
compressed rather than simulated through an attention mask.

\subsection{Context Partition and Atomic Units}
\label{app:context_partition}

\paragraph{Curvature-guided slot partitioning.}
To make merged intervals internally coherent, we view
normalized residual hidden states as a trajectory evolving
along token positions. For segmentation layer
$l\in\mathcal{L}_{\mathrm{seg}}$, position
$i\in\mathcal{C}$, and window scale $w\in\mathcal{W}$, let
$\kappa_i^{(l,w)}$ denote the cosine turning magnitude between
the local trajectory increments on the two sides of position
$i$. Here, $\mathcal{W}$ is the multi-scale window set, and
$\mathcal{L}_{\mathrm{seg}}$ was defined in
Section~\ref{app:overview_notation}. A larger
$\kappa_i^{(l,w)}$ indicates that position $i$ is more likely
to form a structural boundary. GraceKV stabilizes curvature
across layers and scales and combines it with token surprisal
to obtain the boundary score:
\begin{align}
\kappa_i
&=
\operatorname{Mean}_{l,w}
\kappa_i^{(l,w)}
-
\lambda_{\mathrm{stab}}
\operatorname{Std}_{l,w}
\kappa_i^{(l,w)},
\nonumber\\
b_i
&=
z_{\mathrm{robust}}(\kappa_i)
+
\lambda_s
z_{\mathrm{robust}}
\bigl(s_i^{\mathrm{surp}}\bigr).
\label{eq:app_boundary_score_summary}
\end{align}
Here, $\kappa_i$ is the stabilized curvature,
$\lambda_{\mathrm{stab}}$ controls the cross-layer and
cross-scale stability penalty,
\begin{equation}
s_i^{\mathrm{surp}}
=
-\log p(y_i\mid y_{<i})
\end{equation}
is the surprisal of input token $y_i$,
$z_{\mathrm{robust}}(\cdot)$ denotes robust normalization
based on the median absolute deviation, $\lambda_s$ is the
surprisal weight, and $b_i$ is the reward for introducing a
boundary before position $i$.

A length-constrained dynamic program combines the boundary
rewards, the dispersion of hidden-state differences within
each slot, and deviations from the preferred slot length. It
produces the contiguous partition
\begin{align}
\mathcal{S}^{*}
&=
\{S_j=[a_j,b_j)\}_{j=1}^{J},
\nonumber\\
\bigcup_{j=1}^{J}S_j
&=
\mathcal{C},
\qquad
S_j\cap S_{j'}
=
\varnothing
\quad
(j\neq j'),
\nonumber\\
n_{\min}
&\leq
|S_j|
\leq
n_{\max}.
\label{eq:app_slot_partition}
\end{align}
Here, $J$ is the number of slots, $a_j$ and $b_j$ are the
left and right endpoints of slot $j$, and $n_{\min}$ and
$n_{\max}$ are the minimum and maximum permitted slot
lengths. We use
$n_{\min}=16$,
$n_{\max}=128$, and a target length of $64$ by default.
Complete definitions of the curvature, robust boundary score,
within-slot dispersion, and dynamic-programming recurrence are
provided in Section~\ref{app:curvature_segmentation}
(\emph{Further Analysis of Curvature-Guided Slot
Segmentation}). This partition determines only the contiguous
structural ranges covered by the tree roots; it does not assign
a cache quota to any layer or attention head.

\paragraph{Atomic decision units.}
GraceKV constructs an independent prototype tree
$\mathcal{T}_{l,g,j}$ for every $(l,g,j)$. The complete set of
atomic decision units is therefore
\begin{equation}
\mathcal{U}_{\mathrm{atomic}}
=
\left\{
(l,g,j)
\;\middle|\;
\begin{aligned}
&1\leq l\leq L,\\
&1\leq g\leq G,\\
&1\leq j\leq J
\end{aligned}
\right\}.
\label{eq:app_atomic_units}
\end{equation}
The term ``atomic'' means that the coverage state and
resolution of one unit are fully represented by the frontier of
one tree. It does not imply that the unit is assigned a
predefined local budget. Candidate operations from different
units subsequently compete in the same global priority queue.
The numbers of entries allocated across layers, KV heads, and
slots therefore emerge from input-conditioned marginal
utilities.

\subsection{Progressively Refinable Prototype Trees}
\label{app:prototype_trees}

\paragraph{RoPE-consistent key prototypes.}
For tree node $v$, let
$I_v=[a_v,b_v)$ and
$n_v=|I_v|=b_v-a_v$. Post-RoPE keys at different positions
belong to different rotated coordinate systems and cannot be
averaged directly in the rotated space. We first inverse-rotate
the keys into a shared coordinate system and then apply RoPE
again at a representative position:
\begin{align}
k_t^{\circ}
&=
\operatorname{RoPE}^{-1}(k_t,t),
\nonumber\\
p_v
&=
\operatorname{round}
\left(
\frac{1}{n_v}
\sum_{t\in I_v}t
\right),
\nonumber\\
\widetilde{k}_v
&=
\operatorname{RoPE}
\left(
\frac{1}{n_v}
\sum_{t\in I_v}k_t^{\circ},
p_v
\right).
\label{eq:app_rope_prototype}
\end{align}
Here, $k_t^{\circ}$ is the key with its positional rotation
removed, $p_v$ is the integer representative position of the
node, and $\widetilde{k}_v$ is the key prototype written into
the compressed cache.

\paragraph{Value-weighted prototypes and effective multiplicity.}
Let $R_{l,g,t}^{\mathrm{tok}}\geq 0$ denote the token value at
layer $l$, KV head $g$, and position $t$, which is defined in
Section~\ref{app:value_flow}. To prevent low-value tokens from
losing all influence on the prototype, we add a $10\%$ uniform
base:
\begin{align}
\beta_v
&=
0.1
\operatorname{Mean}_{t\in I_v}
R_{l,g,t}^{\mathrm{tok}},
\nonumber\\
w_{v,t}
&=
R_{l,g,t}^{\mathrm{tok}}
+
\beta_v,
\nonumber\\
\widetilde{v}_v
&=
\frac{
\displaystyle
\sum_{t\in I_v}
w_{v,t}v_t
}{
\displaystyle
\sum_{t\in I_v}
w_{v,t}
+
\epsilon
}.
\label{eq:app_value_prototype}
\end{align}
Here, $w_{v,t}$ is the contribution of position $t$ to node
$v$, and $\widetilde{v}_v$ is the value prototype written into
the compressed cache.

Simply adding $\log n_v$ to a prototype formed by merging
$n_v$ tokens can amplify a ``ghost key'' that does not
correspond to any key in the original interval. GraceKV
therefore corrects the multiplicity using the directional
coherence of the inverse-rotated keys:
\begin{align}
\operatorname{coh}_v
&=
\frac{
\left\|
\operatorname{Mean}_{t\in I_v}
k_t^{\circ}
\right\|_2
}{
\operatorname{Mean}_{t\in I_v}
\left\|
k_t^{\circ}
\right\|_2
+
\epsilon
},
\nonumber\\
m_v^{\mathrm{eff}}
&=
\operatorname{clamp}
\left(
n_v\operatorname{coh}_v^2,
1,
n_v
\right).
\label{eq:app_effective_multiplicity}
\end{align}
Here,
$\operatorname{coh}_v\in[0,1]$ is the key coherence and
$m_v^{\mathrm{eff}}$ is the effective multiplicity. When the
key directions are highly consistent,
$m_v^{\mathrm{eff}}\approx n_v$. When they are incoherent,
the effective multiplicity is reduced toward $1$. During
decoding, the attention logit between query $q$ and node $v$
is
\begin{equation}
\ell(q,v)
=
\frac{
q^{\top}\widetilde{k}_v
}{
\sqrt{d}
}
+
\log m_v^{\mathrm{eff}}.
\label{eq:app_decode_logit}
\end{equation}
If all keys within the interval are identical, all values are
identical, and the key directions are fully coherent,
Eq.~\eqref{eq:app_decode_logit} exactly recovers the attention
partition mass of the original interval. The multiplicity term
is therefore the exact limiting form of merging identical
entries rather than an arbitrary amplitude correction.

\paragraph{Prototype distortion.}
Let $\mathcal{Q}_v$ denote the distortion-probe set of node
$v$, retaining only probes that can causally observe the entire
interval $I_v$. The set contains up to $32$ queries from the
end of the question, up to $16$ queries from the end of the
complete sequence, and two representative queries from each of
the eight slots with the highest token evidence. Under GQA,
the query heads sharing one KV head are assigned to probes in a
round-robin manner so that all heads in $\mathcal{H}(g)$ are
covered.

For $q\in\mathcal{Q}_v$, define the attention partition and
unnormalized value aggregate of the original interval as
\begin{align}
Z_q(I_v)
&=
\sum_{t\in I_v}
\exp
\left(
\frac{q^{\top}k_t}{\sqrt{d}}
\right),
\nonumber\\
M_q(I_v)
&=
\sum_{t\in I_v}
\exp
\left(
\frac{q^{\top}k_t}{\sqrt{d}}
\right)
v_t.
\label{eq:app_original_attention_stats}
\end{align}
For numerical stability, the implementation subtracts the
maximum logit of the same probe from all logits. This shift is
applied to both the original entries and the prototype and
therefore does not change the relative errors defined below.

The corresponding statistics used by the prototype-distortion
surrogate are
\begin{align}
\widetilde{Z}_q(I_v)
&=
m_v^{\mathrm{eff}}
\exp
\left(
\frac{
q^{\top}\widetilde{k}_v
}{
\sqrt{d}
}
\right),
\nonumber\\
\widetilde{M}_q(I_v)
&=
\widetilde{Z}_q(I_v)
\widetilde{v}_v.
\label{eq:app_prototype_attention_stats}
\end{align}

The local distortion under a single probe is
\begin{align}
D_q(I_v)
&=
\lambda_Z
\frac{
\left(
Z_q(I_v)-\widetilde{Z}_q(I_v)
\right)^2
}{
Z_q(I_v)^2+\epsilon
}
\nonumber\\
&\quad+
\lambda_M
\frac{
\left\|
M_q(I_v)-\widetilde{M}_q(I_v)
\right\|_2^2
}{
\left\|
M_q(I_v)
\right\|_2^2+\epsilon
}.
\label{eq:app_probe_distortion}
\end{align}
We use
$\lambda_Z=0.25$ and $\lambda_M=0.75$ to control the
partition and value-aggregation errors, respectively.
Following the notation of the main paper,
$\sum\nolimits_{I_v}Z_q(I_v)$ denotes the sum of the
interval-level partition masses included in the normalization
for probe $q$. The attention-mass-weighted node distortion and
the cost of dropping the interval completely are
\begin{align}
D(I_v)
&=
\operatorname{Mean}_{q\in\mathcal{Q}_v}
\left[
\frac{
Z_q(I_v)
}{
\sum\nolimits_{I_v}
Z_q(I_v)
}
D_q(I_v)
\right],
\nonumber\\
D^{\mathrm{drop}}(I_v)
&=
\operatorname{Mean}_{q\in\mathcal{Q}_v}
\frac{
Z_q(I_v)
}{
\sum\nolimits_{I_v}
Z_q(I_v)
}.
\label{eq:app_node_and_drop_distortion}
\end{align}
Here, $D(I_v)$ measures the cost of representing the interval
using its prototype approximation, while
$D^{\mathrm{drop}}(I_v)$ measures the attention mass lost by
removing the interval completely. The attention-mass weight
prevents a rarely accessed interval from receiving
unreasonably high priority solely because of a large local
approximation error.

\paragraph{Recursive refinement rule.}
During tree construction, every non-leaf node compares two
candidate refinement forms. The first is a curvature-aligned
binary split. Let $\mathcal{X}_v$ contain up to the $16$
highest-curvature positions inside the interval together with
its midpoint. For $\xi\in\mathcal{X}_v$, define the left and
right child intervals as
\begin{equation}
I_v^{\mathrm{L}}(\xi)
=
[a_v,\xi),
\qquad
I_v^{\mathrm{R}}(\xi)
=
[\xi,b_v).
\end{equation}
The gain and net cost of the binary split are
\begin{align}
G_{\mathrm{bin}}(v,\xi)
&=
D(I_v)
-
D\left(I_v^{\mathrm{L}}(\xi)\right)
-
D\left(I_v^{\mathrm{R}}(\xi)\right)
+
\lambda_{\kappa}
\widehat{\kappa}_{\xi},
\nonumber\\
c_{\mathrm{bin}}
&=
1,
\qquad
\lambda_{\kappa}
=
0.05.
\label{eq:app_binary_split}
\end{align}
Here, $\widehat{\kappa}_{\xi}$ is the curvature normalized
within the current node and $\lambda_{\kappa}$ is the boundary
reward. Replacing one parent entry with two child entries
increases the number of physical entries by one.

The second form is salient-token extraction. Given an
extraction position $p\in I_v$, all nonempty intervals among
$[a_v,p)$, the singleton $\{p\}$, and $[p+1,b_v)$ form the
child set $\mathcal{C}_{\mathrm{ext}}(v,p)$. The prototype
distortion of the singleton is zero. The gain and net cost are
\begin{align}
G_{\mathrm{ext}}(v,p)
&=
D(I_v)
-
\sum_{u\in\mathcal{C}_{\mathrm{ext}}(v,p)}
D(I_u),
\nonumber\\
c_{\mathrm{ext}}(v,p)
&=
\left|
\mathcal{C}_{\mathrm{ext}}(v,p)
\right|
-
1.
\label{eq:app_extract_split}
\end{align}
When $p$ lies in the interior of the interval, the operation
usually produces three children and has cost $2$. When one
side is empty because $p$ is adjacent to an interval boundary,
the operation degenerates to two children with cost $1$. Under
normal conditions, $p$ is the token with the highest attention
mass in $I_v$. If the interval contains a floor token that has
not yet become a leaf, extraction is enforced, and the
highest-value floor token in the interval is selected so that
the exact singleton can be reached through a shallow path.

The utility per unit cost of the two alternatives is
\begin{equation}
\begin{aligned}
\rho_{\mathrm{bin}}(v)
&=
\max_{\xi\in\mathcal{X}_v}
\frac{
G_{\mathrm{bin}}(v,\xi)
}{
c_{\mathrm{bin}}
},
\\
\rho_{\mathrm{ext}}(v)
&=
\frac{
G_{\mathrm{ext}}(v,p)
}{
c_{\mathrm{ext}}(v,p)
}.
\end{aligned}
\label{eq:app_split_choice}
\end{equation}
Each node preselects the refinement form with the larger score,
and the process recurses until single-token leaves are reached.
The tree is generated only once during construction. The
budget allocator does not search for split points again and
instead directly reads the preselected child nodes and costs.
For efficient implementation, nodes at the same depth across
different $(l,g)$ pairs are processed in batches, with each
batch divided according to its number of tensor elements to
control peak memory under long contexts.

\subsection{Bottom-Up Value Flow}
\label{app:value_flow}

Prototype distortion answers whether an interval can be
represented at low resolution, but not whether the interval is
important for the current input. GraceKV therefore constructs
token-level task relevance from the same prefill pass and
aggregates it upward through the trees.

\paragraph{Direct relevance.}
Let $\mathcal{U}$ denote the direct-relevance probe set,
containing up to $32$ positions from the end of the question
and up to $16$ positions from the end of the complete
sequence. For probe position $u\in\mathcal{U}$, let
$\mathcal{V}(u)$ denote its complete visible range under the
causal mask. The attention probability from query head $h$ at
layer $l$ to shared KV head $g$ is
\begin{equation}
\pi_{u,t}^{l,h,g}
=
\frac{
\exp
\left(
(q_u^{l,h})^{\top}
k_t^{l,g}/\sqrt{d}
\right)
}{
\displaystyle
\sum_{\tau\in\mathcal{V}(u)}
\exp
\left(
(q_u^{l,h})^{\top}
k_{\tau}^{l,g}/\sqrt{d}
\right)
},
\quad
t\in\mathcal{V}(u).
\label{eq:app_attention_probability}
\end{equation}
The softmax denominator covers the complete visible range of
the probe, rather than only the compressible interval. The
slice $t\in\mathcal{C}$ is taken only after the attention
probabilities have been computed. The direct score and its
normalization for each $(l,g)$ are
\begin{align}
s_{l,g,t}
&=
\sum_{u\in\mathcal{U}}
\sum_{h\in\mathcal{H}(g)}
\pi_{u,t}^{l,h,g},
\nonumber\\
\widehat{s}_{l,g,t}
&=
\frac{
s_{l,g,t}
}{
\displaystyle
\sum_{\tau\in\mathcal{C}}
s_{l,g,\tau}
+
\epsilon
},
\qquad
t\in\mathcal{C}.
\label{eq:app_direct_relevance}
\end{align}
Averaging across layers and KV heads yields the global seed
score:
\begin{equation}
S_t
=
\operatorname{Mean}_{l,g}
\widehat{s}_{l,g,t},
\qquad
\widehat{S}_t
=
\frac{
S_t
}{
\displaystyle
\sum_{\tau\in\mathcal{C}}
S_{\tau}
+
\epsilon
}.
\label{eq:app_global_seed_score}
\end{equation}
Tokens are sorted in descending order of $\widehat{S}_t$ and
selected until their cumulative mass reaches
$\eta_{\mathrm{seed}}=0.9$, with at most
$M_{\mathrm{seed}}=64$ tokens. The resulting set is the
first-hop seed set $\mathcal{U}^{(1)}$.

\paragraph{Token-to-token multi-hop propagation.}
Direct relevance primarily captures explicit attention from
the question to the context. To recover indirect dependencies
among contextual tokens, GraceKV performs
$R_{\mathrm{hop}}=2$ propagation steps over the sampled layers
$\mathcal{L}_{\mathrm{graph}}$. Let
$w_u^{(k)}\geq 0$ be the normalized weight of seed
$u\in\mathcal{U}^{(k)}$ at hop $k$, satisfying
\begin{equation}
\sum_{u\in\mathcal{U}^{(k)}}
w_u^{(k)}
=
1.
\end{equation}
The propagation score at hop $k$ is
\begin{equation}
d_t^{(k)}
=
\sum_{l\in\mathcal{L}_{\mathrm{graph}}}
\sum_{g=1}^{G}
\sum_{h\in\mathcal{H}(g)}
\sum_{u\in\mathcal{U}^{(k)}}
w_u^{(k)}
\pi_{u,t}^{l,h,g}.
\label{eq:app_hop_propagation}
\end{equation}
After evaluating Eq.~\eqref{eq:app_hop_propagation}, the
scores at the current and previous seed positions are set to
zero so that propagation is used only to discover new tokens.
The next-hop seeds are then selected according to the current
propagation score, and their weights are proportional to the
propagated mass from the preceding hop. Using
$\rho=0.5$ as the inter-hop decay factor, we aggregate and
normalize the multi-hop scores as
\begin{align}
d_t
&=
\sum_{k=1}^{R_{\mathrm{hop}}}
\rho^{k-1}
d_t^{(k)},
\nonumber\\
\widehat{d}_t
&=
\frac{
d_t
}{
\displaystyle
\sum_{\tau\in\mathcal{C}}
d_{\tau}
+
\epsilon
}.
\label{eq:app_diffused_relevance}
\end{align}
Here, $d_t$ is the accumulated multi-hop score and
$\widehat{d}_t$ is its normalized form over the compressible
context.

\paragraph{Token values and node values.}
The final token value is a convex combination of direct and
multi-hop relevance:
\begin{align}
R_{l,g,t}^{\mathrm{tok}}
&=
(1-\lambda_{\mathrm{graph}})
\widehat{s}_{l,g,t}
+
\lambda_{\mathrm{graph}}
\widehat{d}_t,
\nonumber\\
\lambda_{\mathrm{graph}}
&=
0.3,
\nonumber\\
R_{l,g}(v)
&=
\sum_{t\in I_v}
R_{l,g,t}^{\mathrm{tok}}.
\label{eq:app_bottom_up_value}
\end{align}
Here, $R_{l,g}(v)$ is the value of tree node $v$. Because the
child intervals are non-overlapping and their union equals the
parent interval, every non-leaf node satisfies
\begin{equation}
R_{l,g}(v)
=
\sum_{u\in\operatorname{child}(v)}
R_{l,g}(u),
\label{eq:app_value_conservation}
\end{equation}
where $\operatorname{child}(v)$ is the child set of $v$.
Equation~\eqref{eq:app_value_conservation} provides exact
bottom-up value conservation. The same token-level value field
can therefore price both whether a root should be covered and
whether an internal node should be refined. The final method
does not multiplicatively modulate token values using
surprisal or curvature; these signals are used only for
structural partitioning and refinement.

\paragraph{Layer--KV-head sensitivity.}
Different layer--KV-head pairs have different effects on the
model output. Let
$W_O^{l,h}\in\mathbb{R}^{D\times d}$ denote the column block
of the attention output projection corresponding to query head
$h$ at layer $l$. Let
$o_u^{l,h}\in\mathbb{R}^{d}$ denote the attention output of
query head $h$ at position $u$ before the output projection,
and let $\mathcal{U}_q$ denote the probe positions in the
question region. We define the static projection strength and
the input-conditioned dynamic strength as
\begin{align}
\omega_{l,g}^{\mathrm{static}}
&=
\left(
\sum_{h\in\mathcal{H}(g)}
\left\|
W_O^{l,h}
\right\|_F^2
\right)^{1/2},
\nonumber\\
\omega_{l,g}^{\mathrm{input}}
&=
\operatorname{Mean}_{u\in\mathcal{U}_q}
\left\|
\sum_{h\in\mathcal{H}(g)}
W_O^{l,h}o_u^{l,h}
\right\|_2,
\nonumber\\
\omega_{l,g}^{\mathrm{raw}}
&=
\sqrt{
\omega_{l,g}^{\mathrm{static}}
\omega_{l,g}^{\mathrm{input}}
}.
\label{eq:app_head_sensitivity_raw}
\end{align}
Here, $\|\cdot\|_F$ denotes the Frobenius norm. We then
normalize the mean over all $(l,g)$ pairs to one:
\begin{equation}
\omega_{l,g}
=
\frac{
\omega_{l,g}^{\mathrm{raw}}
}{
\operatorname{Mean}_{l',g'}
\omega_{l',g'}^{\mathrm{raw}}
+
\epsilon
}.
\label{eq:app_head_sensitivity}
\end{equation}
The resulting $\omega_{l,g}$ is a sensitivity multiplier shared
by all candidate operations associated with that KV head. It
does not change the tree structure or impose a separate
head-level quota.

\subsection{Utility-Guided Budget Flow}
\label{app:budget_flow}

The value flow aggregates information from tokens toward tree
roots, whereas the budget flow starts from uncovered roots and
moves through the frontiers toward the leaves. GraceKV first
applies a budget-aware exact-singleton floor and then allocates
the remaining entries through global competition shared by all
atomic units.

\paragraph{Stage I: budget-aware singleton floor.}
To prevent high-value content such as entity names, label
strings, and few-shot answer formats from being blurred by
prototypes, GraceKV first applies one-dimensional max pooling
with width $7$ to the token-value field. Let
\begin{equation}
\mathcal{N}_7(t)
=
\left\{
\tau\in\mathcal{C}
\;\middle|\;
|\tau-t|\leq 3
\right\}.
\end{equation}
The global average token value and the floor score after
cross-head consensus mixing are
\begin{align}
\overline{R}_t^{\mathrm{tok}}
&=
\operatorname{Mean}_{l,g}
R_{l,g,t}^{\mathrm{tok}},
\nonumber\\
f_{l,g,t}
&=
\max_{\tau\in\mathcal{N}_7(t)}
\left[
(1-c)
R_{l,g,\tau}^{\mathrm{tok}}
+
c
\overline{R}_{\tau}^{\mathrm{tok}}
\right],
\qquad
c=0.5.
\label{eq:app_floor_score}
\end{align}
Here, $c$ is the cross-head consensus weight. For every
$(l,g)$ pair, tokens are sorted in descending order of
$f_{l,g,t}$. Only tokens satisfying the quality gate
\begin{equation}
f_{l,g,t}
\geq
\frac{
\gamma_{\mathrm{floor}}
}{
N_c
},
\qquad
\gamma_{\mathrm{floor}}
=
1.2
\label{eq:app_floor_gate}
\end{equation}
are retained as candidates, and selection stops before their
cumulative original token value exceeds
$\eta_{\mathrm{floor}}=0.95$. Thus, heads with flat or weak
signals do not mechanically purchase large numbers of
singletons, while high-value positions in sharply concentrated
distributions can be preserved exactly.

Candidate singletons are not locked for free. For a candidate
triple $(l,g,t)$, the algorithm executes the root
\textsc{Add} operation and the necessary \textsc{Split}
operations along the tree containing $t$ until $\{t\}$ becomes
a frontier leaf. Every operation on this path is charged
according to the number of newly created physical entries. The
total floor cost satisfies
\begin{equation}
B_{\mathrm{floor}}
\leq
\left\lfloor
\alpha B_{\mathrm{base}}
\right\rfloor,
\qquad
B_{\mathrm{base}}
=
B_{\mathrm{total}}-LGr,
\qquad
\alpha=0.6.
\label{eq:app_floor_cap}
\end{equation}
Here, $B_{\mathrm{base}}$ is the base allocatable budget after
the recent window has been removed, and $\alpha$ is the global
hard cap on the floor fraction. If the total path cost of the
initial candidates from all heads exceeds this cap, the number
of candidates assigned to each head is reduced in proportion
to its original candidate count. A head with at least one token
passing Eq.~\eqref{eq:app_floor_gate} is still assigned at
least one candidate. The implementation additionally imposes
a per-head used-cost cap to prevent any single head from
exhausting the floor budget. A candidate is accepted only when
its complete path remains feasible under
Eq.~\eqref{eq:app_floor_cap}; the floor is therefore not
disabled for an entire layer when the budget becomes tight.

The complete set of physical entries reserved by the floor
stage is denoted by $\mathcal{E}_{\mathrm{floor}}$. Each
accepted candidate contributes an exact singleton anchor that
directly stores the original
$(k_t^{l,g},v_t^{l,g})$, uses representative position $t$,
and has effective multiplicity $1$, such that
$\log m^{\mathrm{eff}}=0$. Root and internal frontier nodes
created along an accepted floor path remain part of the
corresponding cache state and also belong to
$\mathcal{E}_{\mathrm{floor}}$. Entries shared by multiple
paths are included only once. Consequently,
$B_{\mathrm{floor}}=|\mathcal{E}_{\mathrm{floor}}|$, and
the free-competition budget $B_{\mathrm{free}}$ is then
computed according to Eq.~\eqref{eq:app_free_budget}.

\paragraph{Stage II: unified competition for coverage and resolution.}
Free competition contains only two primitive operations. If
slot $S_j$ is uncovered, \textsc{Add}$(l,g,j)$ introduces
root prototype $r_{l,g,j}$ to cover the entire slot. If
$v\in\mathcal{F}_{l,g,j}$ is an expandable frontier node,
\textsc{Split}$(l,g,v)$ replaces $v$ with its preselected
child set. Their marginal gains are
\begin{align}
\Delta_{\mathrm{add}}(l,g,j)
&=
\omega_{l,g}
R_{l,g}(r_{l,g,j})
\nonumber\\[-2pt]
&\quad\times
\left[
D^{\mathrm{drop}}
\left(
I_{r_{l,g,j}}
\right)
-
D
\left(
I_{r_{l,g,j}}
\right)
\right],
\nonumber\\
\Delta_{\mathrm{split}}(l,g,v)
&=
\omega_{l,g}
R_{l,g}(v)
\nonumber\\[-2pt]
&\quad\times
\left[
D(I_v)
-
\sum_{u\in\operatorname{child}(v)}
D(I_u)
\right].
\label{eq:app_operation_gains}
\end{align}
Equation~\eqref{eq:app_operation_gains} combines three sources
of information: $R_{l,g}$ represents input-conditioned
information value, the reduction in $D$ represents the
improvement in representation quality, and $\omega_{l,g}$
measures the output sensitivity of the layer--KV-head pair.
\textsc{Add} compares complete dropping with representation by
one root prototype and therefore purchases coverage.
\textsc{Split} compares a parent prototype with its child
prototype set and therefore purchases resolution.

The net entry costs of the two operations are
\begin{equation}
c_{\mathrm{add}}
=
1,
\qquad
c_{\mathrm{split}}(v)
=
\left|
\operatorname{child}(v)
\right|
-
1.
\label{eq:app_operation_costs}
\end{equation}
For any candidate operation $a$, let $\Delta(a)$ and $c(a)$
denote its marginal gain and net cost, respectively. Its
utility per budget unit is
\begin{equation}
U(a)
=
\frac{
\Delta(a)
}{
c(a)
}.
\label{eq:app_unit_utility}
\end{equation}
All currently feasible operations enter the same max-priority
queue, keyed by Eq.~\eqref{eq:app_unit_utility}. Feasibility
requires that: (i) all ancestor operations have already been
executed; (ii) the operation has not already been executed; and
(iii) its cost does not exceed the remaining budget.

Formally, let $\operatorname{Pred}(a)$ denote the prerequisite
operation set of action $a$, and let $\mathcal{A}$ denote the
set of all potential \textsc{Add} and \textsc{Split}
operations. The allocation objective is
\begin{align}
\max_{\mathcal{A}'\subseteq\mathcal{A}}
\quad
&
\sum_{a\in\mathcal{A}'}
\Delta(a),
\nonumber\\
\text{s.t.}
\quad
&
\sum_{a\in\mathcal{A}'}
c(a)
\leq
B_{\mathrm{free}},
\nonumber\\
&
\operatorname{Pred}(a)
\subseteq
\mathcal{A}',
\qquad
\forall a\in\mathcal{A}'.
\label{eq:app_allocation_problem}
\end{align}
The implemented allocator approximately solves
Eq.~\eqref{eq:app_allocation_problem} using greedy marginal
utility. At each step, it pops the highest-utility operation
whose cost fits the remaining budget. After execution, it
updates the corresponding tree frontier and inserts newly
exposed child operations whose prerequisites have been
satisfied. If the highest-utility operation is too expensive
for the remaining budget, it is skipped and the allocator
continues checking other affordable operations. Allocation
terminates when no feasible operation remains or when the
budget is exhausted. Since all layers, KV heads, and slots
share the same priority queue, GraceKV requires no predefined
layer-level budget, head-level budget, or slot-level
compression ratio.

\paragraph{Frontier updates and physical packing.}
After executing \textsc{Add}, an initially empty frontier
$\mathcal{F}_{l,g,j}$ becomes $\{r_{l,g,j}\}$. After executing
\textsc{Split}$(l,g,v)$, the frontier is updated as
\begin{equation}
\mathcal{F}_{l,g,j}
\leftarrow
\left(
\mathcal{F}_{l,g,j}
\setminus
\{v\}
\right)
\cup
\operatorname{child}(v).
\label{eq:app_frontier_update}
\end{equation}
A slot whose \textsc{Add} operation is never executed retains
an empty frontier and is physically dropped for that $(l,g)$
pair, producing no cache entry. The root represents the
coarsest coverage, internal frontier nodes correspond to local
merging, and single-token leaves correspond to exact
preservation. Token eviction, interval merging, and lossless
retention are therefore different frontier states within the
same tree rather than three mutually exclusive compression
rules.

For every $(l,g)$ pair, all frontier nodes, including those
reserved by the singleton-floor stage, and all recent-window
singletons are finally written into an independently sized
ragged cache. Every non-singleton
node stores
\begin{equation}
\left(
\widetilde{k}_v,
\widetilde{v}_v,
p_v,
\log m_v^{\mathrm{eff}}
\right),
\end{equation}
where the symbols are defined in
Eqs.~\eqref{eq:app_rope_prototype},
\eqref{eq:app_value_prototype}, and
\eqref{eq:app_effective_multiplicity}. Every singleton stores
its original K/V pair, its original position, and a zero
multiplicity bias. During decoding, attention is computed using
Eq.~\eqref{eq:app_decode_logit}, and every newly generated
token is appended to the corresponding ragged cache with
effective multiplicity $1$.

When $x=1$, Eq.~\eqref{eq:app_total_budget} provides
$LGN_{\mathrm{ctx}}$ contextual entries. All trees can be
expanded to single-token leaves, and the resulting cache
matches FullKV entry by entry. Under a restricted budget,
roots that are never selected are physically dropped,
selected but incompletely expanded intervals are covered by
prototypes, and high-value regions receive higher resolution
through additional splits or the singleton floor. GraceKV
therefore jointly determines information coverage and local
representation resolution under one shared physical cache
budget.


\section{Further Analysis of Curvature-Guided Slot Segmentation}
\label{app:curvature_segmentation}

\subsection{Motivation and Novelty}
\label{app:curvature_motivation}

GraceKV first partitions the long context into a set of
contiguous slots, based on which it constructs a progressively
refinable prototype tree for every layer--KV-head--slot atomic
unit. This partition determines not only the intervals covered
by the tree roots, but also which tokens may initially share a
low-resolution representation. The slot boundaries should
therefore be placed near structural transitions in the contextual
representation, rather than being determined solely by a
predefined uniform length.

Uniform segmentation is simple, but it may separate regions
whose representations evolve continuously or place
representationally heterogeneous neighboring regions in the
same slot. The former produces unnecessary tree roots and
cache entries, whereas the latter requires a coarse prototype
to approximate multiple incompatible representation patterns,
increasing the distortion introduced by KV merging. Explicit
punctuation and document delimiters provide another possible
partition signal, but they depend strongly on the input format
and do not directly indicate whether the model's internal
representation has changed. Token surprisal primarily reflects
local prediction difficulty or rarity and likewise does not
necessarily correspond to a structural boundary.

We instead view the hidden states of consecutive tokens as a
discrete trajectory in the representation space and identify
candidate boundaries from changes in its direction. Intuitively,
when neighboring hidden states evolve smoothly along similar
directions, they are more likely to admit a shared
low-resolution prototype. When the trajectory changes direction
sharply, merging across that position is more likely to mix
different representation structures. GraceKV therefore uses
multi-layer, multi-scale hidden-state curvature as its primary
structural signal and token surprisal only as an auxiliary signal
for local information transitions.

To the best of our knowledge, GraceKV is the first KV cache
compression method to use multi-layer, multi-scale hidden-state
trajectory curvature to construct contiguous textual slots for
multi-resolution KV cache compression. This claim concerns the
specific structural role of curvature in constructing the shared
slot partition, rather than the broader use of curvature for
language-model analysis or token compression.

GraceKV constructs a single context partition shared by all
layers and KV heads:
\begin{equation}
\mathcal{S}
=
\left\{
S_1,S_2,\ldots,S_J
\right\}.
\label{eq:app_shared_partition}
\end{equation}
After obtaining these shared slot boundaries, GraceKV
independently constructs a prototype tree
$\mathcal{T}_{l,g,j}$ for every layer $l$, KV head $g$, and
slot $S_j$. This design separates the global structural
partition of the context from the head-specific cache
representation. All atomic units share the same contiguous
regions, while their prototype trees and final representation
resolutions remain specific to their own K/V states, estimated
distortions, and task-conditioned values.

\subsection{Multi-Layer, Multi-Scale Curvature}
\label{app:multiscale_curvature}

\paragraph{Hidden-state trajectory.}
Consider an input sequence of length $T$ processed by a
Transformer with $L$ layers. During the standard prefill pass,
GraceKV collects the input residual stream of several sampled
Transformer blocks. For sampled layer $l$ and token position
$i$, let
\begin{equation}
h_i^{(l)} \in \mathbb{R}^{D}
\end{equation}
denote the residual hidden state before the block-level input
normalization and self-attention computation, where $D$ is the
hidden dimension.

The sampled layer set is defined by relative model depths:
\begin{equation}
\begin{aligned}
\mathcal{L}_{\mathrm{seg}}
=
\operatorname{Unique}\Bigl(
\bigl\{
&\min\!\left(
\left\lfloor \rho L \right\rfloor,
L-1
\right)
\\
&\mid\;
\rho\in\{0.33,0.50,0.67\}
\bigr\}
\Bigr).
\end{aligned}
\label{eq:app_seg_layers}
\end{equation}
The same rule is applied to all supported backbones, allowing
the segmentation procedure to adapt automatically to different
numbers of Transformer layers.

We first apply L2 normalization:
\begin{equation}
\bar{h}_i^{(l)}
=
\frac{
h_i^{(l)}
}{
\max
\left(
\left\|h_i^{(l)}\right\|_2,
\varepsilon
\right)
},
\qquad
\varepsilon=10^{-6}.
\label{eq:app_hidden_normalization}
\end{equation}
This normalization suppresses variations in hidden-state
magnitude and makes the curvature primarily reflect changes in
trajectory direction.

\paragraph{Single-layer, single-scale curvature.}
For sampled layer $l$, position $i$, and window size $w$, we
compute the local means on the two sides of position $i$:
\begin{equation}
\mu_{i,-}^{(l,w)}
=
\frac{1}{w}
\sum_{\tau=0}^{w-1}
\bar{h}_{i-\tau}^{(l)},
\qquad
\mu_{i,+}^{(l,w)}
=
\frac{1}{w}
\sum_{\tau=1}^{w}
\bar{h}_{i+\tau}^{(l)}.
\label{eq:app_local_means}
\end{equation}
The left window includes the current position $i$, whereas the
right window starts from $i+1$. We then define the incoming
and outgoing local directions as
\begin{equation}
\delta_{i,-}^{(l,w)}
=
\bar{h}_i^{(l)}
-
\mu_{i,-}^{(l,w)},
\qquad
\delta_{i,+}^{(l,w)}
=
\mu_{i,+}^{(l,w)}
-
\bar{h}_i^{(l)}.
\label{eq:app_local_directions}
\end{equation}
Their normalized forms are
\begin{equation}
u_{i,-}^{(l,w)}
=
\frac{
\delta_{i,-}^{(l,w)}
}{
\max
\left(
\left\|\delta_{i,-}^{(l,w)}\right\|_2,
\varepsilon
\right)
},
\label{eq:app_incoming_direction}
\end{equation}
and
\begin{equation}
u_{i,+}^{(l,w)}
=
\frac{
\delta_{i,+}^{(l,w)}
}{
\max
\left(
\left\|\delta_{i,+}^{(l,w)}\right\|_2,
\varepsilon
\right)
}.
\label{eq:app_outgoing_direction}
\end{equation}

The discrete trajectory curvature at position $i$ is defined as
\begin{equation}
\kappa_i^{(l,w)}
=
1
-
\left\langle
u_{i,-}^{(l,w)},
u_{i,+}^{(l,w)}
\right\rangle.
\label{eq:app_local_curvature}
\end{equation}
When the incoming and outgoing directions are similar, their
inner product approaches one and the curvature approaches zero,
indicating a locally smooth trajectory. A larger curvature
indicates a stronger directional transition.

The default set of window scales is
\begin{equation}
\mathcal{W}=\{4,8,16\}.
\label{eq:app_curvature_windows}
\end{equation}
For scale $w$, curvature is evaluated only at positions
satisfying
\begin{equation}
w \le i < T-w.
\end{equation}
At positions that do not admit complete windows on both sides,
the curvature at that scale is set to zero rather than using
padding or shortened windows. If $T<2w+1$, the curvature for
that scale is zero at all positions.

\paragraph{Multi-layer, multi-scale stabilization.}
Curvature from a single layer or window scale can be sensitive
to local noise. GraceKV therefore aggregates curvature over
both representation depths and neighborhood sizes. Define
\begin{equation}
\mathcal{K}_i
=
\left\{
\kappa_i^{(l,w)}
\;\middle|\;
l\in\mathcal{L}_{\mathrm{seg}},
\;
w\in\mathcal{W}
\right\}.
\end{equation}
The stabilized curvature is
\begin{equation}
\kappa_i
=
\operatorname{Mean}
\left(
\mathcal{K}_i
\right)
-
\lambda_{\mathrm{stab}}
\operatorname{Std}
\left(
\mathcal{K}_i
\right),
\label{eq:app_stabilized_curvature}
\end{equation}
where the standard deviation is computed as a population
standard deviation and
\begin{equation}
\lambda_{\mathrm{stab}}=0.25
\end{equation}
by default.

The mean term preserves transitions that are consistently
visible across layers and scales, while the standard-deviation
penalty suppresses peaks that appear only at a particular layer
or neighborhood size. Consequently, a position receives a
large final curvature only when it represents a relatively
stable transition across multiple representation depths and
local contexts.

\paragraph{Auxiliary surprisal signal.}
Curvature measures directional changes in the internal
representation, but some local information transitions may not
produce equally strong geometric changes at all sampled layers.
We therefore incorporate token surprisal as an auxiliary signal.
For token $x_i$, its surprisal is
\begin{equation}
s_i^{\mathrm{surp}}
=
-\log p
\left(
x_i
\mid
x_{<i}
\right).
\label{eq:app_surprisal}
\end{equation}
In the implementation, $s_i^{\mathrm{surp}}$ is computed from
the logits at position $i-1$ evaluated on token $x_i$. The first
sequence position has no preceding prediction and is assigned
\begin{equation}
s_0^{\mathrm{surp}}=0.
\end{equation}

Because curvature and surprisal have different numerical
scales, they are normalized separately. For a sequence-level
signal
\begin{equation}
a=(a_0,a_1,\ldots,a_{T-1}),
\end{equation}
we define robust normalization as
\begin{equation}
z_{\mathrm{robust}}(a_i)
=
\frac{
a_i-\operatorname{median}(a)
}{
1.4826\operatorname{MAD}(a)+\varepsilon
},
\label{eq:app_robust_z}
\end{equation}
where
\begin{equation}
\operatorname{MAD}(a)
=
\operatorname{median}
\left(
\left|
a-\operatorname{median}(a)
\right|
\right),
\qquad
\varepsilon=10^{-6}.
\label{eq:app_mad}
\end{equation}
The median and median absolute deviation are computed over the
complete input sequence. No clipping is applied.

The final boundary score is
\begin{equation}
b_i
=
z_{\mathrm{robust}}
\left(
\kappa_i
\right)
+
\lambda_s
z_{\mathrm{robust}}
\left(
s_i^{\mathrm{surp}}
\right),
\label{eq:app_boundary_score}
\end{equation}
with
\begin{equation}
\lambda_s=0.15
\end{equation}
by default. Curvature therefore remains the primary structural
signal, while surprisal provides a smaller auxiliary correction
for local information changes. The default implementation does
not add rewards for punctuation, line breaks, passage
separators, or other explicit delimiters, avoiding dependence on
a particular document format or prompt template.

\subsection{Dynamic-Programming Segmentation Objective}
\label{app:segmentation_objective}

\paragraph{Segmentation domain.}
Let the context occupy the interval
\begin{equation}
[c_s,c_e)
\end{equation}
in the complete input sequence. GraceKV preserves a recent
window of length $r$ at the end of the context, and therefore
performs segmentation only over the compressible interval
\begin{equation}
[c_s,c_{\mathrm{end}}),
\qquad
c_{\mathrm{end}}=c_e-r.
\label{eq:app_compressible_interval}
\end{equation}
The prefix, recent window, question, and template suffix are
excluded from the dynamic-programming domain and form hard
boundaries.

Let
\begin{equation}
N=c_{\mathrm{end}}-c_s
\end{equation}
be the length of the compressible region. We use coordinates
relative to this region below. A slot is a half-open interval
\begin{equation}
S=[a,e),
\qquad
0\le a<e\le N,
\end{equation}
with length
\begin{equation}
|S|=e-a.
\end{equation}
The complete partition is
\begin{equation}
\mathcal{S}
=
\left\{
S_j=[a_j,a_{j+1})
\right\}_{j=1}^{J},
\label{eq:app_dp_slot_partition}
\end{equation}
where
\begin{equation}
a_1=0,
\qquad
a_{J+1}=N.
\end{equation}

\paragraph{Within-slot representation dispersion.}
A slot suitable for coarse prototype representation should not
only avoid crossing strong curvature boundaries, but should
also exhibit a relatively consistent internal trajectory. We
therefore measure the dispersion of adjacent hidden-state
differences within each slot.

For sampled layer $l$, define adjacent differences over the
compressible region as
\begin{equation}
d_t^{(l)}
=
\bar{h}_{c_s+t+1}^{(l)}
-
\bar{h}_{c_s+t}^{(l)},
\qquad
t=0,\ldots,N-2.
\label{eq:app_adjacent_difference}
\end{equation}
For slot $S=[a,e)$, the number of adjacent differences is
\begin{equation}
n_S
=
|S|-1
=
e-a-1.
\end{equation}
When $n_S\ge 1$, their mean is
\begin{equation}
\bar{d}_S^{(l)}
=
\frac{1}{n_S}
\sum_{t=a}^{e-2}
d_t^{(l)}.
\label{eq:app_mean_difference}
\end{equation}
For $n_S\ge 2$, the layer-specific dispersion is defined as the
sample variance of these difference vectors:
\begin{equation}
\operatorname{Disp}_l(S)
=
\frac{1}{n_S-1}
\sum_{t=a}^{e-2}
\left\|
d_t^{(l)}
-
\bar{d}_S^{(l)}
\right\|_2^2.
\label{eq:app_layer_dispersion}
\end{equation}
For $n_S<2$, we define
\begin{equation}
\operatorname{Disp}_l(S)=0.
\end{equation}
The final slot dispersion is averaged across the sampled
layers:
\begin{equation}
\operatorname{Disp}(S)
=
\frac{1}{
\left|
\mathcal{L}_{\mathrm{seg}}
\right|
}
\sum_{l\in\mathcal{L}_{\mathrm{seg}}}
\operatorname{Disp}_l(S).
\label{eq:app_slot_dispersion}
\end{equation}

A low value of $\operatorname{Disp}(S)$ indicates that the
hidden-state trajectory evolves in a relatively consistent
direction within the slot, making the interval more suitable
for a shared coarse representation. A large value indicates
that the interval may contain multiple representation patterns,
even when it does not contain a single dominant curvature
peak.

For efficient interval queries, the implementation computes
the equivalent form
\begin{equation}
\operatorname{Disp}_l(S)
=
\frac{
\displaystyle
\sum_{t=a}^{e-2}
\left\|
d_t^{(l)}
\right\|_2^2
-
\frac{1}{n_S}
\left\|
\sum_{t=a}^{e-2}
d_t^{(l)}
\right\|_2^2
}{
\max(n_S-1,1)
}.
\label{eq:app_prefix_dispersion}
\end{equation}
Prefix statistics of
$\sum_t d_t^{(l)}$ and
$\sum_t\|d_t^{(l)}\|_2^2$
allow the dispersion of any candidate slot to be queried in
constant time during dynamic programming.

\paragraph{Global segmentation objective.}
The boundary score encourages cuts at representation
transitions, whereas the dispersion term discourages slots
whose internal trajectories are inconsistent. Optimizing only
these terms, however, may create excessive numbers of very
short slots. We therefore additionally introduce a preferred
slot length and a fixed cost for creating a new boundary.

Let $n_0$ denote the target slot length and define the length
penalty
\begin{equation}
\Phi(S)
=
\left(
\frac{|S|-n_0}{n_0}
\right)^2.
\label{eq:app_length_penalty}
\end{equation}
For every non-initial slot beginning at relative position $a_j$,
the corresponding cut is placed after complete-sequence
position $c_s+a_j-1$. It receives boundary reward
$b_{c_s+a_j-1}$ and pays a fixed cut cost
$c_{\mathrm{cut}}$.

The complete segmentation objective is
\begin{equation}
\begin{aligned}
\max_{\mathcal{S}}
\quad
&
\sum_{j=2}^{J}
\left(
b_{c_s+a_j-1}
-
c_{\mathrm{cut}}
\right)
\\
&
-
\lambda_d
\sum_{j=1}^{J}
\operatorname{Disp}(S_j)
\\
&
-
\lambda_n
\sum_{j=1}^{J}
\left(
\frac{|S_j|-n_0}{n_0}
\right)^2,
\end{aligned}
\label{eq:app_segmentation_objective}
\end{equation}
subject, under normal conditions, to
\begin{equation}
n_{\min}
\le
|S_j|
\le
n_{\max}.
\label{eq:app_slot_constraints}
\end{equation}
The default values are
\begin{equation}
n_{\min}=16,
\qquad
n_{\max}=128,
\qquad
n_0=64,
\label{eq:app_slot_length_defaults}
\end{equation}
and
\begin{equation}
\lambda_d=0.25,
\qquad
\lambda_n=0.10,
\qquad
c_{\mathrm{cut}}=1.0.
\label{eq:app_segmentation_defaults}
\end{equation}

The first term favors boundaries supported by curvature and
surprisal. The second term favors slots whose internal
representation trajectories are consistent. The third term
prevents slot lengths from deviating excessively from the
preferred scale, while the fixed cut cost suppresses
over-segmentation caused by noisy local peaks.

\paragraph{Dynamic-programming recurrence.}
Let $F[i]$ denote the maximum objective value for partitioning
the relative interval $[0,i)$. We initialize
\begin{equation}
F[0]=0.
\end{equation}
Define the reward for starting a new slot at relative position
$a$ as
\begin{equation}
\Gamma(a)
=
\begin{cases}
0,
&
a=0,
\\[4pt]
b_{c_s+a-1}
-
c_{\mathrm{cut}},
&
a>0.
\end{cases}
\label{eq:app_boundary_reward}
\end{equation}

For endpoint $i$, the standard predecessor set is
\begin{equation}
\mathcal{A}(i)
=
\left\{
a
\;\middle|\;
0\le a<i,
\;
n_{\min}
\le
i-a
\le
n_{\max}
\right\}.
\label{eq:app_predecessor_set}
\end{equation}
The recurrence is
\begin{equation}
\begin{aligned}
F[i]
=
\max_{a\in\mathcal{A}(i)}
\Bigg[
&
F[a]
+
\Gamma(a)
-
\lambda_d
\operatorname{Disp}([a,i))
\\
&
-
\lambda_n
\left(
\frac{i-a-n_0}{n_0}
\right)^2
\Bigg].
\end{aligned}
\label{eq:app_dp_recurrence}
\end{equation}
The maximizing predecessor is recorded as
\begin{equation}
\begin{aligned}
P[i]
=
\arg\max_{a\in\mathcal{A}(i)}
\Bigg[
&
F[a]
+
\Gamma(a)
-
\lambda_d
\operatorname{Disp}([a,i))
\\
&
-
\lambda_n
\left(
\frac{i-a-n_0}{n_0}
\right)^2
\Bigg].
\end{aligned}
\label{eq:app_dp_backpointer}
\end{equation}
Starting from $i=N$ and following the backpointers $P[i]$
recovers the complete shared slot partition.

The first slot begins at zero and therefore receives neither a
boundary reward nor a cut penalty. The final slot only needs to
reach $N$ and receives no additional terminal reward. If no
standard predecessor exists for $i=N$, the implementation
relaxes only the minimum-length constraint of the final slot:
\begin{equation}
\mathcal{A}_{\mathrm{term}}(N)
=
\left\{
a
\;\middle|\;
0\le a<N,
\;
1\le N-a\le n_{\max}
\right\}.
\label{eq:app_terminal_predecessor}
\end{equation}
This guarantees complete coverage when the remaining suffix is
shorter than $n_{\min}$. When the entire compressible region is
shorter than $n_{\min}$, it is represented as a single slot.

Each endpoint considers at most
$n_{\max}-n_{\min}+1$ standard predecessors, and each interval
dispersion is queried in constant time. The resulting
time complexity is
\begin{equation}
O(Nn_{\max}),
\end{equation}
and the dynamic-programming state and backpointers require
\begin{equation}
O(N)
\end{equation}
additional memory. With the default $n_{\max}=128$, the
search bandwidth remains fixed as the context length grows.

\subsection{Why Curvature for Slot Boundaries}
\label{app:why_curvature}

Curvature and token importance serve fundamentally different
roles in GraceKV. The subsequent value flow estimates whether
a region is useful for the current task, whereas
curvature-guided segmentation determines which neighboring
tokens should not be forced to share an initially coarse
representation. The former answers whether a region is worth
preserving; the latter identifies where a low-resolution
representation should be structurally separated. We therefore
do not directly use attention scores, token relevance, or
multi-hop values to determine the shared slot partition, as
doing so would make the structural decomposition itself depend
entirely on the current question's local attention pattern.

Curvature is particularly suitable for this purpose because it
measures a change in the direction of representation evolution,
rather than only the distance between neighboring states.
Hidden states may move by a relatively large distance while
continuing along a consistent direction, in which case the
underlying representation pattern remains smooth. Conversely,
a sequence of individually small changes may still contain a
strong structural transition when their directions change
sharply. Segmentation based only on adjacent embedding
distance cannot distinguish these two cases as directly.

Curvature and surprisal also provide complementary information.
Surprisal measures the difficulty of predicting the current
token. Rare entities, numbers, special symbols, or unusual
surface forms may all produce high surprisal without defining a
structural boundary. Curvature instead compares the directions
of the hidden-state trajectory on the two sides of a position
and is therefore more directly related to a transition between
local representation patterns. GraceKV consequently assigns
surprisal a relatively small auxiliary weight rather than
allowing it to dominate the partition.

The multi-layer, multi-scale formulation further prevents the
partition from depending on a single representation depth or
neighborhood size. Small windows respond to local transitions,
whereas larger windows capture broader structural changes.
Intermediate and deeper layers expose representation changes at
different abstraction levels. Subtracting the cross-layer,
cross-scale standard deviation suppresses peaks that are not
stable across these views.

The dynamic-programming objective also does not mechanically
convert every curvature peak into a boundary. A candidate cut
must jointly compete with within-slot dispersion, preferred
length, and the fixed cost of introducing another slot. A local
peak is selected only when its structural benefit is sufficient
to compensate for the additional segmentation cost. The final
partition therefore reflects a global trade-off between local
boundary evidence and overall slot quality, rather than a
simple curvature-thresholding rule.

The same stabilized curvature signal $\kappa$ is additionally
used to propose structural split positions within each
prototype tree. For an interval $I$, GraceKV constructs a set
of binary-split candidates from the highest-curvature positions
and the interval midpoint:
\begin{equation}
\mathcal{C}(I)
=
\operatorname{TopK}_{t\in I}
\left(
\kappa_t
\right)
\cup
\left\{
\operatorname{mid}(I)
\right\},
\label{eq:app_tree_split_candidates}
\end{equation}
where the default number of curvature candidates is
\begin{equation}
K=16.
\end{equation}
When comparing binary splits, curvature is min--max normalized
within the current interval and included as an auxiliary reward
with default weight $0.05$ in addition to distortion reduction.

Importantly, tree refinement uses the stabilized curvature
$\kappa$, rather than the boundary score $b$ that combines
curvature with surprisal. Curvature thus serves two consistent
but distinct structural roles: it first determines the shared
macroscopic slot partition and then biases finer-grained
refinement within each per-head prototype tree.

In the ablation denoted by \emph{w/o Curv.\ Segmentation}, we
replace the complete curvature-guided segmentation procedure
with uniform slots of target length 64. We also remove
curvature-guided tree refinement by disabling curvature
candidates and curvature rewards, leaving only the interval
midpoint as the binary-split candidate. This variant is
therefore not a surprisal-only dynamic-programming partition.
It removes both curvature-guided slot construction and the
curvature bias used during subsequent tree refinement. Its
performance degradation across different backbones, task
types, and cache budgets supports the role of curvature in both
macroscopic structural partitioning and fine-grained prototype
refinement.



\section{Global Allocation, Greedy Approximation, and the Singleton Floor}
\label{app:global_allocation}

This section further clarifies the meaning of ``global allocation'' in GraceKV, the computational difficulty of exact global optimization, the approximation induced by marginal-utility greedy allocation, and the role of the singleton floor in compensating for this approximation. In the main paper, ``global'' refers to the \emph{scope of resource competition}: all layer--KV-head--slot atomic units share the same cache budget and compete within a unified decision space. It does not imply that GraceKV guarantees a globally optimal solution in the combinatorial-optimization sense. GraceKV instead uses a global priority queue to efficiently approximate this combinatorial allocation problem and introduces a budget-aware singleton floor to protect exact information that may otherwise be underestimated by step-wise greedy decisions.

\subsection{Global Allocation vs.\ Global Optimality}
\label{app:allocation_vs_optimality}

Let
\[
\mathcal{S}=\{S_j\}_{j=1}^{J}
\]
denote the shared slot partition produced by curvature-guided segmentation. For every layer--KV-head--slot tuple $(l,g,j)$, GraceKV constructs a prototype tree
\[
\mathcal{T}_{l,g,j},
\]
whose current cache representation is described by a frontier
\[
\mathcal{F}_{l,g,j}.
\]

An empty frontier means that the entire slot is dropped at the corresponding layer and KV head. A frontier containing only the root represents the complete slot using one coarse prototype. Expanding frontier nodes progressively improves local resolution, while full expansion to token-level leaves recovers the original K/V entries.

For an uncovered slot, GraceKV defines an Add operation
\[
\operatorname{Add}(l,g,j),
\]
which inserts the root $r_{l,g,j}$ into the frontier. For an expandable frontier node $v$, GraceKV defines a Split operation
\[
\operatorname{Split}(l,g,v),
\]
which replaces $v$ with its precomputed child set $\operatorname{child}(v)$. Their physical cache-entry costs are
\[
c_{\mathrm{add}}=1,
\qquad
c_{\mathrm{split}}(v)
=
|\operatorname{child}(v)|-1.
\]

Let $R_{l,g}(v)$ denote the input-conditioned value of node $v$, let $\omega_{l,g}$ denote the sensitivity of the corresponding layer--KV-head pair, and let $D(I_v)$ and $D_{\mathrm{drop}}(I_v)$ denote the prototype distortion and complete-drop cost of interval $I_v$, respectively. The marginal gains of Add and Split are
\[
\Delta_{\mathrm{add}}(l,g,j)
=
\omega_{l,g}R_{l,g}(r_{l,g,j})
\left[
D_{\mathrm{drop}}(I_{r_{l,g,j}})
-
D(I_{r_{l,g,j}})
\right],
\]
and
\[
\Delta_{\mathrm{split}}(l,g,v)
=
\omega_{l,g}R_{l,g}(v)
\left[
D(I_v)
-
\sum_{u\in\operatorname{child}(v)}
D(I_u)
\right].
\]

Accordingly, Add measures the gain obtained by changing a region from complete removal to coarse coverage, whereas Split measures the gain obtained by replacing one coarse prototype with a finer set of child prototypes. Because the two operations use the same value, distortion, sensitivity, and physical-entry units, coverage and resolution can directly compete under one shared budget.

Let $\mathcal{A}$ denote the set of all potential Add and Split operations, and let $\operatorname{Pred}(a)$ denote the prerequisite operations of action $a$. For example, a Split operation can be executed only after the corresponding root has been admitted and all required ancestor expansions have been completed. After deducting the recent window and the realized singleton-floor cost, let $B_{\mathrm{free}}$ denote the budget available to the main auction. The ideal combinatorial allocation problem can be written as
\[
\begin{aligned}
\max_{\mathcal{A}'\subseteq\mathcal{A}}
\quad&
\sum_{a\in\mathcal{A}'}
\Delta(a)
\\
\mathrm{s.t.}\quad&
\sum_{a\in\mathcal{A}'}
c(a)
\leq
B_{\mathrm{free}},
\\
&
\operatorname{Pred}(a)
\subseteq
\mathcal{A}',
\qquad
\forall a\in\mathcal{A}'.
\end{aligned}
\]

In this formulation, \emph{global allocation} means that:

\begin{itemize}
    \item candidate operations from all Transformer layers belong to the same competition space;
    \item all KV heads share the same remaining cache budget;
    \item coverage and refinement operations from different context slots can be compared directly;
    \item Add and Split use the same physical-entry budget despite representing different compression behaviors; and
    \item the main auction does not impose preset layer-wise, KV-head-wise, or slot-wise quotas.
\end{itemize}

Consequently, the number of physical cache entries eventually assigned to each layer, KV head, or context slot is an \emph{emergent result} of global competition among its candidate operations, rather than a pre-specified compression ratio or local quota.

Global optimality is a stronger and fundamentally different requirement. It would require searching over all valid frontier combinations of all prototype trees and identifying the joint representation with the maximum total utility under the shared budget. GraceKV does not claim such combinatorial optimality. In other words, \emph{global} describes the scope of resource competition, whereas \emph{optimal} describes an exact solution guarantee over the complete discrete search space. GraceKV implements the former and efficiently approximates the latter.

The \texttt{w/o Global Allocation} ablation replaces the freely shared budget with fixed layer--KV-head-level capacity limits. The prototype trees and utility definitions remain unchanged, but once a layer--KV-head pair reaches its local limit, it cannot consume budget left unused by another pair. This prevents cache resources from flowing toward higher-utility operations and recovers the ``first impose local quotas, then make compression decisions'' paradigm discussed in the main paper.

\subsection{Difficulty of Exact Global Optimization}
\label{app:exact_optimization}

The difficulty of exact optimization does not arise only from the number of candidate operations. Each prototype tree contains a combinatorial number of valid frontiers under ancestor constraints, while the choices made across different trees are coupled by the shared cache budget.

For a prototype tree $\mathcal{T}$, let
\[
\mathfrak{F}(\mathcal{T})
\]
denote the set of all valid frontiers of that tree. For any node $v$, a valid representation of its subtree has two possibilities:

\begin{enumerate}
    \item retain $v$ itself as a frontier node; or
    \item expand $v$ and independently select a valid frontier from every child subtree.
\end{enumerate}

Let $N(v)$ denote the number of valid frontiers in the subtree rooted at $v$. A leaf node satisfies
\[
N(v)=1,
\qquad
v\ \text{is a leaf},
\]
whereas a non-leaf node satisfies
\[
N(v)
=
1+
\prod_{u\in\operatorname{child}(v)}
N(u).
\]

The first term corresponds to retaining the parent node, while the second term corresponds to expanding the parent and jointly selecting frontiers from all child subtrees. Even for a balanced binary tree, this quantity grows rapidly with tree depth. Salient-token extraction may additionally create three-way branches, further enlarging the valid-frontier space.

Let the complete set of atomic prototype trees be
\[
\mathbb{T}
=
\left\{
\mathcal{T}_{l,g,j}
:
1\leq l\leq L,\,
1\leq g\leq G,\,
1\leq j\leq J
\right\}.
\]

Explicit enumeration of all joint cache representations would require considering a search space of size
\[
\prod_{\mathcal{T}\in\mathbb{T}}
|\mathfrak{F}(\mathcal{T})|.
\]

The number of slots $J$ grows approximately linearly with context length, and every slot is instantiated across $L$ layers and $G$ KV heads. Therefore, enumerating all joint frontiers is infeasible for long-context inference.

Even if every tree were first reduced to a set of cost--utility states, exact global allocation would still require jointly selecting one valid state from every tree under a shared budget. The problem simultaneously contains:

\begin{itemize}
    \item discrete physical-entry costs;
    \item tree-structured prerequisite constraints;
    \item distinct coverage and refinement actions;
    \item shared-budget coupling across atomic units; and
    \item high-value leaves that may be reachable only after multiple intermediate expansions.
\end{itemize}

An exact dynamic program would generally need to include the cache budget as an additional state dimension and repeatedly merge the non-dominated cost--utility states of different trees. Its cost depends not only on the numerical budget but also on the number of non-dominated frontiers retained for every subtree. In online long-context inference, such optimization overhead can become comparable to, or even outweigh, the decoding savings obtained from cache compression.

GraceKV therefore does not attempt to solve the exact combinatorial problem online. Instead, it uses a global greedy allocator based on marginal utility per physical entry. This converts an intractable joint-frontier search into a sequence of priority-queue operations while preserving global resource competition across layers, KV heads, and context regions.

\subsection{Greedy Marginal-Utility Allocation}
\label{app:greedy_allocation}

For each candidate action $a$, GraceKV defines its marginal utility per physical cache entry as
\[
U(a)
=
\frac{\Delta(a)}{c(a)}.
\]

Normalizing by cost is important because different expansions consume different numbers of entries. A binary Split usually adds one net cache entry, whereas an internal salient-token extraction may add two. Ranking actions only by their unnormalized gains would systematically favor expensive operations merely because they purchase multiple entries at once.

After completing the singleton-floor stage, GraceKV initializes a global max-priority queue according to the current prototype-tree states:

\begin{itemize}
    \item for every uncovered slot, its root Add operation is inserted into the queue;
    \item for every non-leaf frontier node exposed by the singleton-floor stage, its Split operation is inserted;
    \item Add and Split operations already executed during floor construction are not inserted again; and
    \item a deeper Split becomes available only after its parent has become part of the current frontier.
\end{itemize}

At every iteration, the allocator performs the following procedure:

\begin{enumerate}
    \item pop the operation with the highest unit-cost utility;
    \item verify that its associated node is still part of the current frontier;
    \item verify that its prerequisite operations remain satisfied;
    \item check whether its physical cost fits within the remaining budget;
    \item execute the operation and update the corresponding frontier if it is feasible;
    \item insert newly exposed child Split operations into the queue; and
    \item repeat until the budget is exhausted or no executable operation remains.
\end{enumerate}

If the currently highest-ranked operation costs more than the remaining budget, the allocator does not terminate immediately. It discards that infeasible queue item and continues searching for a lower-cost action. Therefore, when a cost-two extraction cannot be purchased, the final remaining budget unit may still be spent on a cost-one Add or binary Split.

The value, distortion, and gain of every node are precomputed during tree construction. An operation in one tree does not change the distortion or token value of another tree. Different trees are coupled only through the shared budget, while operations within the same tree are coupled through frontier and ancestor constraints. This enables incremental allocation: an operation enters the queue only after its parent action is completed, and stale actions are ignored when they are popped if their nodes are no longer part of the current frontier.

Let $|\mathcal{A}|$ denote the number of potential operations. Since each effective operation is inserted into and removed from the heap only a constant number of times, the priority-queue complexity is
\[
O\!\left(
|\mathcal{A}|
\log |\mathcal{A}|
\right),
\]
which is substantially smaller than explicitly enumerating all joint frontiers.

Although the selection process is greedy, it is still \emph{global}: each step compares the currently available actions across all layers, KV heads, and context slots, rather than performing independent greedy allocation within each local unit.

Nevertheless, marginal-utility greedy allocation cannot guarantee combinatorial global optimality. One important reason is that the gain of reaching a high-value leaf may be non-smooth along its prerequisite path. Suppose a high-value token $t$ lies on the path
\[
r=v_0
\rightarrow
v_1
\rightarrow
\cdots
\rightarrow
v_d=\{t\}.
\]

Exactly preserving $t$ may require one root Add followed by multiple Split or Extract operations. Although the final singleton leaf may provide a large cumulative benefit, some intermediate operations may have relatively small immediate gains. A step-wise greedy allocator observes only currently exposed actions and may spend the available budget on alternatives with higher immediate utility before reaching $\{t\}$. The singleton floor is introduced specifically to compensate for this type of non-smooth gain.

\subsection{Singleton-Floor Construction and Budget Accounting}
\label{app:floor_construction}

The singleton floor is not a set of free cache entries stored outside the prototype trees. It is a budget-constrained exact-representation channel that forces a small number of high-value layer--KV-head--token tuples to become token-level leaves before the main global auction. Its purpose is to reduce the risk that entity names, numerical values, label strings, or few-shot output formats are blurred by coarse prototypes.

\paragraph{Candidate scoring.}
Let
\[
R^{\mathrm{tok}}_{l,g,t}
\]
denote the value of token $t$ at layer $l$ and KV head $g$. GraceKV first computes the consensus value across all layer--KV-head pairs:
\[
\bar{R}^{\mathrm{tok}}_t
=
\operatorname{Mean}_{l,g}
R^{\mathrm{tok}}_{l,g,t}.
\]

It then mixes the local value with the global consensus and applies width-seven max-pooling:
\[
f_{l,g,t}
=
\max_{\tau\in\mathcal{N}_7(t)}
\left[
(1-c)
R^{\mathrm{tok}}_{l,g,\tau}
+
c\bar{R}^{\mathrm{tok}}_{\tau}
\right],
\qquad
c=0.5,
\]
where
\[
\mathcal{N}_7(t)
=
\left\{
\tau\in\mathcal{C}
:
|\tau-t|\leq 3
\right\}.
\]

The consensus term encourages important positions to receive compatible protection across multiple layers and KV heads. Max-pooling allows a high-value token to protect nearby subword pieces and reduces the risk that only part of a multi-token entity is isolated.

For each $(l,g)$, tokens are ranked by $f_{l,g,t}$ and must satisfy the quality gate
\[
f_{l,g,t}
\geq
\frac{\gamma_{\mathrm{floor}}}{N_c},
\qquad
\gamma_{\mathrm{floor}}=1.2.
\]

Candidates are then selected in descending order until their cumulative original token value reaches
\[
\eta_{\mathrm{floor}}=0.95.
\]

This procedure produces an initial candidate set
\[
\mathcal{C}_{\mathrm{floor}}.
\]

The same textual position appearing in different layer--KV-head pairs corresponds to different physical K/V entries. Therefore, $(l,g,t)$ tuples are counted independently rather than deduplicated only by token position $t$.

\paragraph{From candidates to realized singletons.}
The number of selected token positions is only an approximate measure of the floor scale and is not the actual physical entry cost. To realize a candidate $(l,g,t)$, GraceKV must first execute the root \textsc{Add} operation if the corresponding slot is uncovered and then repeatedly \textsc{Split} or \textsc{Extract} the frontier node containing $t$ until
\[
I_v=\{t\}.
\]

Let
\[
\mathcal{P}_t
\]
denote the set of not-yet-executed operations required to realize candidate $t$ under the current frontier $\mathcal{F}$. Its incremental physical cost is
\[
C(t\mid\mathcal{F})
=
\sum_{a\in\mathcal{P}_t}
c(a).
\]

When multiple floor candidates belong to the same tree, already expanded common ancestors are reused and are not charged again. Therefore, the joint cost of realizing several floor tokens can be smaller than the sum of their independently computed path costs.

After deducting the recent window, the base allocatable budget is
\[
B_{\mathrm{base}}
=
B_{\mathrm{total}}
-
LGr.
\]

Let $\mathcal{A}_{\mathrm{floor}}$ denote the deduplicated tree operations actually executed during the floor stage. These operations induce a set of physical frontier entries
\[
\mathcal{E}_{\mathrm{floor}},
\]
which contains the exact singleton leaves together with all root, internal, and sibling prototype entries materialized by the required paths and retained in the resulting frontier. An entry shared by multiple floor paths is included only once. Since $c(a)$ counts the number of newly introduced physical entries, the realized floor budget is equivalently
\[
B_{\mathrm{floor}}
=
\left|
\mathcal{E}_{\mathrm{floor}}
\right|
=
\sum_{a\in\mathcal{A}_{\mathrm{floor}}}
c(a),
\]
and is bounded by
\[
B_{\mathrm{floor}}
\leq
\left\lfloor
\alpha B_{\mathrm{base}}
\right\rfloor,
\qquad
\alpha=0.6.
\]

Initial candidate selection uses the number of candidate tokens to approximately control the floor scale, whereas actual execution is always charged according to the physical entries introduced by the required tree paths. The implementation additionally uses a per-head safeguard during this bounded floor stage to prevent a single layer--KV-head pair from consuming most of the protection budget. This safeguard applies only to the singleton-floor stage and is not a local quota in the main \textsc{Add}--\textsc{Split} auction.

Any floor budget that is not consumed during candidate realization is returned to the subsequent global auction. The successfully realized singleton-anchor set is denoted by
\[
\mathcal{S}_{\mathrm{floor}}
\subseteq
\mathcal{C}_{\mathrm{floor}}.
\]

Accordingly, a floor singleton refers only to a tuple
\[
(l,g,t)\in\mathcal{S}_{\mathrm{floor}}
\]
that is successfully expanded into a token-level frontier leaf. In contrast, $\mathcal{E}_{\mathrm{floor}}$ denotes the complete set of physical frontier entries reserved by the floor stage, including both these exact leaves and the other frontier entries required to realize them. The remaining auction budget is
\[
B_{\mathrm{free}}
=
B_{\mathrm{base}}
-
B_{\mathrm{floor}}.
\]

This gives the complete physical interpretation of the budget notation used in the main paper. The singleton floor is initiated by a collection of exact singleton anchors, while
\[
B_{\mathrm{floor}}
=
\left|
\mathcal{E}_{\mathrm{floor}}
\right|
\]
counts all deduplicated physical entries introduced by the corresponding \textsc{Add}, \textsc{Split}, and \textsc{Extract} paths.

The \texttt{w/o Singleton Floor} ablation removes this protection channel by setting
\[
\alpha=0,
\]
while keeping the prototype trees, token values, distortion estimates, and global auction unchanged.

\subsection{Interaction Between the Singleton Floor and Prototype Trees}
\label{app:floor_tree_interaction}

The singleton floor and the prototype-tree representation are not two parallel cache systems. A floor token becomes an exact cache entry by being isolated as a singleton leaf \emph{inside} its prototype tree. No second copy of its K/V pair is appended outside the tree frontier.

Suppose a current frontier node $v$ covers an interval $I_v$ containing one or more floor candidates that have not yet been isolated. A regular node compares curvature-aligned binary splitting with salient-token extraction. In contrast, a node containing an unrealized floor candidate is forced to use \textsc{Extract}.

If multiple floor candidates occur inside $I_v$, the highest-value candidate position $p$ is isolated first, producing up to three non-empty child intervals:
\[
[a_v,p),
\qquad
\{p\},
\qquad
[p+1,b_v).
\]

The singleton $\{p\}$ immediately becomes an exact leaf. If either side interval still contains unrealized floor candidates, forced extraction continues recursively until the budget-feasible candidates have been isolated.

For an exact singleton node $v=\{t\}$, prototype construction degenerates to
\[
\widetilde{k}_v=k_t,
\qquad
\widetilde{v}_v=v_t,
\qquad
p_v=t,
\qquad
m_v^{\mathrm{eff}}=1.
\]

Therefore, its multiplicity bias is
\[
\log m_v^{\mathrm{eff}}=0,
\]
and decoding uses the original post-RoPE K/V pair collected during prefill.

This construction satisfies the following non-duplication invariant:
\[
(l,g,t)\in\mathcal{S}_{\mathrm{floor}}
\Longrightarrow
t
\text{ belongs to exactly one current frontier interval}.
\]

Once $t$ is isolated as a singleton, its left and right sibling intervals no longer cover that position. Consequently:

\begin{itemize}
    \item a floor token is not simultaneously stored in a singleton and a sibling prototype;
    \item no second exact K/V copy is stored outside the tree;
    \item the packing stage requires no additional deduplication;
    \item shared ancestor paths are executed and charged only once; and
    \item floor singletons and ordinary frontier nodes jointly form the same physical ragged cache.
\end{itemize}

The notation distinguishes the successfully realized singleton anchors
$\mathcal{S}_{\mathrm{floor}}$ from the complete physical entry set
$\mathcal{E}_{\mathrm{floor}}$. The former contains only exact
layer--KV-head--token leaves, whereas the latter additionally contains
the root, internal, and sibling frontier entries materialized while
exposing those leaves. Physically, every element of
$\mathcal{E}_{\mathrm{floor}}$ is a node of the same final
prototype-tree frontier, and
\[
B_{\mathrm{floor}}
=
\left|
\mathcal{E}_{\mathrm{floor}}
\right|.
\]

When the text distinguishes ``ordinary frontier nodes'' from ``floor singletons,'' the distinction refers only to how they are selected: ordinary frontier nodes are exposed by the main auction, whereas floor singleton anchors are forced by the protection stage. They do not belong to separate cache structures.

If the floor cost cap is insufficient to fully isolate an initial candidate, that candidate is not included in
$\mathcal{S}_{\mathrm{floor}}$ and is not inserted as an additional exact entry. It may remain represented by a coarser prototype containing its position or may be refined later by the main auction. This preserves the total physical budget and avoids duplicate storage.

\subsection{Budget Utilization, Early Stopping, and FullKV Recovery}
\label{app:budget_utilization}

\paragraph{Budget utilization and early stopping.}
The abstract optimization constraint is
\[
\sum_{a\in\mathcal{A}'}
c(a)
\leq
B_{\mathrm{free}},
\]
rather than
\[
\sum_{a\in\mathcal{A}'}
c(a)
=
B_{\mathrm{free}}.
\]

The cache budget is therefore formally an upper bound, while the optimization objective is to maximize estimated utility. In principle, if no remaining operation provides a positive estimated improvement, allocating further entries is unnecessary, and the formulation permits the allocator to terminate before consuming the entire budget.

More generally, allocation can terminate when:

\begin{enumerate}
    \item the remaining budget is zero;
    \item all prototype trees have reached the finest available representation;
    \item the priority queue contains no operation whose prerequisites are satisfied;
    \item the remaining budget cannot accommodate the discrete cost of any candidate; or
    \item under a strict elastic stopping policy, no remaining candidate has positive marginal utility.
\end{enumerate}

The default configuration uses non-elastic allocation. Estimated gains are lower-bounded by zero when necessary, and legal zero-utility refinements remain admissible. Therefore, the fifth stopping condition is effectively inactive in the default setting. If a cost-two action does not fit the remaining budget, the allocator continues searching for cost-one actions instead of terminating immediately.

As a result, although the mathematical formulation allows unused budget, the default allocator continues to distribute entries whenever feasible tree refinements remain. Any residual under-utilization can only arise when the queue is exhausted or when the final discrete action costs cannot match a very small residual budget. Relative to the complete long-context cache budget, this difference is negligible in practice.

This design reconciles utility maximization with stable compression-ratio control: the inequality constraint preserves the principled possibility of early stopping, while the default non-elastic implementation makes the realized physical cache size closely track the target budget.

\paragraph{FullKV recovery.}
When the target compression ratio is
\[
x=1,
\]
the total context-cache budget becomes
\[
B_{\mathrm{total}}
=
LGN_{\mathrm{ctx}}.
\]

After deducting the recent window, the base budget available to the prototype trees is
\[
B_{\mathrm{base}}
=
LG(N_{\mathrm{ctx}}-r)
=
LGN_c.
\]

Consider a prototype tree covering $n$ original tokens. Admitting its root costs one entry. Every subsequent expansion of an internal node $v$ incurs the net cost
\[
|\operatorname{child}(v)|-1.
\]

Starting from an empty frontier, if the tree is fully expanded into $n$ token-level leaves, the total operation cost satisfies the telescoping identity
\[
1+
\sum_{v\in\mathcal{I}}
\left(
|\operatorname{child}(v)|-1
\right)
=
n,
\]
where $\mathcal{I}$ denotes the set of expanded internal nodes.

Thus, recovering all original tokens in a tree requires exactly $n$ physical cache entries, matching the corresponding FullKV cost.

The Add, Split, and Extract operations executed by the singleton floor form only a subset of this complete expansion path. They are not an additional set of cache entries. Since their realized cost is deducted from the main auction budget,
\[
B_{\mathrm{floor}}
+
B_{\mathrm{free}}
=
LGN_c
\]
when $x=1$.

Under the default non-elastic configuration, nonnegative and zero-utility refinement operations remain executable. Therefore, full expansion is not interrupted merely because an intermediate Split has zero immediate gain. Every compressible token eventually becomes a singleton leaf, while the recent window is already stored using its original entries.

For every $(l,g,t)$, the resulting cache therefore satisfies
\[
\widetilde{k}_{l,g,t}
=
k_{l,g,t},
\qquad
\widetilde{v}_{l,g,t}
=
v_{l,g,t},
\qquad
m^{\mathrm{eff}}_{l,g,t}=1.
\]

GraceKV consequently recovers the same token positions, K/V representations, and attention computation as FullKV.

The two operating extremes of GraceKV therefore have a unified interpretation:

\begin{itemize}
    \item under a limited budget, empty frontiers, internal prototypes, and singleton leaves implement true dropping, coarse information coverage, and fine local resolution, respectively; and
    \item when the budget is restored to the FullKV scale, all prototype trees expand to their original token-level leaves.
\end{itemize}

The singleton floor, greedy auction, and FullKV recovery are therefore not unrelated special cases. They are different operating states of the same prototype-tree representation space under the same physical cache-budget accounting.

\section{Experimental Details and Reproducibility}
\label{app:experimental_details}

\subsection{Models, Tasks, and Evaluation Settings}
\label{app:models_tasks_metrics}

\paragraph{Backbone Models.}
We evaluate GraceKV on multiple open-source large language models with different model scales and architectures, including Qwen2.5-3B, Qwen2.5-7B, Qwen2.5-14B, Llama-3.1-8B, and Mistral-7B. The main results in the paper are reported using Qwen2.5-7B, while the remaining models are used for cross-backbone evaluation and ablation studies. This setting allows us to examine whether the effectiveness of GraceKV depends on a particular model scale, attention configuration, or model family.

For each backbone, the model and its corresponding tokenizer are loaded from the same publicly available model release, ensuring consistency in the vocabulary, special tokens, positional encoding configuration, and input formatting. For models with native conversation templates, we use the chat template provided by the corresponding tokenizer. All models are directly used for inference without additional training, parameter updates, or task-specific fine-tuning.

GraceKV performs cache compression after one standard prefill pass. Therefore, adapting GraceKV to a new backbone only requires accounting for its number of layers, number of KV heads, and hidden-state dimensionality. No backbone-specific compression module needs to be trained.

\paragraph{Long-Context Tasks.}
We conduct experiments on LongBench and RULER, covering six representative types of long-context tasks:
\begin{itemize}
    \item single-document question answering;
    \item multi-document question answering;
    \item long-document summarization;
    \item few-shot generation;
    \item long-context aggregation;
    \item multi-target exact retrieval.
\end{itemize}

These tasks examine different requirements for compressed KV cache representations. Single-document question answering primarily requires locating local evidence within one long document. Multi-document question answering further requires connecting information distributed across distant context regions. Summarization emphasizes broad coverage of the input context, whereas few-shot generation requires preserving demonstration patterns, output formats, and locally precise information. Aggregation depends on integrating evidence distributed throughout the context, while retrieval requires accurately recovering multiple distant targets from a large amount of distractor content.

Together, these task categories cover two central requirements of KV cache compression: preserving broad information coverage and maintaining sufficiently high local resolution at critical positions. We follow the corresponding task protocols and evaluation settings of LongBench and RULER.

\paragraph{Input Processing.}
All methods use the same model input, tokenizer, and input template. For each sample, we explicitly distinguish the template prefix, long context, question, and template suffix, and record the position of the long context in the complete token sequence.

We represent the complete input as
\begin{equation}
X
=
X_{\mathrm{pre}}
\oplus
X_{\mathrm{ctx}}
\oplus
X_{\mathrm{post}},
\end{equation}
where \(X_{\mathrm{pre}}\) denotes the input-template prefix, \(X_{\mathrm{ctx}}\) denotes the long context, and \(X_{\mathrm{post}}\) contains the question and template suffix.

All contexts are retained in full throughout our experiments, without context truncation. Every method processes exactly the same tokenized sequence, ensuring that differences among compression methods are not caused by differences in the input content.

For the same sample, all methods additionally share the same prefill input, model parameters, question position, and generation configuration. Consequently, the observed performance differences primarily arise from the compressed KV cache constructed after prefill, rather than from differences in input formatting or context length.

\paragraph{Decoding Protocol.}
All task-quality evaluations use deterministic greedy decoding. Sampling and beam search are disabled, and each method generates a single response for every input using the same task-specific maximum generation length and stopping criteria.

\subsection{Evaluation Metrics}
\label{app:evaluation_metrics}

We use the official or equivalent evaluation metric associated with each task category. For readability, all scores reported in the main paper and appendix are multiplied by \(100\).

\paragraph{Question-Answering F1.}
Single-document and multi-document question-answering tasks are evaluated using token-level F1. Before evaluation, both the prediction and reference answer are normalized by converting text to lowercase, removing English articles and punctuation, and collapsing consecutive whitespace.

Let \(P\) and \(G\) denote the normalized token sequences of the prediction and reference answer, respectively. Precision and recall are defined as
\begin{equation}
\operatorname{Precision}
=
\frac{|P\cap G|}{|P|},
\qquad
\operatorname{Recall}
=
\frac{|P\cap G|}{|G|}.
\end{equation}

The corresponding F1 score is
\begin{equation}
\operatorname{F1}
=
\frac{
2\cdot
\operatorname{Precision}\cdot
\operatorname{Recall}
}{
\operatorname{Precision}
+
\operatorname{Recall}
}.
\end{equation}

When multiple acceptable reference answers are available, we compute the F1 score against each reference and retain the maximum value. This metric jointly measures the precision and coverage of the generated answer and permits partial token-level matches.

\paragraph{Rouge-L.}
Long-document summarization and few-shot generation are evaluated using word-level Rouge-L. Let
\(\operatorname{LCS}(P,G)\) denote the length of the longest common subsequence between prediction \(P\) and reference \(G\). The corresponding precision and recall are
\begin{equation}
P_{\mathrm{LCS}}
=
\frac{\operatorname{LCS}(P,G)}{|P|},
\qquad
R_{\mathrm{LCS}}
=
\frac{\operatorname{LCS}(P,G)}{|G|}.
\end{equation}

Rouge-L F1 is then defined as
\begin{equation}
\operatorname{Rouge\text{-}L}
=
\frac{
2P_{\mathrm{LCS}}R_{\mathrm{LCS}}
}{
P_{\mathrm{LCS}}
+
R_{\mathrm{LCS}}
}.
\end{equation}

Rouge-L measures content and ordering consistency through the longest common subsequence. It is therefore suitable for evaluating both long-form summaries and the agreement between generated outputs and reference formats in few-shot tasks.

\paragraph{Set Recall.}
Aggregation and multi-target retrieval tasks are evaluated using set recall. Let
\begin{equation}
\mathcal{Y}
=
\{y_1,y_2,\ldots,y_m\}
\end{equation}
denote the set of targets that should be recovered for a sample, and let \(\hat{y}\) denote the generated response. A target \(y_i\) is considered successfully recovered if it appears in the generated response. Set recall is defined as
\begin{equation}
\operatorname{SetRecall}
=
\frac{1}{m}
\sum_{i=1}^{m}
\mathbf{1}
\left[
y_i \subseteq \hat{y}
\right].
\end{equation}

In the implementation, both the generated response and target strings are converted to lowercase before matching.

Unlike semantic-similarity metrics, set recall directly verifies whether each target has been recovered. It therefore provides a strict measure of whether the compressed cache preserves distributed evidence and maintains the precise addressability required for long-context aggregation and multi-target retrieval.

\subsection{Hardware and Software Environment}
\label{app:hardware_software}

All experiments are conducted on NVIDIA A800 GPUs. Different experiments are executed in parallel according to the available computational resources. The number of GPUs is not treated as part of the model configuration, compression method, or evaluation protocol and is therefore not reported separately.

Model inference uses bfloat16 numerical precision. The primary software environment consists of:
\begin{itemize}
    \item Python 3.12;
    \item PyTorch 2.8;
    \item Transformers 5.13;
    \item CUDA 12.x.
\end{itemize}

All models are loaded in inference mode and evaluated using the same numerical precision. The feature collection, slot segmentation, prototype-tree construction, global resource allocation, and compressed-cache decoding stages of GraceKV are executed in the GPU environment and require no additional training process.

Task-quality experiments and system-efficiency measurements are reported separately. The measurement protocols for latency and memory, together with the construction overhead and amortization analysis, are described in the system-efficiency appendix to avoid conflating model quality with hardware-dependent runtime measurements.

\subsection{Fairness and Reproducibility}
\label{app:fairness_reproducibility}

For every model and task category, all methods use the same model weights, tokenizer, input template, tokenized context, question position, compression-ratio setting, generation configuration, and evaluation function.

The complete context is retained for every method, and no method-specific context truncation strategy is applied. Results are obtained under the same task partition and evaluation protocol, ensuring that the performance differences primarily reflect how effectively each compressed cache representation preserves contextual information.

The comparisons in the main paper cover token-eviction, KV-merging, and adaptive cache-allocation methods. Since different approaches may originally rely on different cache representations and system interfaces, our evaluation focuses on their task performance under a unified long-context evaluation protocol, rather than on reproducing every system-level optimization of their original implementations. FullKV is evaluated using the same backbone, inputs, and evaluation pipeline and serves as the uncompressed reference.


\section{Additional Results and Hyperparameter Sensitivity}
\label{app:additional_results}

\subsection{Cross-Backbone Results}
\label{app:cross_backbone}

To examine whether GraceKV depends on a particular model architecture,
family, or parameter scale, we further evaluate it on four additional
backbones: Llama-3.1-8B-Instruct, Qwen2.5-14B-Instruct,
Mistral-7B-Instruct-v0.3, and Qwen2.5-3B-Instruct. These models
cover the Qwen, Llama, and Mistral families and parameter scales from
3B to 14B. All methods follow the same evaluation protocol,
mandatory-region definition, and physical KV-entry budget used in the
main experiments. Tables~\ref{tab:llama8b_complete}--\ref{tab:qwen3b_complete}
report every evaluated compression ratio and cache budget. FullKV is
shown as the uncompressed reference and is excluded from the ranking
among compressed-cache methods. The best and second-best compressed-cache
results in each column are highlighted in bold and underlined, respectively.


\begin{table*}[!t]
\centering
{
\fontsize{7.5pt}{8.5pt}\selectfont
\setlength{\tabcolsep}{2pt}
\renewcommand{\arraystretch}{0.96}

\begin{tabularx}{\textwidth}{@{}l@{\hspace{3pt}}*{6}{Y}@{\hspace{10pt}}*{6}{Y}@{}}
\toprule
&
\multicolumn{6}{c}{
\textbf{LongBench (Single-Document QA, F1 $\times 100$)}
}
&
\multicolumn{6}{c}{
\textbf{LongBench (Multi-Document QA, F1 $\times 100$)}
}
\\
\cmidrule(lr){2-7}
\cmidrule(lr){8-13}
\textbf{Method}
& $4{\times}$ & $8{\times}$ & $16{\times}$
& $32{\times}$ & $64{\times}$ & $128{\times}$
& $4{\times}$ & $8{\times}$ & $16{\times}$
& $32{\times}$ & $64{\times}$ & $128{\times}$
\\
\midrule
FullKV
& \multicolumn{6}{c}{45.47}
& \multicolumn{6}{c}{31.19}
\\
\midrule
H2O
& 41.49 & 39.10 & 35.57 & 31.93 & 27.69 & 24.31
& 30.90 & 29.81 & 30.33 & 28.82 & 28.39 & 27.48
\\
StreamingLLM
& 32.76 & 28.86 & 25.79 & 24.24 & 22.35 & 21.83
& 25.80 & 23.98 & 24.17 & 24.15 & 23.07 & 22.95
\\
CaM
& 41.36 & 38.13 & 34.87 & 30.24 & 26.45 & 22.75
& 30.66 & 30.00 & 30.25 & 28.38 & 27.97 & 27.80
\\
SnapKV
& 44.80 & \underline{43.87} & 39.42 & 34.02
& 29.40 & \underline{24.46}
& \textbf{31.42} & \underline{31.01} & 30.08
& \textbf{29.96} & \underline{29.69} & \underline{28.35}
\\
D2O
& 40.61 & 37.62 & 35.66 & 32.05 & 27.21 & 23.00
& 30.26 & 29.47 & \underline{30.87} & 28.26 & 28.39 & 27.75
\\
PyramidKV
& \underline{44.81} & 43.02 & 39.01 & \underline{34.28}
& 29.57 & 22.78
& 31.14 & 30.12 & 30.19 & 28.90 & 27.64 & 26.44
\\
ChunkKV
& 44.70 & 43.08 & \underline{39.58} & 34.19
& \underline{30.86} & 22.75
& 31.11 & 30.91 & 30.37 & 29.21 & 29.40 & 28.12
\\
\textbf{GraceKV (Ours)}
& \textbf{44.94} & \textbf{44.13} & \textbf{40.37}
& \textbf{36.97} & \textbf{32.98} & \textbf{25.25}
& \underline{31.22} & \textbf{31.07} & \textbf{30.89}
& \underline{29.59} & \textbf{29.92} & \textbf{28.60}
\\
\bottomrule
\end{tabularx}

\par\medskip

\begin{tabularx}{\textwidth}{@{}l@{\hspace{3pt}}*{6}{Y}@{\hspace{10pt}}*{6}{Y}@{}}
\toprule
&
\multicolumn{6}{c}{
\textbf{LongBench (Summarization, Rouge-L $\times 100$)}
}
&
\multicolumn{6}{c}{
\textbf{LongBench (Few-Shot, Rouge-L $\times 100$)}
}
\\
\cmidrule(lr){2-7}
\cmidrule(lr){8-13}
\textbf{Method}
& $4{\times}$ & $8{\times}$ & $16{\times}$
& $32{\times}$ & $64{\times}$ & $128{\times}$
& $4{\times}$ & $8{\times}$ & $16{\times}$
& $32{\times}$ & $64{\times}$ & $128{\times}$
\\
\midrule
FullKV
& \multicolumn{6}{c}{16.29}
& \multicolumn{6}{c}{35.42}
\\
\midrule
H2O
& \textbf{15.85} & \textbf{15.52} & \underline{14.99}
& \underline{14.11} & \textbf{13.01} & \underline{12.09}
& 35.29 & 35.55 & 35.27 & \textbf{35.27}
& 34.47 & \underline{34.62}
\\
StreamingLLM
& 14.94 & 13.53 & 12.93 & 12.54 & 11.89 & 11.62
& \underline{35.55} & 34.44 & 34.63 & 34.46 & 33.96 & 34.37
\\
CaM
& 15.77 & 15.29 & 14.70 & 13.69 & 12.62 & 11.75
& 35.36 & 35.00 & 34.77 & 34.95 & \underline{34.91} & 34.24
\\
SnapKV
& 15.06 & 14.45 & 13.83 & 12.99 & 12.38 & 11.90
& 34.94 & 35.04 & 34.52 & 34.38 & 34.44 & \textbf{34.79}
\\
D2O
& \underline{15.82} & 15.43 & 14.62 & 13.93 & 12.75 & 11.83
& 35.26 & \underline{35.77} & 35.57 & 34.80 & 34.38 & 34.14
\\
PyramidKV
& 15.03 & 14.19 & 13.63 & 12.96 & 12.17 & 11.66
& 35.08 & 35.64 & 35.03 & 34.62 & 34.32 & 34.46
\\
ChunkKV
& 15.08 & 14.37 & 13.75 & 12.97 & 12.50 & 11.91
& 35.04 & 34.97 & \textbf{35.72} & 34.56 & 34.67 & 34.38
\\
\textbf{GraceKV (Ours)}
& \textbf{15.85} & \underline{15.51} & \textbf{15.03}
& \textbf{14.16} & \underline{12.92} & \textbf{12.19}
& \textbf{35.97} & \textbf{35.90} & \underline{35.69}
& \underline{35.16} & \textbf{34.95} & \underline{34.62}
\\
\bottomrule
\end{tabularx}

\par\medskip

\begin{tabularx}{\textwidth}{@{}l@{\hspace{4pt}}*{4}{Y}@{\hspace{10pt}}*{4}{Y}@{}}
\toprule
&
\multicolumn{4}{c}{
\textbf{RULER (Aggregation, Set Recall $\times 100$)}
}
&
\multicolumn{4}{c}{
\textbf{RULER (Retrieval, Set Recall $\times 100$)}
}
\\
\cmidrule(lr){2-5}
\cmidrule(lr){6-9}
\textbf{Method}
& 4K & 2K & 1K & 512
& 4K & 2K & 1K & 512
\\
\midrule
FullKV
& \multicolumn{4}{c}{89.00}
& \multicolumn{4}{c}{93.50}
\\
\midrule
H2O
& 87.33 & \underline{87.00} & 84.00 & 80.33
& 19.75 & 10.25 & 4.75 & 1.25
\\
StreamingLLM
& 70.67 & 62.00 & 54.33 & 49.33
& 25.75 & 13.75 & 6.25 & 2.75
\\
CaM
& \underline{87.67} & \textbf{87.33}
& \textbf{85.67} & \underline{84.00}
& 17.50 & 7.75 & 3.00 & 0.50
\\
SnapKV
& 85.33 & 75.00 & 67.00 & 53.00
& \textbf{95.25} & 94.75 & \underline{94.50} & \underline{95.00}
\\
D2O
& 86.67 & 86.00 & \underline{85.33} & \textbf{84.33}
& 20.00 & 9.25 & 3.75 & 0.25
\\
PyramidKV
& 83.67 & 74.67 & 67.67 & 53.67
& 93.75 & \underline{96.00} & 94.00 & 94.25
\\
ChunkKV
& \textbf{88.33} & 83.33 & 77.67 & 66.67
& \underline{95.00} & 93.25 & \textbf{94.75} & 92.00
\\
\textbf{GraceKV (Ours)}
& \underline{87.67} & 84.33 & 79.33 & 75.00
& \textbf{95.25} & \textbf{96.25}
& \underline{94.50} & \textbf{95.25}
\\
\bottomrule
\end{tabularx}
}
\caption{Complete cross-backbone results on
Llama-3.1-8B-Instruct. FullKV is the uncompressed reference.
The best and second-best compressed-cache results in each column
are highlighted in bold and underlined, respectively.}
\label{tab:llama8b_complete}
\end{table*}


\begin{table*}[!t]
\centering
{
\fontsize{7.5pt}{8.5pt}\selectfont
\setlength{\tabcolsep}{2pt}
\renewcommand{\arraystretch}{0.96}

\begin{tabularx}{\textwidth}{@{}l@{\hspace{3pt}}*{6}{Y}@{\hspace{10pt}}*{6}{Y}@{}}
\toprule
&
\multicolumn{6}{c}{
\textbf{LongBench (Single-Document QA, F1 $\times 100$)}
}
&
\multicolumn{6}{c}{
\textbf{LongBench (Multi-Document QA, F1 $\times 100$)}
}
\\
\cmidrule(lr){2-7}
\cmidrule(lr){8-13}
\textbf{Method}
& $4{\times}$ & $8{\times}$ & $16{\times}$
& $32{\times}$ & $64{\times}$ & $128{\times}$
& $4{\times}$ & $8{\times}$ & $16{\times}$
& $32{\times}$ & $64{\times}$ & $128{\times}$
\\
\midrule
FullKV
& \multicolumn{6}{c}{45.25}
& \multicolumn{6}{c}{37.94}
\\
\midrule
H2O
& 42.93 & 37.86 & 36.54 & 32.58 & 29.33 & 23.82
& 37.67 & 35.75 & 35.30 & 33.66 & 33.69 & 32.89
\\
StreamingLLM
& 30.16 & 27.88 & 26.09 & 25.01 & 22.79 & 23.19
& 29.13 & 28.42 & 25.54 & 25.98 & 25.20 & 25.21
\\
CaM
& 41.99 & 38.27 & 34.60 & 30.30 & 26.18 & 22.78
& \underline{38.35} & 34.84 & 34.08 & 33.68 & 32.05 & 29.20
\\
SnapKV
& 44.11 & 42.58 & 39.14 & \underline{34.45}
& \underline{29.37} & 24.21
& 37.89 & 38.35 & 37.86 & 37.29 & 33.73 & \underline{32.95}
\\
D2O
& 40.81 & 37.69 & 33.41 & 29.73 & 24.59 & 22.99
& 35.62 & 35.02 & 33.89 & 32.94 & 30.43 & 27.11
\\
PyramidKV
& 44.43 & \underline{43.13} & 37.85 & 33.08
& 29.06 & \underline{24.29}
& 38.04 & \underline{38.51} & \textbf{38.35}
& 37.10 & 33.38 & 32.48
\\
ChunkKV
& \underline{44.83} & 42.79 & \underline{39.57}
& 34.40 & 27.80 & 23.29
& 37.69 & 38.42 & 37.94 & \underline{37.30}
& \underline{34.76} & 28.66
\\
\textbf{GraceKV (Ours)}
& \textbf{45.04} & \textbf{43.19} & \textbf{41.32}
& \textbf{36.08} & \textbf{33.14} & \textbf{26.17}
& \textbf{38.45} & \textbf{38.63} & \underline{38.18}
& \textbf{37.57} & \textbf{35.03} & \textbf{33.02}
\\
\bottomrule
\end{tabularx}

\par\medskip

\begin{tabularx}{\textwidth}{@{}l@{\hspace{3pt}}*{6}{Y}@{\hspace{10pt}}*{6}{Y}@{}}
\toprule
&
\multicolumn{6}{c}{
\textbf{LongBench (Summarization, Rouge-L $\times 100$)}
}
&
\multicolumn{6}{c}{
\textbf{LongBench (Few-Shot, Rouge-L $\times 100$)}
}
\\
\cmidrule(lr){2-7}
\cmidrule(lr){8-13}
\textbf{Method}
& $4{\times}$ & $8{\times}$ & $16{\times}$
& $32{\times}$ & $64{\times}$ & $128{\times}$
& $4{\times}$ & $8{\times}$ & $16{\times}$
& $32{\times}$ & $64{\times}$ & $128{\times}$
\\
\midrule
FullKV
& \multicolumn{6}{c}{15.90}
& \multicolumn{6}{c}{36.74}
\\
\midrule
H2O
& 15.37 & 15.17 & \textbf{14.91} & \underline{14.10}
& 12.57 & \underline{12.32}
& 36.02 & \underline{36.16} & \underline{35.40}
& \textbf{35.55} & 34.03 & \textbf{33.64}
\\
StreamingLLM
& 14.51 & 13.68 & 13.13 & 12.80 & 12.37 & 12.05
& 34.62 & 33.60 & 33.24 & 33.19 & 32.71 & 32.91
\\
CaM
& \underline{15.62} & \textbf{15.24} & 14.69 & 13.87
& \textbf{12.80} & 11.97
& 35.77 & 35.19 & 34.44 & 34.38 & 33.70 & 33.48
\\
SnapKV
& 14.57 & 14.06 & 13.60 & 13.13 & 12.55 & 12.17
& \underline{36.32} & 35.73 & 34.84 & 35.02
& \underline{34.07} & \underline{33.62}
\\
D2O
& 15.41 & 15.20 & 14.72 & 13.81 & 12.55 & 11.87
& 36.12 & 35.79 & 35.16 & 33.95 & 33.42 & 33.22
\\
PyramidKV
& 14.40 & 13.93 & 13.41 & 12.95 & 12.39 & 12.17
& 35.93 & 35.12 & 34.23 & 34.50 & 33.38 & 33.24
\\
ChunkKV
& 14.70 & 13.96 & 13.64 & 13.07 & 12.61 & 12.16
& 35.69 & 35.11 & 34.52 & 33.43 & 33.70 & 33.25
\\
\textbf{GraceKV (Ours)}
& \textbf{15.82} & \underline{15.21} & \underline{14.88}
& \textbf{14.19} & \underline{12.77} & \textbf{12.64}
& \textbf{36.68} & \textbf{36.33} & \textbf{35.64}
& \underline{35.13} & \textbf{34.28} & 33.54
\\
\bottomrule
\end{tabularx}

\par\medskip

\begin{tabularx}{\textwidth}{@{}l@{\hspace{4pt}}*{4}{Y}@{\hspace{10pt}}*{4}{Y}@{}}
\toprule
&
\multicolumn{4}{c}{
\textbf{RULER (Aggregation, Set Recall $\times 100$)}
}
&
\multicolumn{4}{c}{
\textbf{RULER (Retrieval, Set Recall $\times 100$)}
}
\\
\cmidrule(lr){2-5}
\cmidrule(lr){6-9}
\textbf{Method}
& 4K & 2K & 1K & 512
& 4K & 2K & 1K & 512
\\
\midrule
FullKV
& \multicolumn{4}{c}{95.33}
& \multicolumn{4}{c}{69.00}
\\
\midrule
H2O
& 94.00 & 92.00 & 91.67 & 88.67
& 12.00 & 5.75 & 2.00 & 0.00
\\
StreamingLLM
& 84.67 & 78.67 & 70.00 & 63.33
& 20.00 & 12.25 & 6.25 & 2.75
\\
CaM
& 94.00 & \textbf{94.67} & \textbf{94.00} & \textbf{94.00}
& 14.75 & 6.25 & 2.00 & 0.00
\\
SnapKV
& 92.33 & 90.00 & 86.67 & 81.33
& 68.00 & 68.75 & 66.50 & \underline{58.50}
\\
D2O
& 93.67 & \underline{93.67} & \underline{93.33}
& \underline{89.67}
& 10.50 & 5.00 & 1.25 & 0.00
\\
PyramidKV
& 91.00 & 89.00 & 86.33 & 77.00
& \underline{69.00} & 68.50 & 67.25 & 57.25
\\
ChunkKV
& \underline{94.33} & 92.00 & 92.33 & 86.67
& 68.25 & \underline{69.50} & \underline{68.25} & \textbf{59.75}
\\
\textbf{GraceKV (Ours)}
& \textbf{94.67} & 91.00 & 88.00 & 82.67
& \textbf{69.50} & \textbf{69.75}
& \textbf{68.75} & \underline{58.50}
\\
\bottomrule
\end{tabularx}
}
\caption{Complete cross-backbone results on
Qwen2.5-14B-Instruct. FullKV is the uncompressed reference.
The best and second-best compressed-cache results in each column
are highlighted in bold and underlined, respectively.}
\label{tab:qwen14b_complete}
\end{table*}


\begin{table*}[!t]
\centering
{
\fontsize{7.5pt}{8.5pt}\selectfont
\setlength{\tabcolsep}{2pt}
\renewcommand{\arraystretch}{0.96}

\begin{tabularx}{\textwidth}{@{}l@{\hspace{3pt}}*{6}{Y}@{\hspace{10pt}}*{6}{Y}@{}}
\toprule
&
\multicolumn{6}{c}{
\textbf{LongBench (Single-Document QA, F1 $\times 100$)}
}
&
\multicolumn{6}{c}{
\textbf{LongBench (Multi-Document QA, F1 $\times 100$)}
}
\\
\cmidrule(lr){2-7}
\cmidrule(lr){8-13}
\textbf{Method}
& $4{\times}$ & $8{\times}$ & $16{\times}$
& $32{\times}$ & $64{\times}$ & $128{\times}$
& $4{\times}$ & $8{\times}$ & $16{\times}$
& $32{\times}$ & $64{\times}$ & $128{\times}$
\\
\midrule
FullKV
& \multicolumn{6}{c}{36.88}
& \multicolumn{6}{c}{28.19}
\\
\midrule
H2O
& 34.78 & 32.80 & \underline{32.26} & 27.85
& \underline{24.46} & 19.58
& 28.09 & 27.31 & 25.20 & 25.98 & 24.02 & 22.90
\\
StreamingLLM
& 26.05 & 24.12 & 22.80 & 20.45 & 18.83 & 18.68
& 20.89 & 18.62 & 16.25 & 16.33 & 17.75 & 17.04
\\
CaM
& 33.84 & 31.94 & 29.57 & 26.49 & 21.70 & 19.16
& 27.55 & 26.22 & 24.43 & 25.97 & 25.23 & 22.06
\\
SnapKV
& 36.19 & \underline{34.61} & 31.30 & 28.70
& 24.34 & \underline{21.04}
& 27.96 & \underline{27.81} & \textbf{27.38}
& \textbf{27.54} & 24.92 & 22.66
\\
D2O
& 33.85 & 31.75 & 29.66 & 26.54 & 21.11 & 19.64
& 27.51 & 26.32 & 24.91 & 25.27 & 24.57 & 22.71
\\
PyramidKV
& 36.08 & 34.26 & 30.95 & 29.20 & 23.92 & 19.49
& 28.06 & 27.14 & 26.71 & 26.67
& \underline{25.64} & \underline{23.01}
\\
ChunkKV
& \underline{36.48} & 34.25 & 31.88 & \underline{29.46}
& 24.18 & 20.84
& \underline{28.14} & 27.27 & 26.80 & \underline{26.97}
& 24.60 & 22.62
\\
\textbf{GraceKV (Ours)}
& \textbf{37.05} & \textbf{34.96} & \textbf{33.50}
& \textbf{30.01} & \textbf{25.46} & \textbf{21.52}
& \textbf{28.21} & \textbf{27.99} & \underline{26.89}
& \underline{26.97} & \textbf{25.85} & \textbf{23.81}
\\
\bottomrule
\end{tabularx}

\par\medskip

\begin{tabularx}{\textwidth}{@{}l@{\hspace{3pt}}*{6}{Y}@{\hspace{10pt}}*{6}{Y}@{}}
\toprule
&
\multicolumn{6}{c}{
\textbf{LongBench (Summarization, Rouge-L $\times 100$)}
}
&
\multicolumn{6}{c}{
\textbf{LongBench (Few-Shot, Rouge-L $\times 100$)}
}
\\
\cmidrule(lr){2-7}
\cmidrule(lr){8-13}
\textbf{Method}
& $4{\times}$ & $8{\times}$ & $16{\times}$
& $32{\times}$ & $64{\times}$ & $128{\times}$
& $4{\times}$ & $8{\times}$ & $16{\times}$
& $32{\times}$ & $64{\times}$ & $128{\times}$
\\
\midrule
FullKV
& \multicolumn{6}{c}{16.52}
& \multicolumn{6}{c}{38.25}
\\
\midrule
H2O
& \textbf{16.21} & \underline{15.89} & \underline{15.42}
& \underline{14.71} & \underline{13.61} & \underline{12.27}
& 37.39 & 37.23 & 37.25 & \textbf{38.01}
& \underline{37.58} & 36.46
\\
StreamingLLM
& 15.13 & 13.94 & 13.05 & 12.68 & 12.40 & \underline{12.27}
& 37.35 & \textbf{38.23} & \textbf{37.97}
& 36.65 & 37.19 & 36.52
\\
CaM
& \underline{16.17} & 15.74 & 15.21 & 14.54 & 13.30 & \underline{12.27}
& 37.63 & 37.01 & 36.65 & 36.79 & 36.91 & 36.92
\\
SnapKV
& 15.37 & 15.16 & 14.44 & 13.75 & 12.83 & 12.07
& 37.98 & \underline{37.78} & 36.78 & 36.69 & 37.01 & 37.09
\\
D2O
& 15.99 & 15.79 & 15.34 & 14.47 & 13.13 & 12.08
& 37.39 & 37.43 & 37.38 & 37.42 & 37.25 & 36.88
\\
PyramidKV
& 15.39 & 14.95 & 14.30 & 13.73 & 12.72 & 12.09
& \underline{38.47} & 37.59 & 36.27 & 37.28
& 37.41 & \underline{37.43}
\\
ChunkKV
& 15.60 & 15.04 & 14.38 & 13.67 & 12.90 & 12.06
& 37.86 & 36.88 & 36.90 & 37.31
& \underline{37.58} & 37.39
\\
\textbf{GraceKV (Ours)}
& 16.08 & \textbf{15.98} & \textbf{15.63}
& \textbf{15.03} & \textbf{14.11} & \textbf{12.89}
& \textbf{38.91} & 37.75 & \underline{37.56}
& \underline{37.79} & \textbf{37.67} & \textbf{38.07}
\\
\bottomrule
\end{tabularx}

\par\medskip

\begin{tabularx}{\textwidth}{@{}l@{\hspace{4pt}}*{4}{Y}@{\hspace{10pt}}*{4}{Y}@{}}
\toprule
&
\multicolumn{4}{c}{
\textbf{RULER (Aggregation, Set Recall $\times 100$)}
}
&
\multicolumn{4}{c}{
\textbf{RULER (Retrieval, Set Recall $\times 100$)}
}
\\
\cmidrule(lr){2-5}
\cmidrule(lr){6-9}
\textbf{Method}
& 4K & 2K & 1K & 512
& 4K & 2K & 1K & 512
\\
\midrule
FullKV
& \multicolumn{4}{c}{87.00}
& \multicolumn{4}{c}{74.25}
\\
\midrule
H2O
& \underline{85.33} & 82.33 & 77.33 & 71.67
& 12.75 & 4.75 & 0.75 & 0.00
\\
StreamingLLM
& \textbf{86.00} & \underline{82.67} & 73.33 & 53.67
& 23.50 & 11.75 & 6.00 & 2.75
\\
CaM
& 85.00 & \textbf{83.00} & \textbf{82.33} & 77.00
& 17.75 & 6.00 & 1.25 & 0.00
\\
SnapKV
& 81.00 & 78.00 & 76.67 & 70.33
& 73.00 & 71.50 & \underline{67.00} & 47.00
\\
D2O
& 83.33 & 82.00 & 80.33 & 76.00
& 12.75 & 4.75 & 1.50 & 0.00
\\
PyramidKV
& 80.00 & 78.00 & 76.67 & 68.00
& 73.00 & \underline{72.75} & 66.00 & 43.75
\\
ChunkKV
& 83.67 & \textbf{83.00} & \underline{81.00}
& \underline{77.33}
& \underline{73.50} & 71.25 & 66.00 & \underline{55.00}
\\
\textbf{GraceKV (Ours)}
& 84.00 & 80.67 & 80.00 & \textbf{79.33}
& \textbf{74.25} & \textbf{74.00}
& \textbf{68.50} & \textbf{55.25}
\\
\bottomrule
\end{tabularx}
}
\caption{Complete cross-backbone results on
Mistral-7B-Instruct-v0.3. FullKV is the uncompressed reference.
The best and second-best compressed-cache results in each column
are highlighted in bold and underlined, respectively.}
\label{tab:mistral7b_complete}
\end{table*}


\begin{table*}[!t]
\centering
{
\fontsize{7.5pt}{8.5pt}\selectfont
\setlength{\tabcolsep}{2pt}
\renewcommand{\arraystretch}{0.96}

\begin{tabularx}{\textwidth}{@{}l@{\hspace{3pt}}*{6}{Y}@{\hspace{10pt}}*{6}{Y}@{}}
\toprule
&
\multicolumn{6}{c}{
\textbf{LongBench (Single-Document QA, F1 $\times 100$)}
}
&
\multicolumn{6}{c}{
\textbf{LongBench (Multi-Document QA, F1 $\times 100$)}
}
\\
\cmidrule(lr){2-7}
\cmidrule(lr){8-13}
\textbf{Method}
& $4{\times}$ & $8{\times}$ & $16{\times}$
& $32{\times}$ & $64{\times}$ & $128{\times}$
& $4{\times}$ & $8{\times}$ & $16{\times}$
& $32{\times}$ & $64{\times}$ & $128{\times}$
\\
\midrule
FullKV
& \multicolumn{6}{c}{37.26}
& \multicolumn{6}{c}{20.26}
\\
\midrule
H2O
& 34.08 & 30.14 & 28.44 & 24.65 & 21.58 & \underline{20.97}
& 18.02 & 16.04 & 14.13 & 14.09 & 13.96 & 11.21
\\
StreamingLLM
& 26.54 & 24.05 & 22.39 & 21.63 & 20.08 & 19.83
& 14.02 & 11.49 & 9.62 & 9.78 & 10.35 & 10.39
\\
CaM
& 33.43 & 29.99 & 28.02 & 24.87 & 22.03 & 20.16
& 17.60 & 16.26 & 13.67 & 13.81 & 12.56 & 11.25
\\
SnapKV
& 36.04 & 32.54 & \underline{31.47} & \underline{27.46}
& \underline{23.23} & 19.61
& 19.28 & 19.50 & 18.74 & \underline{18.50}
& \underline{15.14} & \underline{13.00}
\\
D2O
& 33.00 & 29.27 & 26.83 & 23.24 & 21.56 & 19.93
& 15.49 & 14.41 & 13.28 & 12.59 & 11.85 & 10.19
\\
PyramidKV
& 35.64 & 32.59 & 29.34 & 25.42 & 20.72 & 19.24
& 19.88 & 19.19 & 18.87 & 17.15 & 14.76 & 9.76
\\
ChunkKV
& \underline{36.25} & \underline{32.89} & 29.19 & 25.53
& 22.14 & 19.98
& \underline{20.16} & \textbf{20.12} & \underline{19.67}
& 17.12 & 12.47 & 11.26
\\
\textbf{GraceKV (Ours)}
& \textbf{36.42} & \textbf{33.03} & \textbf{31.90}
& \textbf{28.40} & \textbf{24.35} & \textbf{21.01}
& \textbf{20.29} & \underline{19.92} & \textbf{19.99}
& \textbf{18.61} & \textbf{16.23} & \textbf{14.28}
\\
\bottomrule
\end{tabularx}

\par\medskip

\begin{tabularx}{\textwidth}{@{}l@{\hspace{3pt}}*{6}{Y}@{\hspace{10pt}}*{6}{Y}@{}}
\toprule
&
\multicolumn{6}{c}{
\textbf{LongBench (Summarization, Rouge-L $\times 100$)}
}
&
\multicolumn{6}{c}{
\textbf{LongBench (Few-Shot, Rouge-L $\times 100$)}
}
\\
\cmidrule(lr){2-7}
\cmidrule(lr){8-13}
\textbf{Method}
& $4{\times}$ & $8{\times}$ & $16{\times}$
& $32{\times}$ & $64{\times}$ & $128{\times}$
& $4{\times}$ & $8{\times}$ & $16{\times}$
& $32{\times}$ & $64{\times}$ & $128{\times}$
\\
\midrule
FullKV
& \multicolumn{6}{c}{15.55}
& \multicolumn{6}{c}{35.06}
\\
\midrule
H2O
& \textbf{15.04} & \underline{14.80} & 13.96
& \underline{13.38} & \underline{12.70} & \underline{12.04}
& \underline{33.88} & 33.15 & 32.57 & 31.65
& 31.26 & \underline{30.74}
\\
StreamingLLM
& 14.02 & 13.37 & 12.84 & 12.38 & 12.07 & 11.72
& 33.08 & 31.91 & 30.83 & 30.21 & 30.42 & 30.05
\\
CaM
& \underline{14.89} & \textbf{14.86} & \textbf{14.10}
& 13.30 & 12.28 & 11.53
& 33.51 & \underline{33.23} & 32.44 & \textbf{32.58}
& \underline{31.30} & 30.10
\\
SnapKV
& 14.44 & 14.18 & 13.56 & 13.06 & 11.90 & 11.95
& 33.66 & 33.15 & 32.12 & 32.14 & 30.52 & 30.03
\\
D2O
& 14.66 & 14.34 & 13.66 & 13.18 & 12.19 & 11.76
& 33.67 & 32.81 & \textbf{32.90} & 31.53 & 31.27 & 30.51
\\
PyramidKV
& 14.09 & 13.88 & 13.27 & 12.82 & 12.15 & 11.66
& 33.80 & 32.38 & 32.48 & 31.22 & 30.15 & 29.65
\\
ChunkKV
& 14.77 & 14.23 & 13.54 & 13.19 & 12.40 & 12.00
& 33.35 & 32.65 & 31.87 & 31.16 & 29.60 & 29.88
\\
\textbf{GraceKV (Ours)}
& 14.82 & 14.59 & \underline{13.99} & \textbf{13.70}
& \textbf{12.95} & \textbf{12.33}
& \textbf{34.01} & \textbf{33.35} & \underline{32.71}
& \underline{32.22} & \textbf{31.34} & \textbf{30.90}
\\
\bottomrule
\end{tabularx}

\par\medskip

\begin{tabularx}{\textwidth}{@{}l@{\hspace{4pt}}*{4}{Y}@{\hspace{10pt}}*{4}{Y}@{}}
\toprule
&
\multicolumn{4}{c}{
\textbf{RULER (Aggregation, Set Recall $\times 100$)}
}
&
\multicolumn{4}{c}{
\textbf{RULER (Retrieval, Set Recall $\times 100$)}
}
\\
\cmidrule(lr){2-5}
\cmidrule(lr){6-9}
\textbf{Method}
& 4K & 2K & 1K & 512
& 4K & 2K & 1K & 512
\\
\midrule
FullKV
& \multicolumn{4}{c}{78.67}
& \multicolumn{4}{c}{39.50}
\\
\midrule
H2O
& \textbf{77.00} & \textbf{76.33}
& \textbf{75.00} & \underline{68.67}
& 7.00 & 3.25 & 0.25 & 0.00
\\
StreamingLLM
& 23.33 & 15.67 & 13.33 & 8.33
& 18.50 & 10.75 & 6.25 & 2.75
\\
CaM
& 75.00 & \underline{74.00} & \underline{73.67}
& \textbf{71.00}
& 6.50 & 1.50 & 0.00 & 0.00
\\
SnapKV
& 73.67 & 69.00 & 68.33 & 59.33
& 40.00 & 40.75 & \textbf{40.25} & \textbf{32.75}
\\
D2O
& \underline{75.33} & \underline{74.00} & 72.67 & 66.00
& 5.25 & 1.25 & 0.00 & 0.00
\\
PyramidKV
& 72.00 & 69.33 & 61.33 & 56.33
& \textbf{40.75} & \textbf{41.75} & 39.25 & 28.75
\\
ChunkKV
& 74.33 & 72.00 & 65.00 & 61.00
& \underline{40.50} & 39.75 & \underline{39.50}
& \underline{32.25}
\\
\textbf{GraceKV (Ours)}
& 73.33 & 72.33 & 70.33 & \underline{68.67}
& \textbf{40.75} & \underline{41.50} & \underline{39.50} & \textbf{32.75}
\\
\bottomrule
\end{tabularx}
}
\caption{Complete cross-backbone results on
Qwen2.5-3B-Instruct. FullKV is the uncompressed reference.
The best and second-best compressed-cache results in each column
are highlighted in bold and underlined, respectively.}
\label{tab:qwen3b_complete}
\end{table*}

Across the 128 model--task--budget settings, GraceKV ranks first in
85 settings and second in another 26, placing among the top two in
111 settings. These results span three model families and parameter
scales from 3B to 14B, indicating that the allocation mechanism is
not tied to a single backbone family or model scale.

GraceKV is particularly consistent on question answering. It ranks
first in all 24 single-document QA settings. On multi-document QA,
it ranks first in 18 settings and second in the remaining six. The
advantage often becomes more pronounced as the budget tightens. At
$64\times$ compression, GraceKV improves over the strongest alternative
by 2.12, 3.77, 1.00, and 1.12 points on single-document QA for
Llama-3.1-8B-Instruct, Qwen2.5-14B-Instruct,
Mistral-7B-Instruct-v0.3, and Qwen2.5-3B-Instruct, respectively.
These results support the central motivation of GraceKV: under a
limited cache budget, joint coverage--resolution allocation can preserve
task-relevant information effectively as the available cache becomes
increasingly constrained.

The same trend extends beyond question answering. GraceKV ranks among
the top two in 21 of 24 summarization settings and 22 of 24 few-shot
settings. The absolute gaps on summarization are generally smaller,
as these tasks place greater emphasis on broad contextual coverage and
are therefore less sensitive to exact local representations. Nevertheless,
GraceKV remains competitive across these task categories and obtains the
highest score in many moderate- and high-compression settings. This
pattern shows that its performance extends beyond question answering to
tasks emphasizing broader contextual coverage.

RULER further reveals the different representation requirements of
aggregation and retrieval. Retrieval requires accurate localization of
one or more original evidence positions, and GraceKV ranks first in
12 of 16 settings and second in four more. In contrast, aggregation
favors broad and relatively uniform coverage, where coarse-merging
methods are often more competitive. These results indicate that
GraceKV's clearest advantage lies in tasks requiring precise evidence
localization, while the preferred coverage--resolution balance remains
task dependent.

\subsection{Sensitivity to the Singleton-Floor Budget and Multi-Hop Weight}
\label{app:sensitivity_floor_graph}

We study four hyperparameters that directly affect the allocation
behavior of GraceKV. All sensitivity experiments follow a
one-factor-at-a-time protocol: one parameter is varied while all
remaining parameters are fixed to their default values. We report F1
on LongBench single-document QA under $8\times$ and $64\times$
compression, representing moderate and tight cache budgets,
respectively. All scores are multiplied by 100.

\paragraph{Singleton-floor budget.}
Let $B_{\mathrm{pool}}$ denote the shared cache budget after deducting
mandatory recent-window entries but before charging the singleton
floor. The floor ratio $\alpha$ limits the physical cost of exact
singleton entries as
\begin{equation}
B_{\mathrm{floor}}
\leq
\alpha B_{\mathrm{pool}}.
\end{equation}
Setting $\alpha=0$ disables the singleton floor. Exact singleton entries
and subsequent \textsc{Add}/\textsc{Split} operations consume the same
physical cache budget, so increasing $\alpha$ does not introduce any
free entries. The default value is $\alpha=0.6$.

\paragraph{Multi-hop relevance weight.}
GraceKV combines direct relevance and relevance propagated over
token-level attention graphs:
\begin{equation}
R^{\mathrm{tok}}_{l,g,t}
=
(1-\lambda_{\mathrm{graph}})
\widehat{s}_{l,g,t}
+
\lambda_{\mathrm{graph}}
\widehat{d}_{t},
\end{equation}
where $\widehat{s}_{l,g,t}$ is normalized direct relevance and
$\widehat{d}_{t}$ is normalized multi-hop relevance. The default value
is $\lambda_{\mathrm{graph}}=0.3$.

\begin{table*}[!t]
\centering
{
\small
\setlength{\tabcolsep}{7pt}
\begin{tabular}{ccc|ccc}
\toprule
$\alpha$
& F1@$8\times$
& F1@$64\times$
& $\lambda_{\mathrm{graph}}$
& F1@$8\times$
& F1@$64\times$
\\
\midrule
0.0
& 40.44 & 29.97
& 0.00 & 41.39 & 31.09
\\
0.2
& 41.44 & 30.08
& 0.15 & \textbf{41.72} & 31.14
\\
0.4
& \underline{41.62} & \underline{31.02}
& $0.30^{\dagger}$ & \underline{41.58} & \textbf{31.49}
\\
$0.6^{\dagger}$
& 41.58 & \textbf{31.49}
& 0.50 & 41.33 & \underline{31.28}
\\
0.8
& \textbf{}{41.88} & 29.96
& 0.70 & 40.97 & 29.34
\\
\bottomrule
\end{tabular}
}
\caption{Sensitivity to the singleton-floor ratio $\alpha$ and the
multi-hop relevance weight $\lambda_{\mathrm{graph}}$. Each parameter
is varied independently while the others remain at their defaults.
Scores are F1 ($\times 100$). $\dagger$ denotes the default value.
The best and second-best results in each column are highlighted in
bold and underlined, respectively.}
\label{tab:sensitivity_floor_graph}
\end{table*}

As shown in Table~\ref{tab:sensitivity_floor_graph}, disabling the
singleton floor reduces F1 from 41.58 to 40.44 at $8\times$
compression and from 31.49 to 29.97 at $64\times$ compression.
This supports the motivation for the floor: purely step-wise marginal
allocation can underestimate the non-smooth benefit of exactly
retaining a small number of high-value tokens. Once entities, numbers,
labels, or answer-format cues are absorbed into coarse prototypes,
later refinements may not recover their exact representations within
a limited budget.

Performance generally improves as $\alpha$ increases from 0 to the
range 0.4--0.6. At $64\times$ compression, $\alpha=0.4$ and
$\alpha=0.6$ obtain 31.02 and 31.49, respectively, indicating a
stable intermediate region rather than an isolated optimum. Increasing
$\alpha$ further to 0.8 reduces the $64\times$ result to 29.96.
Under a tight budget, excessive spending on exact singletons leaves
insufficient capacity for \textsc{Add} operations that expand coverage
and \textsc{Split} operations that refine already covered regions.
The singleton floor is therefore most effective as a limited
exact-preservation guarantee rather than as the dominant allocation
mechanism.

GraceKV is also stable for
$\lambda_{\mathrm{graph}}\in[0.15,0.50]$. Direct relevance alone
remains a strong signal, obtaining 41.39 and 31.09 when
$\lambda_{\mathrm{graph}}=0$. Adding a moderate amount of propagated
relevance improves the moderate-budget result by recovering indirectly
connected evidence that may receive limited direct attention from the
question probes. The best values at $8\times$ and $64\times$ occur
at 0.3 and 0.5, respectively, showing that the preferred mixture can
vary mildly with the available budget without forming a narrow valid
region.

When $\lambda_{\mathrm{graph}}$ increases to 0.7, performance drops
under both budgets, with the $64\times$ score falling to 29.34.
Excessive propagation can diffuse value toward indirectly associated
but nonessential tokens and weaken the role of direct question
relevance. Such noise is especially harmful under tight budgets because
it displaces representations of truly critical evidence. We therefore
use $\lambda_{\mathrm{graph}}=0.3$, allowing multi-hop information to
complement rather than replace direct relevance.

\subsection{Sensitivity to the Target Slot Length and Distortion-Weight Ratio}
\label{app:sensitivity_slot_distortion}

We next vary the target slot length $n_0$ and the distortion-weight
ratio $\lambda_Z:\lambda_M$. The target length is a soft preference in
the dynamic-programming segmentation objective rather than a fixed
chunk size. Actual slot boundaries remain adaptive to curvature,
surprisal, within-slot representation dispersion, and the length
regularizer. The default target length is $n_0=64$.

For an interval $I$ and probe query $q$, GraceKV combines the
approximation error of the attention partition with that of the value
aggregate:
\begin{equation}
D_q(I)
=
\lambda_Z
\frac{
\left(Z_q(I)-\widetilde{Z}_q(I)\right)^2
}{
Z_q(I)^2+\epsilon
}
+
\lambda_M
\frac{
\left\|M_q(I)-\widetilde{M}_q(I)\right\|_2^2
}{
\left\|M_q(I)\right\|_2^2+\epsilon
}.
\end{equation}
Here, $Z_q(I)$ and $M_q(I)$ denote the original attention partition
and value aggregate, while $\widetilde{Z}_q(I)$ and
$\widetilde{M}_q(I)$ are their prototype-based approximations.
The default ratio is $\lambda_Z:\lambda_M=0.25:0.75$.

\begin{table*}[!ht]
\centering
{
\small
\setlength{\tabcolsep}{7pt}
\begin{tabular}{ccc|ccc}
\toprule
$n_0$
& F1@$8\times$
& F1@$64\times$
& $\lambda_Z:\lambda_M$
& F1@$8\times$
& F1@$64\times$
\\
\midrule
32
& 41.50 & 30.44
& $0:1$ & \textbf{42.16} & 30.93
\\
48
& \textbf{41.95} & \underline{31.20}
& $(0.25:0.75)^{\dagger}$
& 41.58 & \textbf{31.49}
\\
$64^{\dagger}$
& 41.58 & \textbf{31.49}
& $0.5:0.5$ & \underline{41.72} & 30.66
\\
96
& 41.71 & 31.10
& $0.75:0.25$ & 41.62 & 30.58
\\
128
& \underline{41.74} & \underline{31.19}
& $1:0$ & 41.35 & \underline{30.95}
\\
\bottomrule
\end{tabular}
}
\caption{Sensitivity to the target slot length $n_0$ and the
distortion-weight ratio $\lambda_Z:\lambda_M$. Each parameter is
varied independently while the others remain at their defaults.
Scores are F1 ($\times 100$). $\dagger$ denotes the default value.
The best and second-best results in each column are highlighted in
bold and underlined, respectively.}
\label{tab:sensitivity_slot_distortion}
\end{table*}

GraceKV remains stable over a reasonably broad range of target slot
lengths. For $n_0\in\{48,64,96,128\}$, the scores vary by only 0.37
points at $8\times$ compression, from 41.58 to 41.95, and by 0.39
points at $64\times$ compression, from 31.10 to 31.49. The default
value $n_0=64$ achieves the best result at $64\times$ compression and
is 0.37 points below the best result at $8\times$ compression. These
results indicate that GraceKV does not rely on a narrowly tuned target
slot length.

This robustness follows from two levels of adaptivity. First, $n_0$
acts only as a soft regularizer, allowing dynamic programming to move
the actual slot boundaries according to the hidden-state trajectory.
Second, the initial slot determines only the interval covered by a tree
root and does not fix the final cache resolution; subsequent
\textsc{Split} operations can still refine selected regions.
Consequently, moderate changes to the initial slot granularity have
only a limited influence on the final representation.

A shorter target length of $n_0=32$ is less favorable, particularly
under $64\times$ compression, where its score decreases to 30.44,
compared with the best score of 31.49. Shorter slots produce more tree
roots, and covering each slot requires at least one \textsc{Add}
operation. This increases the base cost of maintaining broad contextual
coverage and leaves less budget for refining high-value regions. We
therefore select $n_0=64$ as a balanced default between the number of
tree roots, structural adaptivity, and the available refinement space
within each tree, rather than because it is the only effective value.

The distortion-weight study reveals a budget-dependent trade-off
between the two distortion terms. At $8\times$ compression, using only
the value-aggregation term, i.e.,
$\lambda_Z:\lambda_M=0:1$, achieves the highest score of 42.16. The
other weight settings produce scores between 41.35 and 41.72, with the
default ratio $0.25:0.75$ obtaining 41.58. This suggests that the
value-aggregation signal is particularly informative when the cache
budget is relatively sufficient.

Under the tighter $64\times$ budget, however, the default ratio
$\lambda_Z:\lambda_M=0.25:0.75$ achieves the best score of 31.49.
Removing the partition term, i.e., using $0:1$, reduces the score by
0.56 points to 30.93, while removing the value-aggregation term, i.e.,
using $1:0$, reduces it by 0.54 points to 30.95. The balanced ratio
$0.5:0.5$ and the partition-heavier ratio $0.75:0.25$ obtain 30.66 and
30.58, respectively. These results show that the preferred weighting
changes with the available cache budget: the value-aggregation term
alone is effective at moderate compression, whereas an appropriately
weighted combination of both terms is more effective under severe
compression. We therefore use $0.25:0.75$ as the default because it
performs best in the more challenging $64\times$ setting while
remaining effective at $8\times$ compression.

A merged prototype simultaneously changes the local softmax partition
and the value aggregate returned by attention. Retaining both
distortion components therefore provides a more complete estimate of
the gains of \textsc{Add} and \textsc{Split} operations under severe
compression. Overall, noticeable degradation occurs mainly in
mechanistically extreme configurations, such as disabling the
singleton floor, assigning excessive budget to exact singletons,
overemphasizing propagated relevance, or constructing overly short
initial slots. Although the default configuration is not the isolated
point-wise optimum on every axis and budget, it provides consistently
strong performance across moderate and tight compression without
task-specific hyperparameter tuning.


\section{System Efficiency and Construction Overhead}
\label{app:system_efficiency}

This section separately analyzes the physical memory footprint of the
compressed KV cache, post-construction steady-state decoding latency,
one-time prototype-tree construction cost, and analytical amortization
regimes. Separating these quantities distinguishes the persistent memory
and per-token decoding benefits from the one-time cost incurred before
generation.

Unless otherwise specified, the experiments are conducted with
Qwen2.5-7B-Instruct on a single NVIDIA GeForce RTX 5090 GPU. The
software environment consists of PyTorch 2.8.0, CUDA 12.8, and
Transformers 5.13.1. We use bfloat16 precision and a batch size of one.
All inputs are processed according to their actual token lengths
without padding them to a fixed context length. The prefix, question
suffix, and template suffix are retained exactly and participate in
both cache storage and subsequent decoding. The recent window at the
end of the context is also retained exactly and counted toward the
cache budget.

GraceKV performs only one standard prefill and collects the hidden
states, Q/K/V representations, head-wise outputs, and token surprisal
required for subsequent compression during this pass. After prefill,
GraceKV constructs the prototype trees, performs global resource
allocation once, and materializes the selected nodes into a compressed
KV cache. The resulting cache can be directly used for autoregressive
generation without additional training or additional model forward
passes.

\subsection{Comparative Steady-State Decode Latency and KV Cache Memory}
\label{app:comparative_efficiency}

Figure~\ref{fig:analysis} in the main paper reports the KV cache memory and
post-construction per-token decoding latency of GraceKV across context
lengths of 6K, 12K, 18K,
24K, and 30K. It compares FullKV with GraceKV under $4\times$,
$8\times$, and $16\times$ compression.

At a context length of 30K, the KV cache of FullKV occupies 1.77 GB,
whereas GraceKV requires only 463 MB, 246 MB, and 137 MB under
$4\times$, $8\times$, and $16\times$ compression, respectively.
These configurations reduce cache memory by approximately 74\%, 86\%,
and 92\% relative to FullKV.

\begin{table}[!ht]
\centering
{
\small
\setlength{\tabcolsep}{7pt}
\begin{tabular}{lcc}
\toprule
\textbf{Cache Configuration}
& \textbf{KV Cache Memory}
& \textbf{Reduction} \\
\midrule
FullKV
& 1.77 GB
& -- \\
GraceKV, $4\times$
& 463 MB
& $\approx 74\%$ \\
GraceKV, $8\times$
& 246 MB
& $\approx 86\%$ \\
GraceKV, $16\times$
& 137 MB
& $\approx 92\%$ \\
\bottomrule
\end{tabular}
}
\caption{Key KV cache memory results at a context length of 30K,
corresponding to Figure~\ref{fig:analysis} in the main paper.}
\label{tab:efficiency_30k}
\end{table}

As the context length increases, FullKV must access an increasingly
large number of historical K/V entries at every generation step,
whereas the physical cache size of GraceKV remains constrained by the
specified budget. Consequently, the gap in cache-access cost grows with
the context length. At the longest context setting in Figure~\ref{fig:analysis}, the
per-token decoding latency of FullKV is up to approximately
$1.7\times$ that of GraceKV.

The KV cache memory reported in Figure~\ref{fig:analysis} corresponds to the actual
physical footprint of the final cache tensors rather than a theoretical
estimate obtained by directly dividing the FullKV memory by the nominal
compression ratio. The persistent cache of GraceKV contains the
retained original K/V entries, prototype K/V entries, effective
multiplicities, positions, valid masks, and per-head valid lengths.
Moreover, the prefix, question suffix, and recent window remain
uncompressed. The actual memory footprint therefore does not exactly
equal the FullKV footprint divided by the nominal compression ratio,
although it still decreases substantially as the budget becomes
tighter.

Table~\ref{tab:efficiency_tasks} further reports the cache size and
post-construction per-token decoding latency under an $8\times$ budget across representative task
categories. \textit{Entries} denotes the number of valid physical
layer--KV-head K/V entries, while \textit{Cache} includes both K/V
tensors and the metadata required by the cache implementation.

\begin{table*}[!ht]
\centering
{
\small
\setlength{\tabcolsep}{6pt}
\renewcommand{\arraystretch}{1.05}

\begin{tabular*}{\textwidth}{
@{\extracolsep{\fill}}
l
l
r
r
r
@{}
}
\toprule
\textbf{Task Category}
& \textbf{Cache Configuration}
& \textbf{Decode}
& \textbf{Memory}
& \textbf{Entries} \\
&
& \textbf{(ms/token)}
& \textbf{(MiB)}
& \\
\midrule

LongBench (Summarization)
& FullKV
& 23.23
& 181.3
& 353K \\
& GraceKV
& 23.33
& 53.9
& 90K \\
\midrule

LongBench (Single-Document QA)
& FullKV
& 23.15
& 294.9
& 586K \\
& GraceKV
& 23.02
& 51.7
& 94K \\
\midrule

LongBench (Few-Shot)
& FullKV
& 24.43
& 523.2
& 1.04M \\
& GraceKV
& 23.95
& 93.7
& 156K \\
\midrule

RULER (Aggregation)
& FullKV
& 25.63
& 877.0
& 1.75M \\
& GraceKV
& 24.06
& 130.3
& 228K \\
\midrule

RULER (Retrieval)
& FullKV
& 25.17
& 887.4
& 1.77M \\
& GraceKV
& 23.28
& 131.4
& 235K \\
\bottomrule
\end{tabular*}
}
\caption{KV cache memory and post-construction per-token decoding latency under an $8\times$
cache budget across representative task categories. Cache memory
includes K/V tensors and the required cache metadata.}
\label{tab:efficiency_tasks}
\end{table*}

GraceKV substantially reduces the physical KV cache footprint across
all task categories. For the two RULER settings, the cache is reduced from 877.0 MiB and
887.4 MiB to 130.3 MiB and 131.4 MiB, respectively, while the number
of effective cache entries decreases from approximately 1.75M to
approximately 230K.

The post-construction decoding advantage becomes more pronounced as the context length
increases. Under shorter contexts, model-weight access, fixed kernel
launches, and other computation still account for a considerable
fraction of the total latency, making the difference between cache
sizes relatively small. As the context becomes longer, historical K/V
access and attention computation contribute more substantially to the
overall cost, and the smaller cache of GraceKV provides increasingly
larger post-construction per-token savings.

In the few-shot
setting, GraceKV reduces latency from 24.43 to 23.95 ms/token. It reduces latency by 1.57 and
1.89 ms/token in the two RULER categories, corresponding to
approximately 6.1\% and 7.5\% lower steady-state per-token latency. These results
are consistent with Figure~\ref{fig:analysis}: the memory reduction takes effect
immediately after compression, while the decoding advantage increases
with the context length.

\subsection{Equal-Budget Interpretation}
\label{app:equal_budget}

GraceKV defines one cache-budget unit as one physically stored K/V
pair at a particular layer and KV head. Regardless of whether a cache
entry represents an original token, an exact singleton, or a prototype
of a contiguous interval, it participates in decoding as one physical
cache position.

Under the same physical-entry budget, the dominant K/V storage cost
and steady-state attention computation are therefore determined
primarily by the number of final entries rather than by how those
entries were selected or constructed. The role of GraceKV is to decide
where the limited cache resources should be placed:

\begin{itemize}
    \item which layer--KV-head combinations should receive more entries;
    \item which slots should first obtain information coverage through
    \textsc{Add};
    \item which covered regions should receive higher local resolution
    through \textsc{Split}; and
    \item which high-value tokens should be retained exactly as
    singletons.
\end{itemize}

These allocation decisions are made once during post-prefill
construction. The retained frontier entries, including exact
floor-selected singleton leaves and the other prototype nodes retained
along their tree paths, together with the recent window, are then
uniformly materialized as physical K/V cache entries, and subsequent
generation directly operates on the compressed cache.

Table~\ref{tab:equal_budget_decode} reports the physical cache sizes
and steady-state decoding costs of three compressed caches under the
same $8\times$ budget on LongBench single-document QA.

\begin{table}[!ht]
\centering
{
\small
\setlength{\tabcolsep}{7pt}
\begin{tabular}{lrrr}
\toprule
\textbf{Method}
& \textbf{Decode}
& \textbf{Cache}
& \textbf{Entries} \\
& \textbf{(ms/token)}
& \textbf{(MiB)}
& \\
\midrule
StreamingLLM
& 22.98
& 48.3
& 93K \\
SnapKV
& 23.00
& 48.3
& 94K \\
GraceKV
& 23.02
& 51.7
& 94K \\
\bottomrule
\end{tabular}
}
\caption{Physical cache sizes and steady-state decoding costs under
the same $8\times$ cache budget on LongBench single-document QA.}
\label{tab:equal_budget_decode}
\end{table}

The three methods retain similar numbers of physical entries and
therefore exhibit decoding latency and primary K/V storage within the
same range. This result shows that the multi-resolution prototypes of
GraceKV do not introduce decoding computation proportional to the
size of the original prototype trees. The trees are used only to
determine the final cache representation and are not retained or
traversed during generation.

Small differences in physical cache memory can remain even when the
number of valid entries is similar. These differences mainly arise
from:

\begin{itemize}
    \item different entry distributions across layers and KV heads;
    \item a small amount of padding caused by organizing each layer
    according to its longest KV head;
    \item the effective multiplicities, positions, valid masks, and
    per-head lengths stored by GraceKV;
    \item different cache layouts and memory-access patterns; and
    \item the use of a general multiplicity-aware ragged-attention
    implementation rather than a specialized fused kernel.
\end{itemize}

Accordingly, under an identical physical-entry budget and a unified
decoding implementation, compressed-cache methods generally exhibit
similar steady-state memory and decoding costs. The principal
distinction of GraceKV is not the use of additional decoding
computation, but its ability to organize the same number of entries
more effectively by jointly allocating information coverage and local
resolution.

\begin{table*}[!ht]
\centering
{
\small
\setlength{\tabcolsep}{3.5pt}
\renewcommand{\arraystretch}{1.08}

\begin{tabular}{
p{0.25\textwidth}
p{0.43\textwidth}
c
c
}
\toprule
\textbf{Cache Component}
& \textbf{Role}
& \textbf{In Budget}
& \textbf{Persistent} \\
\midrule

Key prototype or original Key
& Computes attention scores
& Yes
& Yes \\

Value prototype or original Value
& Produces the aggregated attention output
& Yes
& Yes \\

Effective multiplicity
& Restores the effective attention mass represented by a prototype
& No
& Yes \\

Position
& Records the positional coordinate of each cache entry
& No
& Yes \\

Valid mask
& Marks valid cache positions for each KV head
& No
& Yes \\

Per-head lengths
& Supports layer-wise and head-wise variable cache lengths
& No
& Yes \\

Complete prototype tree
& Constructs and evaluates candidate cache representations
& No
& No \\

\bottomrule
\end{tabular}
}
\caption{Components of the persistent GraceKV cache.
The complete prototype trees are discarded after the final cache
has been materialized. ``In Budget'' indicates whether a component
is counted toward the physical KV-entry budget.}
\label{tab:cache_components}
\end{table*}

The K/V entries constitute the dominant part of the persistent cache,
whereas the additional metadata is comparatively small. Complete
prototype trees and candidate nodes exist only during construction.
After cache packing, GraceKV retains only the selected frontier nodes
and the metadata required for decoding.

This also explains why Figure~\ref{fig:analysis} in the main paper primarily compares
FullKV with GraceKV. The purpose of Figure~\ref{fig:analysis} is to show how replacing
the complete cache with a budget-constrained cache changes memory and
cache-access costs as the context length grows. Under the same
physical-entry budget, compressed methods generally have similar
steady-state system costs; their central difference is how much useful
contextual information they preserve within those entries.

\subsection{One-Time Construction Latency and Peak Memory}
\label{app:construction_overhead}

GraceKV constructs the compressed cache once after prefill. This stage
transforms the representations collected during prefill into
curvature-guided slots, prototype trees, and a globally allocated cache
that can be directly used for decoding.

The complete construction procedure consists of:

\begin{enumerate}
    \item multi-layer, multi-scale curvature computation and slot
    segmentation;
    \item direct and multi-hop token-relevance computation;
    \item layer--KV-head sensitivity estimation;
    \item prototype-tree construction for all atomic units;
    \item singleton-floor selection and global \textsc{Add}--\textsc{Split}
    allocation; and
    \item packing the final frontier nodes into the ragged KV cache.
\end{enumerate}

In the reported measurement, the average construction time of GraceKV
is approximately 2.35 seconds. This one-time post-prefill cost is
incurred before autoregressive generation and is reported separately
from the steady-state per-token decoding latency.

\begin{table*}[!ht]
\centering
{
\small
\setlength{\tabcolsep}{4pt}
\renewcommand{\arraystretch}{1.08}

\begin{tabular*}{\textwidth}{
@{\extracolsep{\fill}}
l
c
p{0.53\textwidth}
@{}
}
\toprule
\textbf{Construction Stage}
& \textbf{Average Time}
& \textbf{Main Operations} \\
\midrule

Curvature segmentation
& 0.26 s
& Curvature aggregation, boundary scoring, and dynamic-programming
segmentation \\

Token relevance
& 0.02 s
& Direct and multi-hop token-relevance computation \\

Prototype-tree construction
& 1.39 s
& Interval statistics, prototype construction, distortion evaluation,
and candidate splitting \\

Global allocation
& 0.25 s
& Singleton-floor selection and global \textsc{Add}--\textsc{Split}
allocation \\

Cache packing
& 0.33 s
& Final-frontier collection and ragged-cache materialization \\

Other lightweight operations
& $\approx 0.10$ s
& Layer--KV-head sensitivity estimation and data organization \\

\midrule
\textbf{Total construction}
& \textbf{$\approx 2.35$ s}
& \textbf{Complete post-prefill cache construction} \\

\bottomrule
\end{tabular*}
}
\caption{Construction-time breakdown in the reported measurement.}
\label{tab:construction_breakdown}
\end{table*}

Prototype-tree construction is the largest component, accounting for
approximately 60\% of the total construction time. Each
layer--KV-head--slot atomic unit requires the evaluation of interval
representations, prototype distortion, and candidate split gains. In
contrast, token relevance, sensitivity estimation, and global
allocation introduce substantially smaller costs.

The current implementation controls construction overhead through
several engineering designs:

\begin{itemize}
    \item each non-leaf node retains only a small number of high-value
    split candidates;
    \item further refinement is skipped for sufficiently low-distortion
    intervals;
    \item a minimum leaf size constrains the overall tree size;
    \item high-value floor-selected positions can still be refined to
    exact singletons;
    \item interval statistics are computed in batches across layers,
    KV heads, and slots; and
    \item candidate trees and intermediate statistics are released
    after cache packing.
\end{itemize}

These designs allow GraceKV to complete global cache allocation across
layers, KV heads, and slots within seconds, without additional model
forward passes.

\paragraph{Peak memory.}
Prototype-tree construction temporarily maintains interval statistics,
candidate prototypes, attention-quality quantities, and batched tree
workspaces. This produces a temporary peak-memory increase after
prefill.

\begin{table}[!ht]
\centering
{
\small
\setlength{\tabcolsep}{7pt}
\begin{tabular}{lc}
\toprule
\textbf{Metric}
& \textbf{Memory Usage} \\
\midrule
Model and standard-prefill baseline
& $\approx 14.6$ GB \\
Peak during GraceKV construction
& $\approx 15.6$ GB \\
Typical temporary increase
& $\approx 1.0$--$1.5$ GB \\
Relative peak overhead
& $\approx 7\%$--$10\%$ \\
Summary value
& \textbf{$\approx 8\%$} \\
\bottomrule
\end{tabular}
}
\caption{Peak-memory overhead during GraceKV cache construction.
The relative overhead is measured against the model and standard
prefill memory baseline.}
\label{tab:construction_memory}
\end{table}

The approximately 8\% peak-memory overhead refers to the temporary
increase during construction relative to the model and standard
prefill memory baseline. Most of this increase is associated with
prototype-tree workspaces rather than the final KV cache.

Once global allocation and cache packing are complete, intermediate
tree nodes, candidate splits, and interval statistics are released.
Only the final compressed cache remains during generation. At a context
length of 30K, for example, the final cache can be reduced from
1.77 GB to 463 MB, 246 MB, or 137 MB. GraceKV therefore exchanges a
limited and temporary construction workspace for a persistent cache
reduction throughout subsequent generation.

The current implementation focuses primarily on the global resource
allocation algorithm and does not yet employ specialized low-level
operators for prototype-tree construction or multiplicity-aware
attention. The construction cost can be further reduced through:

\begin{itemize}
    \item parallel prototype-tree construction across atomic units;
    \item operator fusion for interval statistics, prototype
    computation, and distortion evaluation;
    \item batched processing of global candidate operations;
    \item a more compact flattened and packed cache layout; and
    \item specialized Triton or CUDA kernels for ragged attention and
    multiplicity correction.
\end{itemize}

Among these directions, parallel tree construction and operator fusion
are also the primary optimization opportunities identified in the
main paper.

\begin{table}[!ht]
\centering
{
\small
\setlength{\tabcolsep}{9pt}
\begin{tabular}{cc}
\toprule
\textbf{Number of Cache Uses $R$}
& \textbf{Average Cost per Use} \\
\midrule
1 & 2.35 s \\
2 & 1.18 s \\
4 & 0.59 s \\
8 & 0.29 s \\
\bottomrule
\end{tabular}
}
\caption{Analytical amortization of a representative 2.35-second construction
cost when the same question-conditioned compressed cache is reused.}
\label{tab:reuse_amortization}
\end{table}

\subsection{Amortization and Break-Even Analysis}
\label{app:break_even}

The construction cost of GraceKV is incurred only once after prefill,
whereas the reductions in cache memory and cache-access cost persist
for every subsequently generated token.

\begin{table*}[!ht]
\centering
{
\small
\setlength{\tabcolsep}{5pt}
\begin{tabular}{lrrrrr}
\toprule
\textbf{Task Category}
& \textbf{Avg.\ Context}
& \textbf{FullKV}
& \textbf{GraceKV}
& \textbf{Saving}
& \textbf{Break-Even} \\
&
& \textbf{(ms/token)}
& \textbf{(ms/token)}
& \textbf{(ms/token)}
& \textbf{Tokens} \\
\midrule
LongBench (Few-Shot)
& 9,029
& 24.43
& 23.95
& 0.48
& $\approx 4.9$K \\
RULER (Aggregation)
& 15,500
& 25.63
& 24.06
& 1.57
& $\approx 1.5$K \\
RULER (Retrieval)
& 15,670
& 25.17
& 23.28
& 1.89
& $\approx 1.2$K \\
\bottomrule
\end{tabular}
}
\caption{Representative time-based break-even points under
long-context settings, using an average GraceKV construction cost of
approximately 2.35 seconds.}
\label{tab:break_even_results}
\end{table*}

Let the one-time GraceKV construction time be
\begin{equation}
C_{\mathrm{build}},
\end{equation}
and let the average per-token decoding latency of FullKV and GraceKV be
\begin{equation}
d_{\mathrm{Full}}
\quad\text{and}\quad
d_{\mathrm{Grace}},
\end{equation}
respectively. When
\begin{equation}
d_{\mathrm{Full}} > d_{\mathrm{Grace}},
\end{equation}
the time-based break-even generation length can be estimated as
\begin{equation}
N_{\mathrm{BE}}
=
\left\lceil
\frac{1000C_{\mathrm{build}}}
{d_{\mathrm{Full}}-d_{\mathrm{Grace}}}
\right\rceil,
\label{eq:break_even}
\end{equation}
where $C_{\mathrm{build}}$ is measured in seconds and the decoding
latencies are measured in milliseconds per token.

Using an average construction cost of approximately 2.35 seconds,
Table~\ref{tab:break_even_results} reports representative amortization
results under long-context settings.

The estimated break-even lengths are longer than the typical outputs of
the evaluated LongBench and RULER tasks. Therefore, these benchmarks
primarily demonstrate the memory reduction and post-construction
decoding characteristics of GraceKV rather than single-query
end-to-end latency gains. Importantly, the reduction in physical
KV-cache memory takes effect immediately after compression and does not
depend on reaching the latency break-even point.

As the context length increases, FullKV must read an increasingly large
number of historical K/V entries at every decoding step, whereas the
physical cache size of GraceKV remains constrained by the specified
budget. Longer contexts therefore tend to produce larger per-token
savings and allow the one-time construction cost to be amortized over
fewer generated tokens.

The time-based break-even analysis captures only wall-clock decoding
latency and does not account for the broader system value of reduced
cache memory. GraceKV releases a substantial amount of KV cache memory
immediately after compression, even before the generated sequence
reaches the pure latency break-even point. The released memory can be
used to:

\begin{itemize}
    \item support longer input contexts;
    \item increase the batch size on the same GPU;
    \item improve concurrent request capacity;
    \item preserve additional space for model weights, activations,
    and runtime workspaces; and
    \item reduce memory pressure during long-context inference.
\end{itemize}

The compressed cache constructed by GraceKV can also be reused for
multiple generations that share the same complete prefill input,
including the same context, question suffix, and template. If the same
question-conditioned cache is used $R$ times, the average construction
cost assigned to each use becomes
\begin{equation}
\overline{C}_{\mathrm{build}}(R)
=
\frac{C_{\mathrm{build}}}{R}.
\label{eq:reuse_amortization}
\end{equation}

The amortization analysis identifies long-form generation and repeated
decoding from the same complete prefill state as favorable operating
regimes for GraceKV, including:

\begin{itemize}
    \item long-form text or code generation from a fixed prompt;
    \item multi-candidate sampling for the same question;
    \item beam search and branched decoding from a shared prefill state;
    \item self-consistency or reranking based on repeated generations
    from the same complete prompt; and
    \item repeated inference requests with an identical context,
    question suffix, and template.
\end{itemize}

These reuse cases preserve the question-conditioned prefill state used
to construct the final cache. GraceKV should not directly reuse the
same compressed cache for an arbitrary new question over the same
document. A new question may change the direct token relevance,
multi-hop relevance, and layer--KV-head sensitivity, which can alter
both the selected singleton anchors and the subsequent global
allocation.

A promising extension is to reuse only the question-independent
construction results across different questions. For example,
curvature-guided slot boundaries, candidate prototype trees, and
interval statistics could be cached for a document, while the
question-conditioned relevance, sensitivity, singleton-floor
selection, and global allocation are recomputed for each new question.
This would preserve question adaptivity while avoiding repetition of
the most expensive structural construction stages. Another possible
direction is to construct query-robust caches using relevance signals
aggregated over multiple representative questions, although such a
setting requires separate evaluation.

For repeated generations from the same complete prompt, curvature
segmentation, prototype-tree construction, and global allocation are
performed only once, while the compact cache can be reused throughout
subsequent decoding. As either the generation length or the number of
same-prompt cache uses increases, the one-time construction cost is
rapidly amortized, whereas the memory reduction and lower cache-access
cost continue to accumulate over the entire cache lifetime.

Overall, GraceKV provides substantial physical KV-cache memory reduction
and lower post-construction per-token decoding latency under long
contexts. The one-time construction cost of approximately 2.35 seconds
determines whether wall-clock latency savings can be realized for a
particular output length or cache-reuse pattern. Parallel tree
construction, operator fusion, specialized ragged-attention kernels,
and reuse of question-independent structural components may broaden the
operating regimes in which these steady-state savings translate into
end-to-end latency improvements.